\documentclass[10pt,journal,compsoc]{IEEEtran}
\ifCLASSOPTIONcompsoc
  \usepackage[nocompress]{cite}
\else
  \usepackage{cite}
\fi
\ifCLASSINFOpdf
\else
\fi

\usepackage{color}
\usepackage{array}
\usepackage{wrapfig}
\usepackage{units}
\usepackage{multirow}
\usepackage{amsmath}
\usepackage{graphicx}
\usepackage{times}
\usepackage{epsfig}
\usepackage{graphicx}
\usepackage{amsmath}
\usepackage{amssymb}
\usepackage{colortbl}
\usepackage{arydshln}
\usepackage{makecell, pict2e}
\usepackage{rotating}
\usepackage{textcomp}
\usepackage{gensymb}
\usepackage{pifont}
\usepackage{subfig}
\usepackage[export]{adjustbox}

\definecolor{gray}{gray}{0.65}
\definecolor{lightgray}{gray}{0.8}
\usepackage[pagebackref=false,breaklinks=true,letterpaper=true,colorlinks,bookmarks=false]{hyperref}

\def\paragraph{\@startsection{paragraph}{4}{\z@}{1.5ex plus 1.5ex minus 0.5ex}{0ex}{\normalfont\normalsize\bfseries}}

\begin{document}
\title{Person  \hspace{-0.07em}Recognition \hspace{-0.07em}in \hspace{-0.07em}Personal  \hspace{-0.07em}Photo  \hspace{-0.07em}Collections}

\author{Seong~Joon~Oh,\IEEEcompsocitemizethanks{
\IEEEcompsocthanksitem{S. Oh, R. Benenson, and M. Fritz were with the Computer Vision and Multimodal Computing Group, Max Planck Institute for Informatics, Saarbr\"{u}cken, Germany when this work was done; they are currently with LINE Plus (South Korea), Google (Switzerland), and CISPA Helmholtz Center i.G. (Germany), respectively. E-mail: coallaoh@linecorp.com, rodrigo.benenson@gmail.com, and fritz@cispa.saarland.}%
\IEEEcompsocthanksitem{B. Schiele is with the Computer Vision and Multimodal Computing Group, Max Planck Institute for Informatics, Saarbr\"{u}cken, Germany. E-mail: schiele@mpi-inf.mpg.de.}%
}%
        Rodrigo~Benenson,
        Mario~Fritz,
        and~Bernt~Schiele, {\em Fellow, IEEE}
        \thanks{Manuscript received ??; revised ??.}%
}% <-this % stops a space

%\IEEEcompsocitemizethanks{\IEEEcompsocthanksitem M. Shell was with the Department
%of Electrical and Computer Engineering, Georgia Institute of Technology, Atlanta,
%GA, 30332.\protect\\

%E-mail: see http://www.michaelshell.org/contact.html
%\IEEEcompsocthanksitem J. Doe and J. Doe are with Anonymous University.}% <-this % stops an unwanted space


\markboth{IEEE TRANSACTIONS ON PATTERN ANALYSIS AND MACHINE INTELLIGENCE,~Vol.~??, No.~??, ??~20??}%
{Shell \MakeLowercase{\textit{et al.}}: Bare Demo of IEEEtran.cls for Computer Society Journals}



\IEEEtitleabstractindextext{%
\begin{abstract}
People nowadays share large parts of their personal lives through social media. Being able to automatically recognise people in personal photos may greatly enhance user convenience by easing photo album organisation. For human identification task, however, traditional focus of computer vision has been face recognition and pedestrian re-identification. Person recognition in social media photos sets new challenges for computer vision, including non-cooperative subjects (e.g. backward viewpoints, unusual poses) and great changes in appearance. To tackle this problem, we build a simple person recognition framework that leverages convnet features from multiple image regions (head, body, etc.). We propose new recognition scenarios that focus on the time and appearance gap between training and testing samples. We present an in-depth analysis of the importance of different features according to time and viewpoint generalisability. In the process, we verify that our simple approach achieves the state of the art result on the PIPA \cite{Zhang2015CvprPiper} benchmark, arguably the largest social media based benchmark for person recognition to date with diverse poses, viewpoints, social groups, and events. 

Compared the conference version of the paper \cite{oh2015person}, this paper additionally presents (1) analysis of a face recogniser (DeepID2+ \cite{Sun2014ArxivDeepId2plus}), (2) new method \texttt{naeil2} that combines the conference version method \texttt{naeil} and DeepID2+ to achieve state of the art results even compared to post-conference works, (3) discussion of related work since the conference version, (4) additional analysis including the head viewpoint-wise breakdown of performance, and (5) results on the open-world setup.
%In a personal photo collection setting, person recognition involves the identification of individuals across viewpoint (e.g. head orientation, face occludsions) and time (e.g. clothing, hairstyle, and event changes). To tackle this problem, we build a simple person recognition system leveraging a standard convolutional neural network architecture (AlexNet \cite{Krizhevsky2012Nips}) and a specialised deep face feature (DeepID \cite{Sun2014ArxivDeepId2plus}). We verify that this simple approach achieves the state of the art result on a large dataset of Flickr personal photo albums (PIPA \cite{Zhang2015CvprPiper}), and present an in-depth analysis of the problem of recognising people under viewpoint and time variations. In particular, we report the impact of body parts, training data, and supervision signals; compare the robustness of different cues across viewpoint and time; and identify remaining challenges in the open world setting. In addition, the limitations of existing benchmarks (e.g. LFW, PIPA) are discussed and more challenging experimental protocols, with greater viewpoint and time variations, are proposed. To our knowledge, this is the first work to consider recognising people with fully general viewpoint, appearance changes, resolution, and occlusion levels. 
\end{abstract}

}

\maketitle

\IEEEdisplaynontitleabstractindextext

% \fi
\IEEEpeerreviewmaketitle

\IEEEraisesectionheading{\section{Introduction}\label{sec:introduction}}

\IEEEPARstart{W}{ith} the advent of social media and the shift of image capturing mode from digital cameras to smartphones and life-logging devices, users share massive amounts of personal photos online these days. Being able to recognise people in such photos would benefit the users by easing photo album organisation. Recognising people in natural environments poses interesting challenges; people may be focused on their activities with the face not visible, or can change clothing or hairstyle. These challenges are largely new -- traditional focus of computer vision research for human identification has been face recognition (frontal, fully visible faces) or pedestrian re-identification (no clothing changes, standing pose).

\begin{figure}
\begin{centering}
\arrayrulecolor{gray}
\par\end{centering}
\begin{centering}
\includegraphics[bb=0bp 0bp 601bp 600bp,width=0.25\columnwidth,height=0.25\columnwidth]{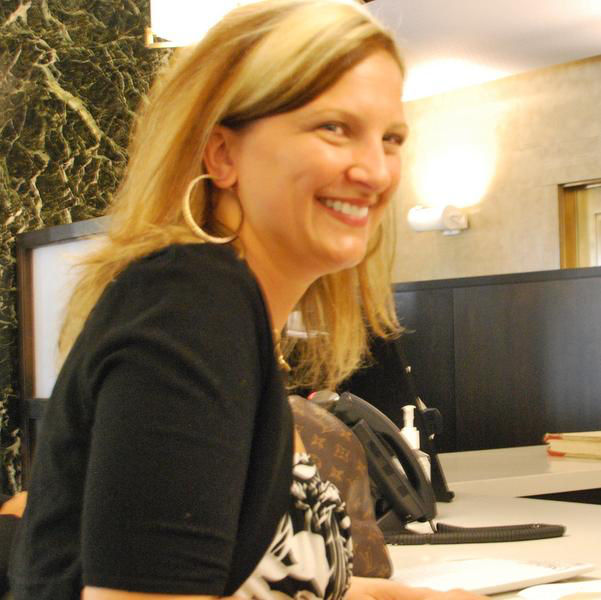}\includegraphics[bb=0bp 0bp 721bp 721bp,width=0.25\columnwidth,height=0.25\columnwidth]{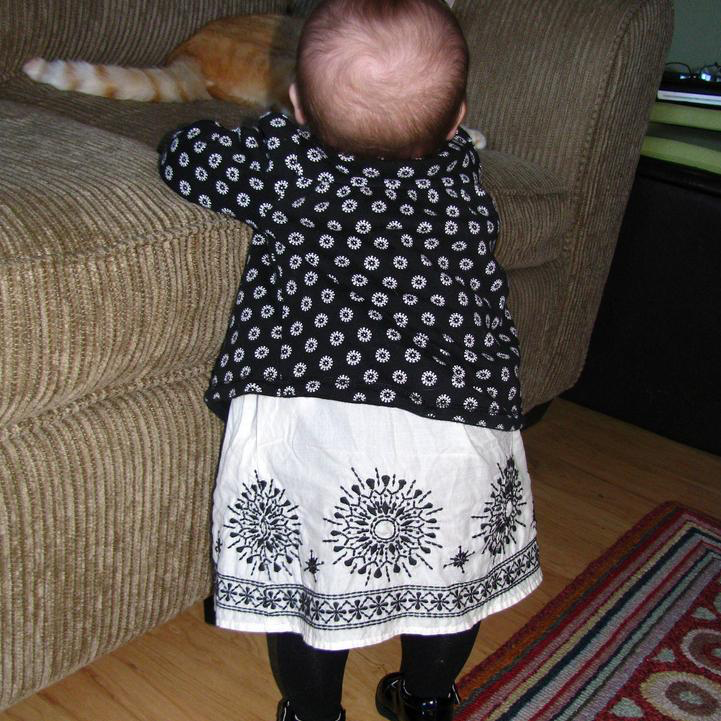}\includegraphics[bb=0bp 0bp 501bp 501bp,width=0.25\columnwidth,height=0.25\columnwidth]{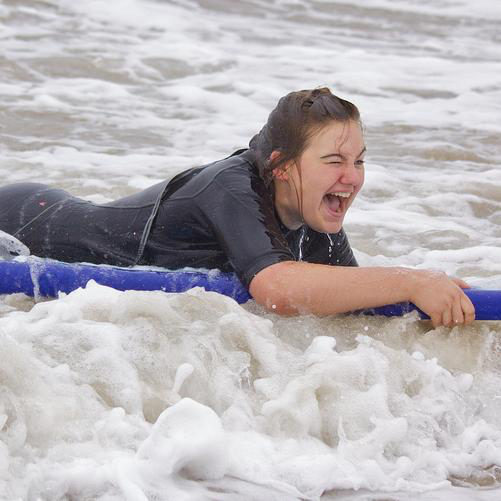}\includegraphics[bb=0bp 0bp 401bp 401bp,width=0.25\columnwidth,height=0.25\columnwidth]{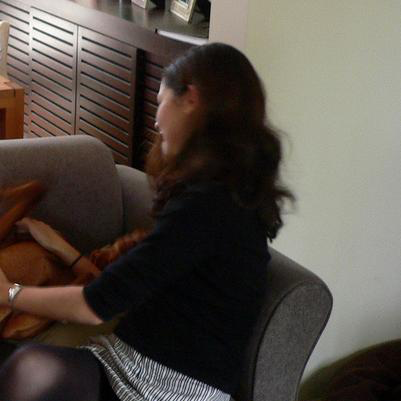}
\par\end{centering}
\vspace{0.5em}

\begin{raggedright}
\begin{tabular}{lcc|ccc|ccc|ccc}
\hspace*{-0.6em}Head & \hspace*{-0.6em}\includegraphics[height=0.8em]{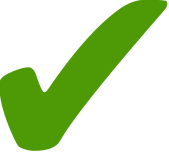} & \hspace*{-0.15em} & \hspace{1.0em} & \includegraphics[height=1em]{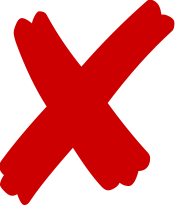} & \hspace{1.0em} & \hspace{1.0em} & \includegraphics[height=1em]{figures/no} & \hspace{1.0em} & \hspace{0.95em} & \includegraphics[height=1em]{figures/no} & \hspace{10em}\tabularnewline
\hspace*{-0.6em}Body & \hspace*{-0.6em}\includegraphics[height=0.8em]{figures/yes} & \hspace*{-0.17em} &  & \includegraphics[height=0.8em]{figures/yes} &  &  & \includegraphics[height=1em]{figures/no} &  &  & \includegraphics[height=1em]{figures/no} & \tabularnewline
\hspace*{-0.6em}{\scriptsize{}Attributes} & \hspace*{-0.6em}\includegraphics[height=0.8em]{figures/yes} & \hspace*{-0.17em} &  & \includegraphics[height=0.8em]{figures/yes} &  &  & \includegraphics[height=0.8em]{figures/yes} &  &  & \includegraphics[height=1em]{figures/no} & \tabularnewline
\hspace*{-0.6em}{\small{}All cues} & \hspace*{-0.6em}\includegraphics[height=0.8em]{figures/yes} & \hspace*{-0.17em} &  & \includegraphics[height=0.8em]{figures/yes} &  &  & \includegraphics[height=0.8em]{figures/yes} &  &  & \includegraphics[height=0.8em]{figures/yes} & \tabularnewline
\end{tabular}
\par\end{raggedright}
\arrayrulecolor{black}\vspace{0.5em}

\caption{\label{fig:teaser}In social media photos, depending on face occlusion or pose, different cues may be effective. For example, the surfer in the third column is not recognised using only head and body cues due to unusual pose. However, she is successfully recognised when additional attribute cues are considered.}
\end{figure}

Intuitively, the ability to recognise faces in the wild \cite{Huang2007Lfw,Sun2014ArxivDeepId2plus} is still an important ingredient. However, when people are engaged in an activity (i.e. not posing) their faces become only partially visible (non-frontal, occluded) or simply fully invisible (back-view). Therefore, additional information is required to reliably recognize people. We explore other cues that include (1) body of a person that contains information about the shape and appearance; (2) human attributes such as gender and age; and (3) scene context. See Figure \ref{fig:teaser} for a list of examples that require increasing number of contextual cues for successful recognition.

This paper presents an in-depth analysis of the person recognition task in the social media type of photos: given a few annotated training images per person, who is this person in the test image? The main contributions of the paper are summerised as follows: 
\begin{itemize}
\item{Propose realistic and challenging person recognition scenarios on the PIPA benchmark (\S\ref{sec:PIPA-dataset}).}
\item{Provide a detailed analysis of the informativeness of different cues, in particular of a face recognition module DeepID2+ \cite{Sun2014ArxivDeepId2plus} (\S\ref{sec:Cues-for-recognition}).}
\item{Verify that our journal version final model \texttt{naeil2} achieves the new state of the art performance on PIPA (\S\ref{sec:Test-set-results}).}
\item{Analyse the contribution of cues according to the amount of appearance and viewpoint changes (\S\ref{sec:challenges-analysis})}. 
\item{Discuss the performance of our methods under the open-world recognition setup (\S\ref{sec:open-world})}
\item Code and data are open source: available at \url{https://goo.gl/DKuhlY}.
\end{itemize}

\section{\label{sec:Related-work}Related work}

\subsection{By data type}

We review work on human identification based on various visual cues. Faces are the most obvious and widely studied cue, while other biometric cues have also been considered. We discuss how our personal photo setup is different from them.

\subsubsection*{Face}

The bulk of previous work on person recognition focuses on faces. The Labeled Faces in the Wild (LFW) \cite{Huang2007Lfw} has been a great testbed for a host of works on the face identification and verification outside the lab setting. The benchmark has saturated, attributing to the deep features \cite{Taigman2014CvprDeepFace,Sun2014ArxivDeepId2plus,Zhou2015ArxivNaiveDeepFace,Schroff2015ArxivFaceNet,Parkhi15,chen2016unconstrained,wen2016discriminative,ranjan2017l2,wang2018additive,deng2018arcface} trained on large scale face databases that outperform the traditional methods involving sophisticated classifiers based on hand-crafted features and metric learning approaches \cite{Guillaumin2009Iccv,Chen2013CvprBlessing,Cao2013IccvTransferLearning,Lu2014ArxivGaussianFace}. While faces are clearly the most discriminative cue for recognising people in personal photos as well, they are often occluded in natural footage - e.g. the person may be engaged in other activities. Since LFW contains largely frontal views of subjects, it does not fully represent the setup we are interested in. LFW is also biased to public figures. 

Some recent face recognition benchmarks have introduced more face occlusions. IARPA Janus Benchmark A (IJB-A) \cite{klare2015pushing} and Celebrities in Frontal Profile (CFP)~\cite{sankaranarayanan2016triplet} datasets include faces with profile viewpoints; however, both IJB-A and CFP do not consider subject fully turning away from the camera (back-view) and the subjects are limited to public figures. Age Database (AgeDB) \cite{moschoglou2017agedb} evaluates the recognition across long time span (years), but is again biased towards public figures; recognition across age gap is a part of our task and we focus on personal photos without celebrity bias. 

MegaFace \cite{kemelmacher2016megaface,nech2017level} is perhaps the largest known open source face database over personal photos on Flickr. However, MegaFace still does not contain any back-view subject and it is not designed to evaluate the ability to combine cues from multiple body regions. Face recognition datasets are not suitable for training and evaluating systems that identify a human from face and other body regions.

\subsubsection*{Pedestrian Re-Identification from RGB Images}

Not only face, but the entire body and clothing patterns have also been explored as cue for human identification. For example, pedestrian re-identification (re-id) tackles the problem of matching pedestrian detections in different camera views. First benchmarks include VIPeR \cite{gray2007VIPeR}, CAVIAR \cite{cheng2011CAVIAR}, CUHK \cite{li2012CUHK1}, while nowadays most re-id papers report results on Market 1501~\cite{zheng2015scalable}, MARS~\cite{zheng2016mars}, CUHK03~\cite{Li2014CvprDeepReID}, and DukeMTMC-reID~\cite{ristani2016performance}. There is an active line of research on pedestrian re-id, starting with hand-crafted features  \cite{Li2013Cvpr,Zhao2013IccvSalienceMatching,Bak2014WacvBrownian} and has moved towards deep feature based schemes \cite{Li2014CvprDeepReID,Yi2014Arxiv,Hu2014Accvw,ahmed2015improved,Cheng_2016_CVPR,Xiao_2016_CVPR,Varior2016,Chen/cvpr2017,zheng2017unlabeled}. 

However, the re-id datasets and benchmarks do not fully cover the social media setup in three aspects. (1) Subjects are pedestrians and mostly appear in the standing pose; in personal photos people may be engaged in a diverse array of activities and poses - e.g. skiing, performing arts, presentation. (2) Typically resolution is low; person recognition in personal photos includes the problem of matching identities across a huge resolution range - from selfies to group photos. 

\subsubsection*{Pedestrian Re-Identification from Depth Images}

In order to identify humans based on body shapes, potentially to enable recognition independent of clothing changes, researchers have proposed depth-based re-identification setups. Datasets include RGBD-ID~\cite{barbosa2012re}, IAS-Lab RGBD-ID \cite{munaro2014one}, and recent SOMAset \cite{barbosa2017looking}. SOMAset in particular has clothing changes enforced in the dataset. There is a line of work~\cite{barbosa2012re,munaro2014one,barbosa2017looking,wu2017robust} that has improved the recognition technology under this setup. While recognition across clothing changes is related to our task of identifying human in personal photos, the RGBD based re-identification typically requires depth information for good performance; for personal photos depth information is unavailable. Moreover, the relevant datasets are collected in controlled lab setup, while personal photos are completely unconstrained. 

\subsubsection*{Other biometric cues}

Traditionally, fingerprints and iris patterns have been considered strong visual biometric cues \cite{maltoni2009handbook,daugman2009iris}. Gaits \cite{CONNOR20181} are also known to be an identity correlated feature. We do not use them explicitly in this work, as such information is not readily given in personal photos.

\subsubsection*{Personal Photos}

Personal photos have distinct characteristics that set new challenges not fully addressed before. For example, people may be engaged in certain activity, not cooperating with the photographer, and people may change clothing over time. Some pre-convnet work have addressed this problem in a small scale \cite{zhang2003automated,song2006context,anguelov2007contextual,Gallagher2008Cvpr}, combining cues from face as well as clothing regions. Among these, Gallagher et al. \cite{Gallagher2008Cvpr} have published the dataset ``Gallagher collection person'' for benchmarking ($\sim\negmedspace 600$ images, 32 identities). It was not until the appearance PIPA dataset \cite{Zhang2015CvprPiper} was there a large-scale dataset of personal photos. The dataset consists of Flickr personal account images (Creative Commons) and is fairly large in scale ($\sim$40k images, $\sim$2k identities), with diverse appearances and subjects with all viewpoints and occlusion levels. Heads are annotated with bounding boxes each with an identity tag. We describe PIPA in greater detail in \S \ref{sec:PIPA-dataset}.

\subsection{By recognition task}

There exist multiple tasks related to person recognition \cite{Gong2014PersonReIdBook} differing mainly in the amount of training and testing data. Face and surveillance re-identification is most commonly done via verification: \emph{given one reference image (gallery) and one test image (probe), do they show the same person?} \cite{Huang2007Lfw,Bedagkar2014IvcPersonReIdSurvey}. In this paper, we consider two recognition tasks. 
\begin{itemize}
\item Closed world identification: \emph{given a single test image (probe), who is this person among the identities that are among the training identities (gallery set)?}
\item Open world recognition \cite{kemelmacher2016megaface} (\S\ref{sec:open-world}) : \emph{given a singe test image (probe), is this person among the training identities (gallery set)? If so, who?}
\end{itemize}
Other related tasks are, face clustering \cite{Cui2007ChiEasyAlbum,Schroff2015ArxivFaceNet,OttoClustering}, finding important people \cite{mathialagan2015vip}, or associating names in text to faces in images  \cite{Everingham2006Bmvc,Everingham2009Ivc}.

\subsection{Prior work on PIPA dataset~\cite{Zhang2015CvprPiper}}

Since the introduction of the PIPA dataset \cite{Zhang2015CvprPiper}, multiple works have proposed different methods for solving the person recognition problem in social media photos. Zhang et al. proposed the Pose Invariant Person Recognition (\texttt{PIPER}) \cite{Zhang2015CvprPiper}, obtaining promising results by combining three ingredients: DeepFace \cite{Taigman2014CvprDeepFace} (face recognition module trained on a large private dataset), poselets \cite{Bourdev2009IccvPoselets} (pose estimation module trained with 2k images and 19 keypoint annotations), and convnet features trained on detected poselets \cite{Krizhevsky2012Nips,Deng2009CvprImageNet}.

Oh et al. \cite{oh2015person}, the conference version of this paper, have proposed a much simpler model \texttt{naeil} that extracts AlexNet cues from multiple \emph{fixed} image regions. In particular, unlike \texttt{PIPER} it does not require data-heavy DeepFace or time-costly poselets; it uses only 17 cues (\texttt{PIPER} uses over 100 cues); it still outperforms \texttt{PIPER}.

There have been many follow-up works since then. Kumar et al. \cite{kumar2017pose} have improved the performance by normalising the body pose using pose estimation. Li et al. \cite{li2017sequential} considered exploiting people co-occurrence statistics. Liu et al. \cite{liu_2017_coco} have proposed to train a person embedding in a metric space instead of training a classifier on a fixed set of identities, thereby making the model more adaptable to unseen identities. We discuss and compare against these works in greater detail in \S\ref{sec:Test-set-results} Some works have exploited the photo-album metadata, allowing the model to reason over different photos \cite{joon16eccv,Li_2016_CVPR}.

In this journal version, we build \texttt{naeil2} from \texttt{naeil} and DeepID2+ \cite{Sun2014ArxivDeepId2plus} to achieve the state of the art result among the published work on PIPA. We provide additional analysis of cues according to time and viewpoint changes.

\section{\label{sec:PIPA-dataset}Dataset and experimental setup}

\subsubsection*{Dataset}

\begin{figure*}
\begin{centering}
\includegraphics[bb=0bp 0bp 2275bp 1536bp,width=1.8\columnwidth]{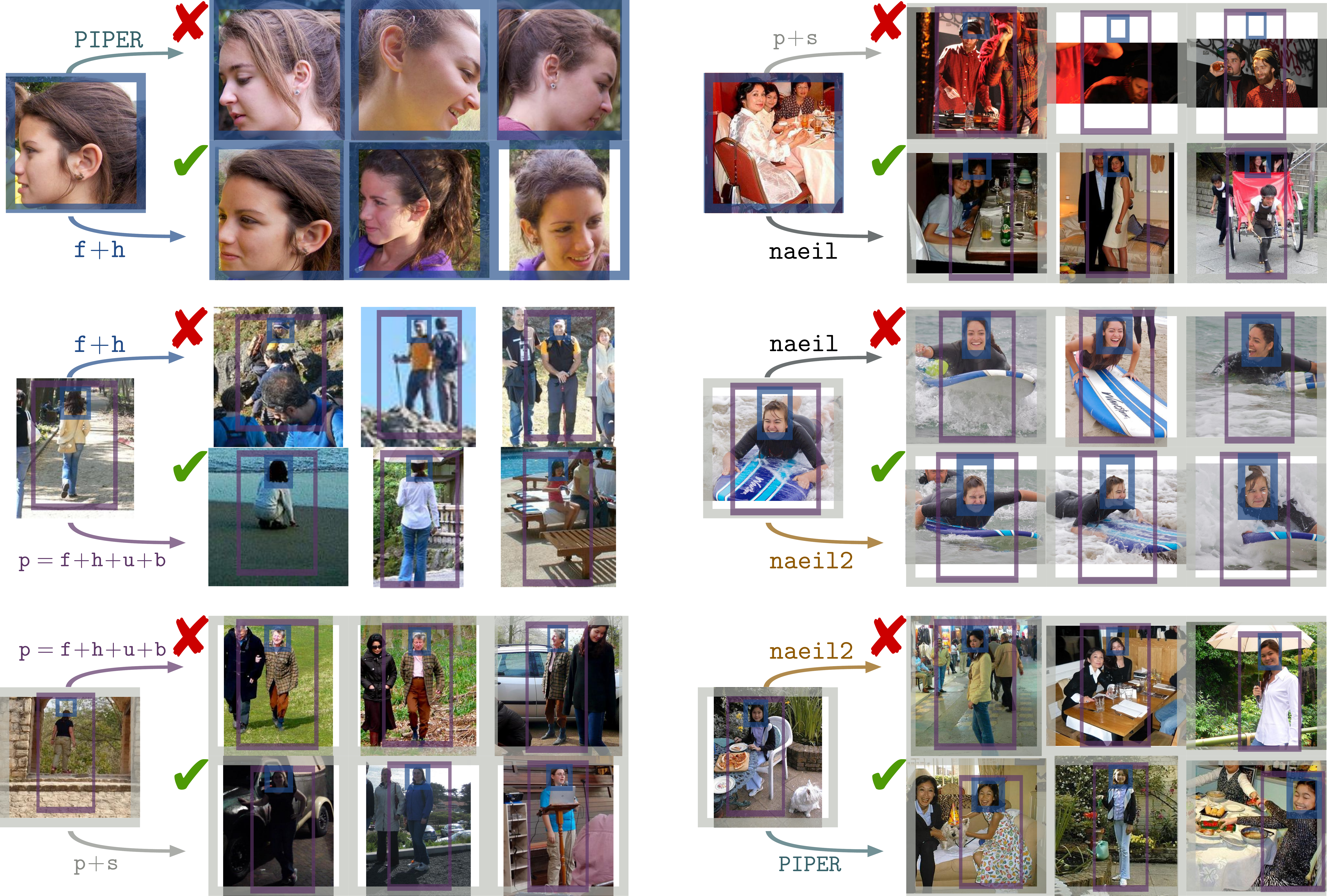}
\par\end{centering}
\begin{raggedright}
\medskip{}
\par\end{raggedright}
\caption{\label{fig:success-O-split} Success and failure cases on the Original split. Single images: test examples. Arrows point to the training samples for the predicted identities. Green and red crosses indicate correct and wrong predictions.}
\vspace{-1em}
\end{figure*}

\begin{figure*}
\begin{centering}
\includegraphics[bb=0bp 0bp 797bp 501bp,width=1.8\columnwidth]{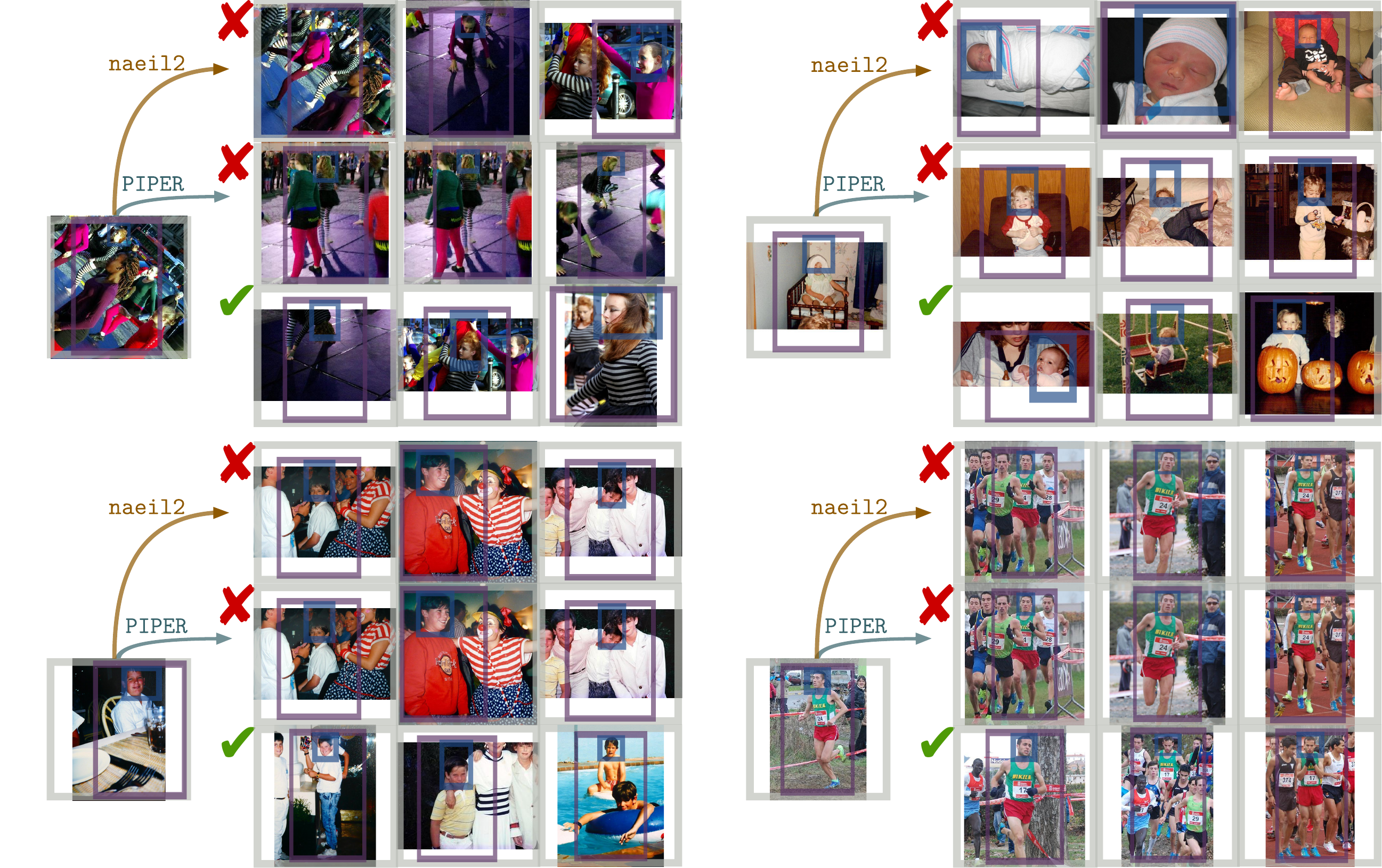}
\par\end{centering}
\begin{raggedright}
\medskip{}
\par\end{raggedright}
\caption{\label{fig:failure-O-split} Failure cases of \texttt{naeil2} and $\texttt{PIPER}$ on the Original split. Single images: test examples. Arrows point to the training samples for the predicted identities. Green and red crosses indicate correct and wrong predictions. Typical hard cases are: 1) uniform clothing (top left), 2) babies (top right), 3) children (bottom left), and 4) annotation errors (bottom right). }
\vspace{-1em}
\end{figure*}

The PIPA dataset (``People In Photo Albums'') \cite{Zhang2015CvprPiper} is, to the best of our knowledge, the first dataset to annotate people's identities even when they are pictured from the back. The annotators labelled instances that can be considered hard even for humans (see qualitative examples in figure \ref{fig:success-O-split}, \ref{fig:failure-O-split}). PIPA features $37\,107$ Flickr personal photo album images (Creative Commons license), with $63\,188$ head bounding boxes of $2\,356$ identities. The head bounding boxes are tight around the skull, including the face and hair; occluded heads are hallucinated by the annotators. The dataset is partitioned into \emph{train}, \emph{val}, \emph{test}, and \emph{leftover} sets, with rough ratio  $45\negmedspace:\negmedspace15\negmedspace:\negmedspace20\negmedspace:\negmedspace20$  percent of the annotated heads. The leftover set is not used in this paper. Up to annotation errors, neither identities nor photo albums by the same uploader are shared among these sets.

\subsubsection*{Task}

At test time, the system is given a photo and ground truth head bounding box corresponding to the test instance (probe). The task is to choose the identity of the test instance among a given set of identities (gallery set, 200$\sim$500 identities) each with $\sim$10 training samples. 

In \S\ref{sec:open-world}, we evaluate the methods when the test instance may be a background person (e.g. bystanders -- no training image given). The system is then also required to determine if the given instance is among the seen identities (gallery set).

\subsubsection*{Protocol}

We follow the PIPA protocol in \cite{Zhang2015CvprPiper} for data utilisation and model evaluation. The \emph{train} set is used for convnet feature training. The \emph{test} set contains the examples for the test identities. For each identity, the samples are divided into $\mbox{\emph{test}}_{0}$ and $\mbox{\emph{test}}_{1}$. For evaluation, we perform a two-fold cross validation by training on one of the splits and testing on the other. The \emph{val} set is likewise split into $\mbox{\emph{val}}_{0}$ and $\mbox{\emph{val}}_{1}$, and is used for exploring different models and tuning hyperparameters.

\subsubsection*{Evaluation}

We use the recognition rate (or accuracy), the rate of correct identity predictions among the test instances. For every experiment, we average two recognition rates obtained from the (training, testing) pairs ($\mbox{\emph{val}}_{0}$, $\mbox{\emph{val}}_{1}$) and ($\mbox{\emph{val}}_{1}$, $\mbox{\emph{val}}_{0}$) -- analogously for \emph{test}.

%and the validation set for exploring and optimising options. The test set is for the evaluation of our methods (table \ref{tab:test-set-accuracy}); it is itself split in two parts, $\mbox{\emph{test}}_{0}$ / $\mbox{\emph{test}}_{1}$, with roughly the same number of instances per identity. Given $\mbox{test}_{0}$ a classifier is learnt for each identity ($11$ examples per identity on average), and these are evaluated on $\mbox{test}_{1}$ (and vice-versa). Later we consider more challenging splits than the PIPA default (\S \ref{subsec:PIPA-splits}).

\subsection{\label{subsec:PIPA-splits}Splits}

\begin{figure*}
\begin{centering}
\hspace*{\fill}\includegraphics[width=1.8\columnwidth]{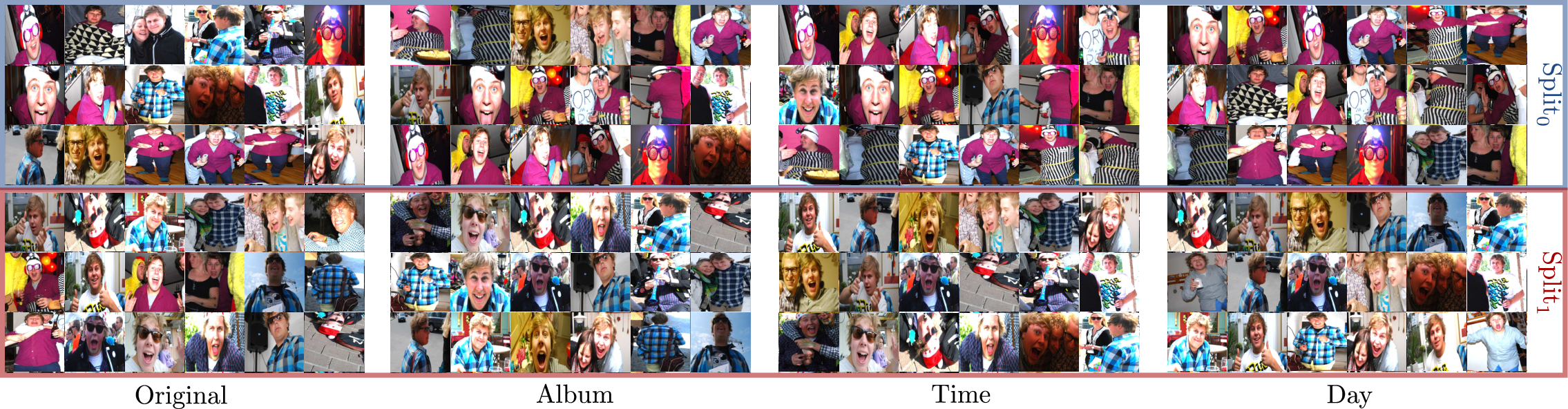}\hspace*{\fill}
\par\end{centering}
\begin{centering}
\vspace{0.5em}
\par\end{centering}
\begin{centering}
\hspace*{\fill}\includegraphics[width=1.8\columnwidth]{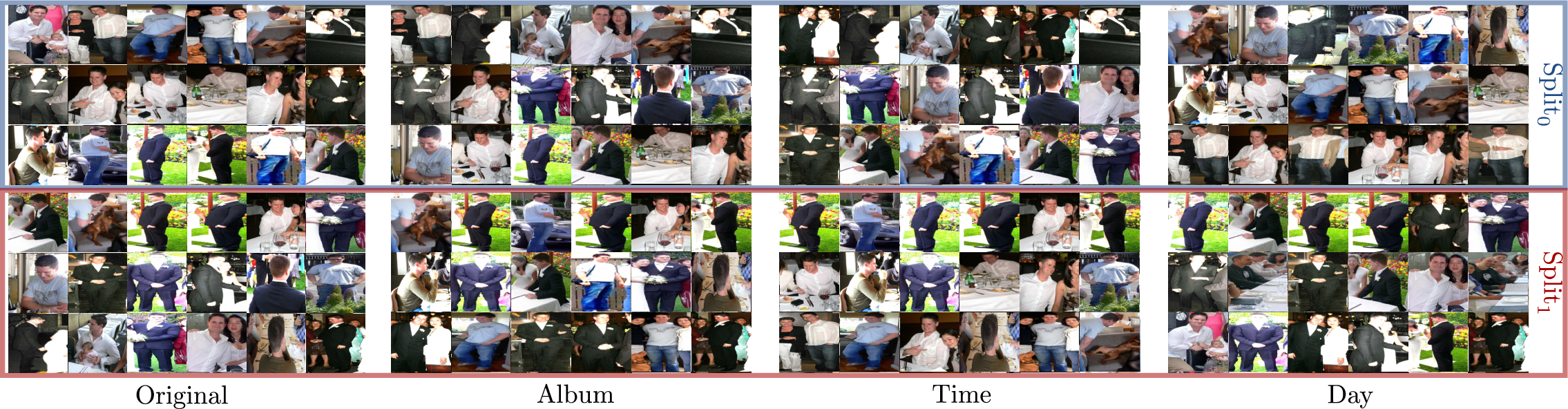}\hspace*{\fill}
\par\end{centering}
\begin{centering}
\vspace{0.5em}
\par\end{centering}
\begin{centering}
\hspace*{\fill}\includegraphics[width=1.8\columnwidth]{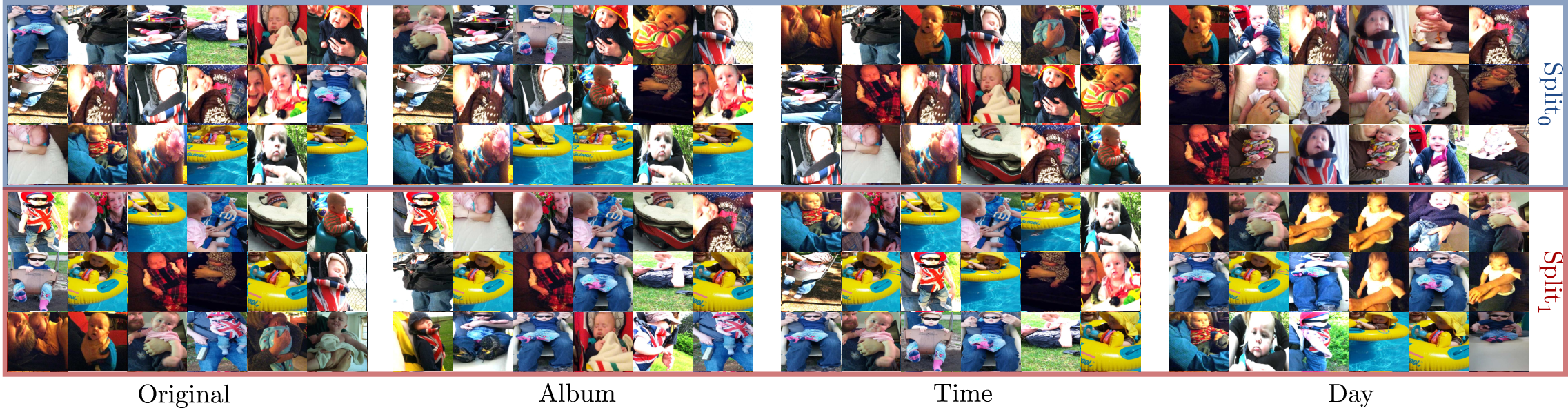}\hspace*{\fill}
\par\end{centering}
\begin{centering}
\vspace{0em}
\par\end{centering}
\caption{\label{fig:splits-visualisation}Visualisation of Original, Album,
Time and Day splits for three identities (rows 1-3). Greater appearance gap is observed from Original to Day splits.}
\end{figure*}

We consider four different ways of splitting the training and testing samples ($\mbox{\emph{val}}_{\nicefrac{0}{1}}$ and $\mbox{\emph{test}}_{\nicefrac{0}{1}}$) for each identity, aiming to evaluate different level of generalisation ability. The first one is from a prior work, and we introduce three new ones. Refer to table \ref{tab:stat-splits} for data statistics and figure \ref{fig:splits-visualisation} for visualisation.

\subsubsection*{Original split $\mathcal{O}$ \cite{Zhang2015CvprPiper}}

The Original split shares many similar examples per identity across the split -- e.g. photos taken in a row. The Original split is thus easy - even nearest neighbour on raw RGB pixels works (\S \ref{subsec:face-rgb-baseline}). In order to evaluate the ability to generalise across long-term appearance changes, we introduce three new splits below. 

\subsubsection*{Album split $\mathcal{A}$ \cite{oh2015person}}

The Album split divides training and test samples for each identity according to the photo album metadata. Each split takes the albums while trying to match the number of samples per identity as well as the total number of samples across the splits. A few albums are shared between the splits in order to match the number of samples. Since the Flickr albums are user-defined and do not always strictly cluster events and occasions, the split may not be perfect.

\subsubsection*{Time split $\mathcal{T}$ \cite{oh2015person}}

The Time split divides the samples according to the time the photo was taken. For each identity, the samples are sorted according to their ``photo-taken-date'' metadata, and then divided according to the newest versus oldest basis. The instances without time metadata are distributed evenly. This split evaluates the temporal generalisation of the recogniser. However, the ``photo-taken-date'' metadata is very noisy with lots of missing data. 

\subsubsection*{Day split $\mathcal{D}$ \cite{oh2015person}}

The Day split divides the instances via visual inspection to ensure the firm ``appearance change'' across the splits. We define two criteria for division: (1) a firm evidence of date change such as \{change of season, continent, event, co-occurring people\} and/or (2) visible changes in \{hairstyle, make-up, head or body wear\}. We discard identities for whom such a division is not possible. After division, for each identity we randomly discard samples from the larger split until the sizes match. If the smaller split has $\leq\negmedspace 4$ instances, we discard the identity altogether. The Day split enables clean experiments for evaluating the generalisation performance across strong appearance and event changes.

\subsection{\label{subsec:face-detection}Face detection}

Instances in PIPA are annotated by humans around their heads (tight around skull). We additionally compute face detections over PIPA for three purposes: (1) to compare the amount of identity information in head versus face (\S\ref{sec:Cues-for-recognition}), (2) to obtain head orientation information for further analysis (\S\ref{sec:challenges-analysis}), and (3) to simulate the scenario without ground truth head box at test time (\S\ref{sec:open-world}). We use the open source DPM face detector \cite{Mathias2014Eccv}.

Given a set of detected faces (above certain detection score threshold) and the ground truth heads, the match is made according to the overlap (intersection over union). For matched heads, the corresponding face detections tell us which DPM component is fired, thereby allowing us to infer the head orientation (frontal or side view). See Appendix \S{\color{red}A} for further details.

\begin{figure}
\begin{centering}
\includegraphics[width=0.8\columnwidth]{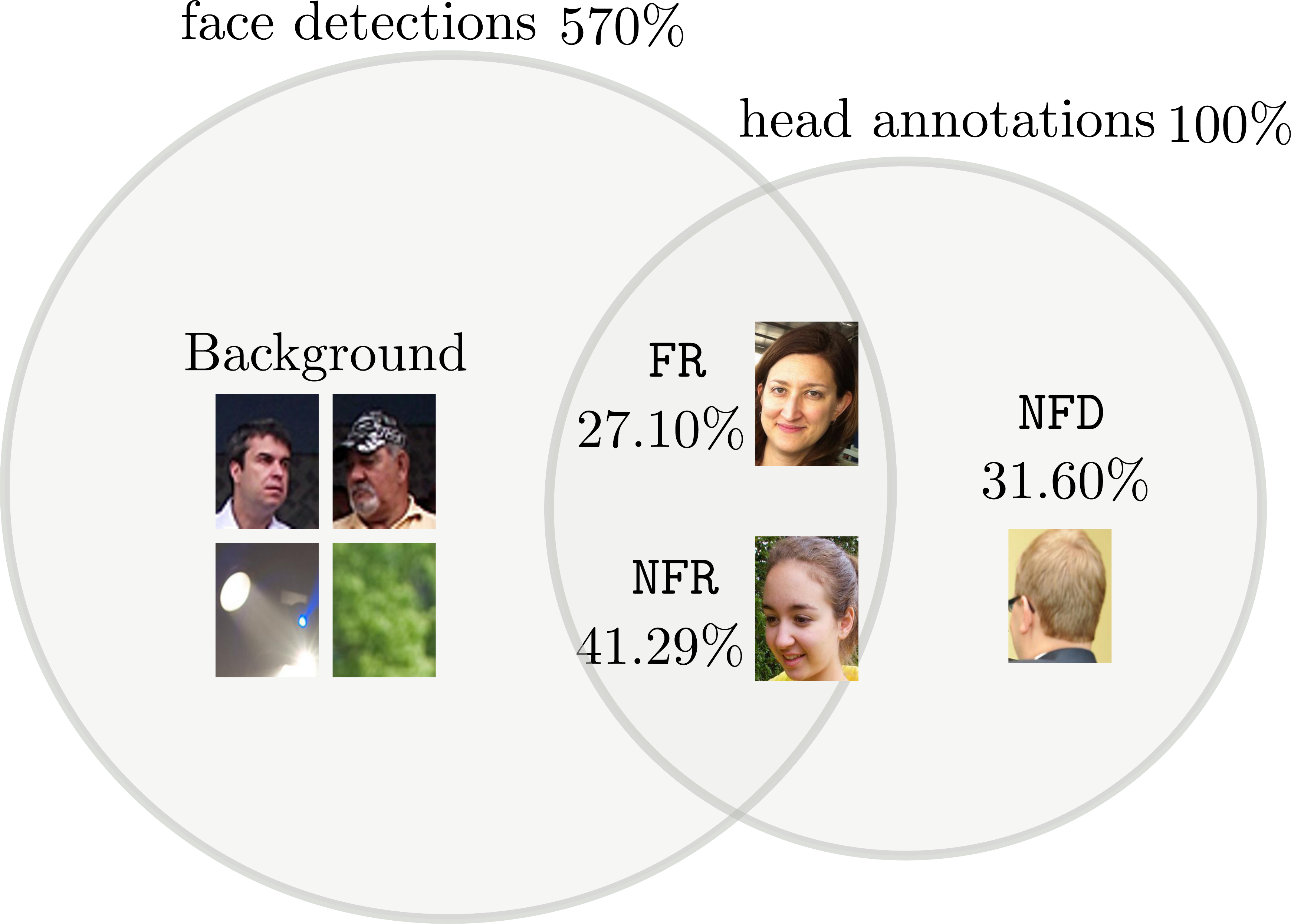}\vspace{-0.5em}
\par\end{centering}
\caption{\label{fig:threeway-diagram}Face detections and head annotations in PIPA. The matches are determined by overlap (intersection over union). For matched faces (heads), the detector DPM component gives the orientation information (frontal versus non-frontal).}
\end{figure}

Using the DPM component, we partition instances in PIPA as follows: (1) detected and frontal ($\texttt{FR}$, 41.29\%), (2) detected and non-frontal ($\texttt{NFR}$, 27.10\%), and (3) no face detected ($\texttt{NFD}$, 31.60\%). We denote detections without matching ground truth head as Background. See figure \ref{fig:threeway-diagram} for visualisation.

%To validate the face detection location, we define $\texttt{h}_{\texttt{det}}$ as the head region regressed from the face detections $\ensuremath{\texttt{f}}$ (per DPM component). The head detection cue $\texttt{h}_{\texttt{det}}$ is compared against the cue based on ground truth head locations $\texttt{h}$. When using $\texttt{h}_{\texttt{det}}$ instead of the ground truth head ($\ensuremath{\texttt{h}}$ in table \ref{tab:validation-set-regions-accuracy}), results drop only $0.45$pp (from $83.88\%$ to $83.43\%$) thus indirectly validating that faces are well localized.

\begin{table}
\begin{centering}
\setlength\tabcolsep{0.5em}
\begin{tabular}{cccccccccccc}
& &  & \multicolumn{4}{c}{\emph{val}} & & \multicolumn{4}{c}{\emph{test}} \tabularnewline
\vspace{-1em}
 &  &  &  &  &  &  & \tabularnewline
\cline{4-7} \cline{9-12} 
\vspace{-1em}
 &  &  &  &  &  &  & \tabularnewline
& & & $\mathcal{O}$ & $\mathcal{A}$ & $\mathcal{T}$ & $\mathcal{D}$ & & $\mathcal{O}$ & $\mathcal{A}$ & $\mathcal{T}$ & $\mathcal{D}$ \tabularnewline
\vspace{-1em}
 &  &  &  &  &  &  & \tabularnewline
\cline{1-2} \cline{4-7} \cline{9-12} 
\vspace{-1em}
 &  &  &  &  &  &  & \tabularnewline
\cline{1-2} \cline{4-7} \cline{9-12} 
\vspace{-0.5em}
 &  &  &  &  &  &  & \tabularnewline
\multirow{2}{*}{{\rotatebox{90}{{spl.0}\hspace{0em}}}}&instance &  & 4820 & 4859 & 4818 & 1076 &  & 6443 & 6497 & 6441 & 2484  \tabularnewline
&identity &  & 366 & 366 & 366 & 65 &  & 581 & 581 & 581 & 199 \tabularnewline
\vspace{-1em}
 &  &  &  &  &  &  &\tabularnewline
\cline{1-2} \cline{4-7} \cline{9-12} 
\vspace{-0.5em}
 &  &  &  &  &  &  & \tabularnewline
\multirow{2}{*}{\rotatebox{90}{spl.1\hspace{0em}}}&instance &  & 4820 & 4783 & 4824 & 1076 &  & 6443 & 6389 & 6445 & 2485  \tabularnewline
&identity &  & 366 & 366 & 366 & 65 &  & 581 & 581 & 581 & 199  \tabularnewline
\vspace{-1em}
 &  &  &  &  &  &  &\tabularnewline
\cline{1-2} \cline{4-7} \cline{9-12} 
\end{tabular}
\par\end{centering}
\vspace{1em}

\caption{\label{tab:stat-splits}Split statistics for \emph{val} and \emph{test} sets. Total number of instances and identites for each split is shown.}
\end{table}

%In the next section, various image regions and the corresponding recognition cues are defined (\S \ref{subsec:Body-regions}), and their validation set performances are compared (\S \ref{sec:how-informative-body-region} to \S \ref{sec:Attributes}). The performance of our final system and comparisons to other methods and baselines are provided in \S \ref{sec:Test-set-results}. \S\ref{sec:failures-analysis} will present an in-depth analysis of the systems, including the performance on the more realistic and challenging PIPA splits. In \S \ref{sec:Towards-open-world}, we report results in the open world setting where ground truth heads are not assumed at test time.

\section{\label{sec:Cues-for-recognition}Cues for recognition}

In this section, we investigate the cues for recognising people in social media photos. We begin with an overview of our model. Then, we experimentally answer the following questions: how informative are fixed body regions (no pose estimation) (\S\ref{sec:how-informative-body-region})? How much does scene context help (\S\textcolor{red}{\ref{sec:Scene}})? Is it head or face (head minus hair and background) that is more informative (\S\ref{sec:Head-or-face})? And how much do we gain by using extended data (\S{\ref{sec:Additional-training-data}} \& \S{\ref{sec:Attributes}})? How effective is a specialised face recogniser (\S{\ref{sec:deepid}})? Studies in this section are based exclusively on the \emph{val} set.

\subsection{Model overview}

\begin{wrapfigure}{O}{0.45\columnwidth}%
\begin{centering}
\par\end{centering}
\centering{}
\includegraphics[height=0.6\columnwidth]{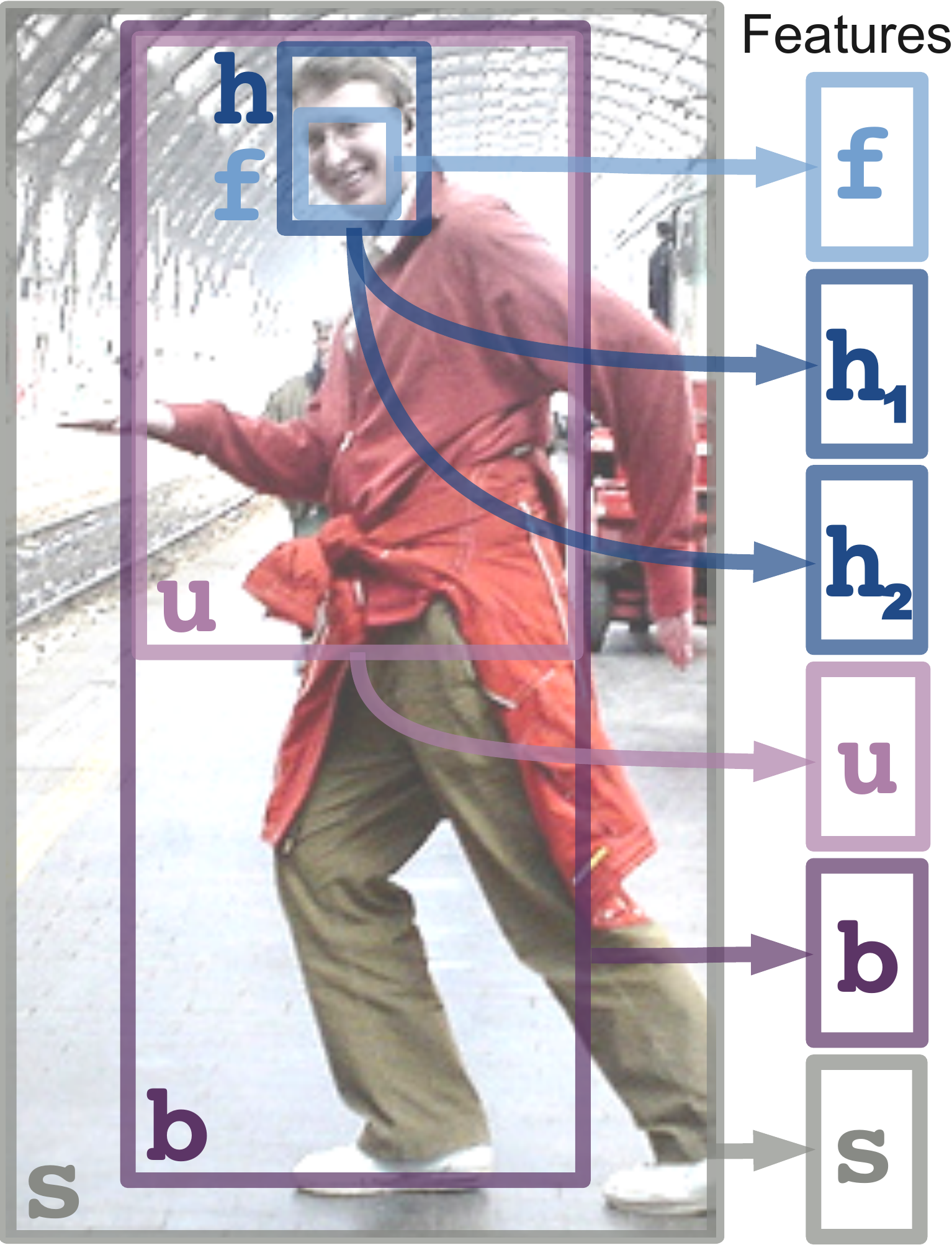}
\caption{\label{fig:image-regions}Regions considered for feature extraction: face $\ensuremath{\texttt{f}}$, head $\ensuremath{\texttt{h}}$, upper body $\ensuremath{\texttt{u}}$, full body $\ensuremath{\texttt{b}}$, and scene $\ensuremath{\texttt{s}}$. More than one cue can be extracted per region (e.g. $\ensuremath{\texttt{h}}_{1}$, $\ensuremath{\texttt{h}}_{2}$ ). %For region definitions and performances, see text.
}
\end{wrapfigure}%

At test time, given a ground truth head bounding box, we estimate five different regions depicted in figure \ref{fig:image-regions}. Each region is fed into one or more convnets to obtain a set of cues. The cues are concatenated to form a feature vector describing the instance. Throughout the paper we write $+$ to denote vector concatenation. Linear SVM classifiers are trained over this feature vector (one versus the rest). In our final system, except for DeepID2+ \cite{Sun2014ArxivDeepId2plus}, all features are computed using the seventh layer (fc7) of AlexNet \cite{Krizhevsky2012Nips} pre-trained for ImageNet classification. The cues only differ amongst each other on the image area and the fine-tuning used (type of data or surrogate task) to alter the AlexNet, except for the DeepID2+ \cite{Sun2014ArxivDeepId2plus} feature.% -- a siamese network is tuned.

\subsection{\label{subsec:Body-regions}Image regions used}

We choose five different image regions based on the ground truth head
annotation (given at test time, see the protocol in \S\ref{sec:PIPA-dataset}). The
head rectangle $\ensuremath{\texttt{h}}$ corresponds to the ground
truth annotation. The full body rectangle $\ensuremath{\texttt{b}}$
is defined as $\left(3\negthinspace\times\negthinspace\mbox{head width},\right.$
$\left.6\negthinspace\times\negthinspace\mbox{head height}\right)$,
with the head at the top centre of the full body. The upper body rectangle
$\ensuremath{\texttt{u}}$ is the upper-half of $\ensuremath{\texttt{b}}$.
The scene region $\ensuremath{\texttt{s}}$ is the whole image containing
the head. 

The face region $\ensuremath{\texttt{f}}$ is obtained using the DPM face detector discussed in \S\ref{subsec:face-detection}. For head boxes with no matching detection (e.g. back views and occluded faces), we regress the face area from the head using the face-head displacement statistics on the \emph{train} set. Five respective image regions are illustrated in figure \ref{fig:image-regions}.

Note that the regions overlap with each other, and that depending
on the person's pose they might be completely off. For example,
$\ensuremath{\texttt{b}}$ for a lying person is likely to contain
more background than the actual body. While precise body parts obtained via pose estimation~\cite{EldarPose,cao2017realtime} may contribute to even better performances~\cite{liu_2017_coco}, we choose not to use it for the sake of efficiency. Our simple region selection scheme still leads to the state of the art performances, even compared to methods that do rely on pose estimation (\S\ref{sec:Test-set-results}).

\subsection{\label{subsec:Implementation}Fine-tuning and parameters}

\begin{figure}
\begin{centering}
\includegraphics[width=0.7\columnwidth]{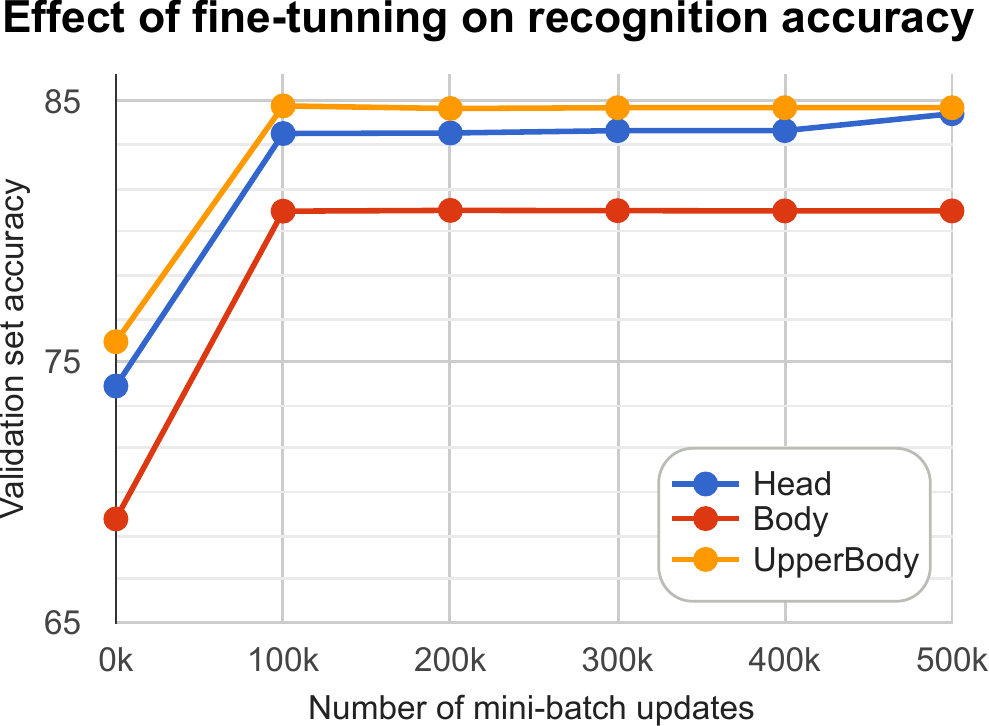}\vspace{-0.5em}
\par\end{centering}
\caption{\label{fig:fine-tuning-effect}PIPA \emph{val} set performance of different
cues versus the SGD iterations in fine-tuning.}
\end{figure}

Unless specified otherwise AlexNet is fine-tuned using the PIPA \emph{train} set ($\sim\negmedspace30\mbox{k}$ instances, $\sim\negmedspace1.5\mbox{k}$ identities), cropped at five different image regions, with $300\mbox{k}$ mini-batch iterations (batch size $50$). We refer to the base cue thus obtained as $\ensuremath{\texttt{f}}$, $\ensuremath{\texttt{h}}$, $\ensuremath{\texttt{u}}$, $\ensuremath{\texttt{b}}$, or $\ensuremath{\texttt{s}}$, depending on the cropped region. On the \emph{val} set we found the fine-tuning to provide a systematic $\sim\negmedspace10$ percent points (pp) gain over the non-fine-tuned AlexNet (figure \ref{fig:fine-tuning-effect}). We use the seventh layer (fc7) of AlexNet for each cue ($4\,096$ dimensions).

We train for each identity a one-versus-all SVM classifier with the regularisation parameter $C=1$; it turned out to be an insensitive parameter in our preliminary experiments. As an alternative, the naive nearest neighbour classifier has also been considered. However, on the \emph{val} set the SVMs consistently outperforms the NNs by a $\sim\negmedspace10\ \mbox{pp}$ margin.

\subsection{\label{sec:how-informative-body-region}How informative is each image
region?}

\begin{table}
\begin{centering}
\par\end{centering}
\begin{centering}
\begin{tabular}{lllcc}
&& Cue & & Accuracy\tabularnewline
\cline{1-1} \cline{3-3} \cline{5-5}
& \vspace{-0.9em}\tabularnewline
\cline{1-1} \cline{3-3} \cline{5-5}
{Chance level} &&&& \hspace{0.5em}1.04\tabularnewline
\cline{1-1} \cline{3-3} \cline{5-5}
Scene (\S\ref{sec:Scene}) && \texttt{$\texttt{s}$} && 27.06\tabularnewline
Body  && $\mbox{\ensuremath{\texttt{b}}}$ && 80.81\tabularnewline
Upper body && $\texttt{u}$ && 84.76\tabularnewline
Head && $\texttt{h}$ && 83.88\tabularnewline
Face (\S\ref{sec:Head-or-face}) && $\texttt{f}$ && 74.45\tabularnewline
\cline{1-1} \cline{3-3} \cline{5-5}
Zoom out && $\texttt{f}$ && 74.45\tabularnewline
 && $\texttt{f}\negthinspace+\negthinspace\texttt{h}$ && 84.80\tabularnewline
 && $\texttt{f}\negthinspace+\negthinspace\texttt{h}\negthinspace+\negthinspace\texttt{u}$ && 90.65\tabularnewline
 & & $\texttt{f}\negthinspace+\negthinspace\texttt{h}\negthinspace+\negthinspace\texttt{u}\negthinspace+\negthinspace\texttt{b}$ && 91.14\tabularnewline
 & & $\texttt{f}\negthinspace+\negthinspace\texttt{h}\negthinspace+\negthinspace\texttt{u}\negthinspace+\negthinspace\texttt{b}\negthinspace+\negthinspace\texttt{s}$ && 91.16\tabularnewline
\cline{1-1} \cline{3-3} \cline{5-5}
Zoom in && \texttt{$\texttt{s}$} && 27.06\tabularnewline
 && $\texttt{s}\negthinspace+\negthinspace\texttt{b}$ && 82.16\tabularnewline
 && $\texttt{s}\negthinspace+\negthinspace\texttt{b}\negthinspace+\negthinspace\texttt{u}$ && 86.39\tabularnewline
 & & $\texttt{s}\negthinspace+\negthinspace\texttt{b}\negthinspace+\negthinspace\texttt{u}\negthinspace+\negthinspace\texttt{h}$ && 90.40\tabularnewline
 && $\texttt{s}\negthinspace+\negthinspace\texttt{b}\negthinspace+\negthinspace\texttt{u}\negthinspace+\negthinspace\texttt{h}\negthinspace+\negthinspace\texttt{f}$ && 91.16\tabularnewline
\cline{1-1} \cline{3-3} \cline{5-5}
Head+body && $\texttt{h}\negthinspace+\negthinspace\texttt{b}$ && 89.42\tabularnewline
%Face+head && $\texttt{f}\negthinspace+\negthinspace\texttt{h}$ && 84.80\tabularnewline
% && $\texttt{f}\negthinspace+\negthinspace\texttt{h}\negthinspace+\negthinspace\texttt{u}$ && 90.65\tabularnewline
% && $\texttt{f}\negthinspace+\negthinspace\texttt{h}\negthinspace+\negthinspace\texttt{b}$ && 90.19\tabularnewline
Full person && $\texttt{P}=\texttt{f}\negthinspace+\negthinspace\texttt{h}\negthinspace+\negthinspace\texttt{u}\negthinspace+\negthinspace\texttt{b}$\hspace*{-1.5em} && 91.14\tabularnewline
Full image && $\texttt{\ensuremath{\mbox{\ensuremath{\texttt{P}}}_{s}}}=\texttt{P}\negthinspace+\negthinspace\texttt{s}$ && 91.16\tabularnewline
\cline{1-1} \cline{3-3} \cline{5-5}
\end{tabular}
\par\end{centering}
\vspace{0.8em}

\caption{\label{tab:validation-set-regions-accuracy}PIPA \emph{val} set accuracy of cues based on different image regions and their concatenations ($+$ means concatenation).}
\end{table}

Table \ref{tab:validation-set-regions-accuracy} shows the \emph{val} set results of each region individually and in combination. Head $\ensuremath{\texttt{h}}$ and upper body $\ensuremath{\texttt{u}}$ are the strongest individual cues. Upper body is more reliable than the full body $\ensuremath{\texttt{b}}$ because the lower body is commonly occluded or cut out of the frame, and thus is usually a distractor. Scene $\ensuremath{\texttt{s}}$ is, unsurprisingly, the weakest individual cue, but it still useful information for person recognition (far above chance level). Importantly, we see that all cues complement each other, despite overlapping pixels. Overall, our features and combination strategy are effective.

\subsection{\label{sec:how-to-select-regions-otherwise}Empirical justification for the regions $\texttt{fhubs}$}

\begin{figure}
\centering
\footnotesize
\setlength\tabcolsep{0.15em}
\begin{tabular}{ccccc}
{\rotatebox{90}{\hspace{2.5em}Head ($\texttt{h}$) size \hspace{0.0em}}}
&\includegraphics[width=0.23\columnwidth]{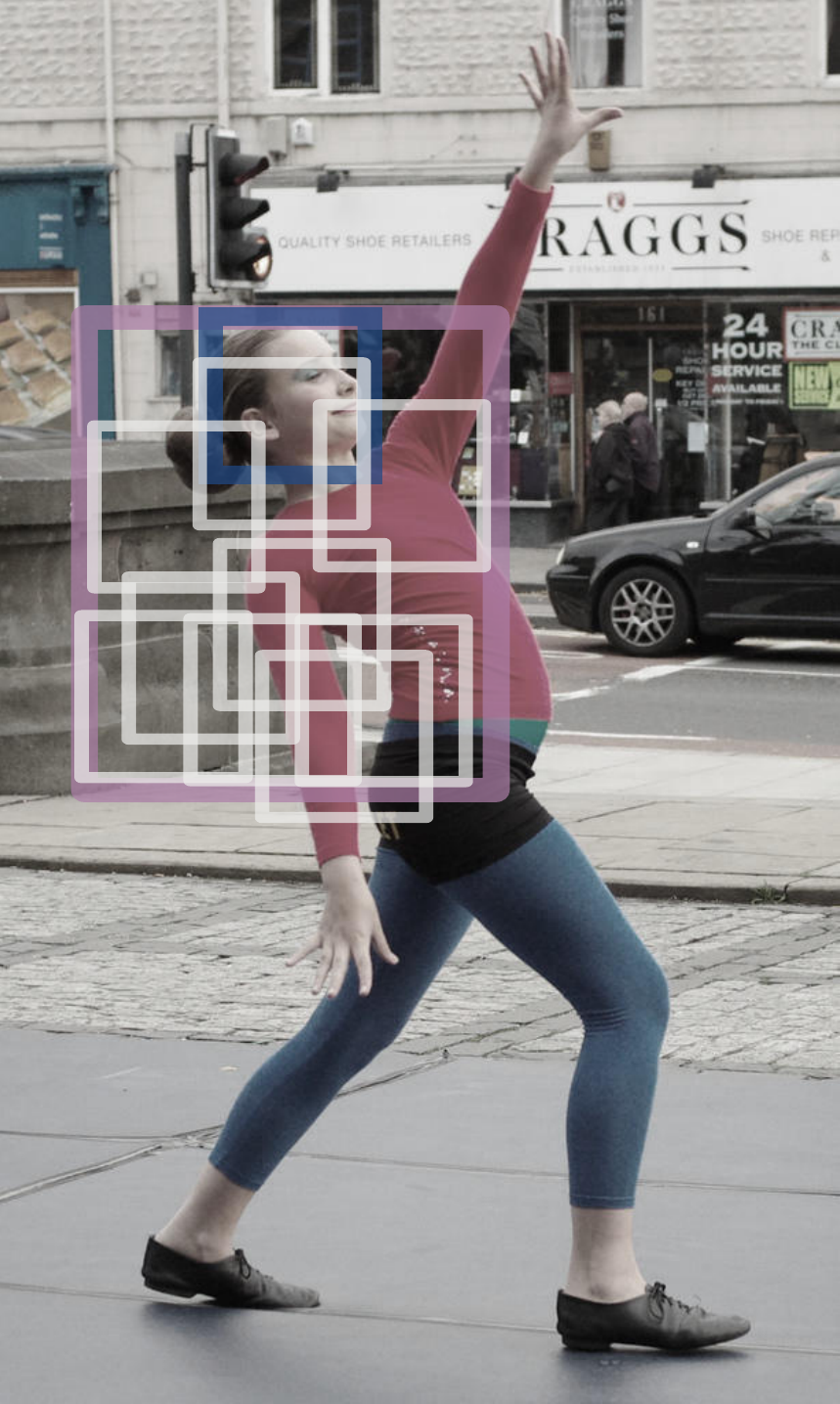} &\includegraphics[width=0.23\columnwidth]{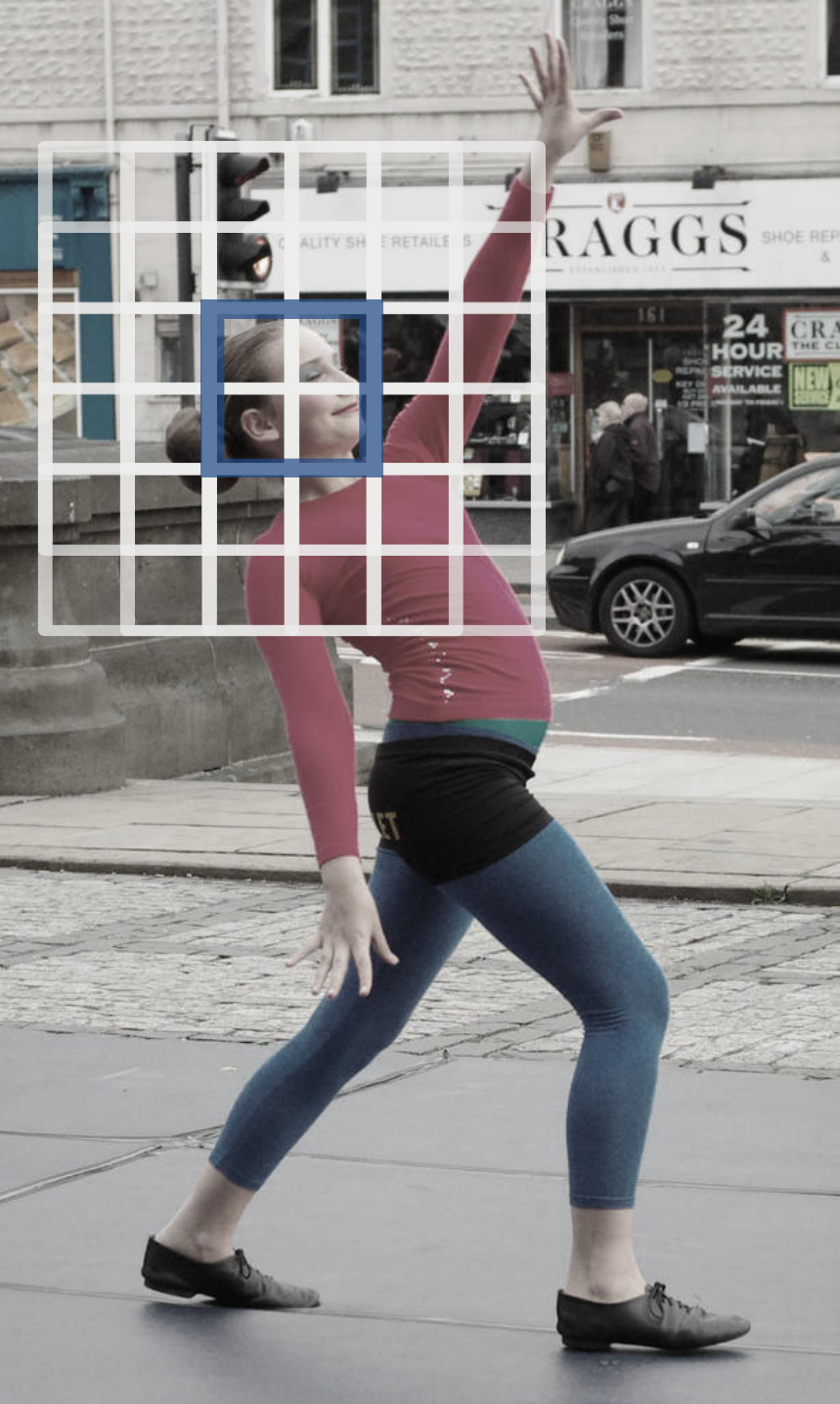} &\includegraphics[width=0.23\columnwidth]{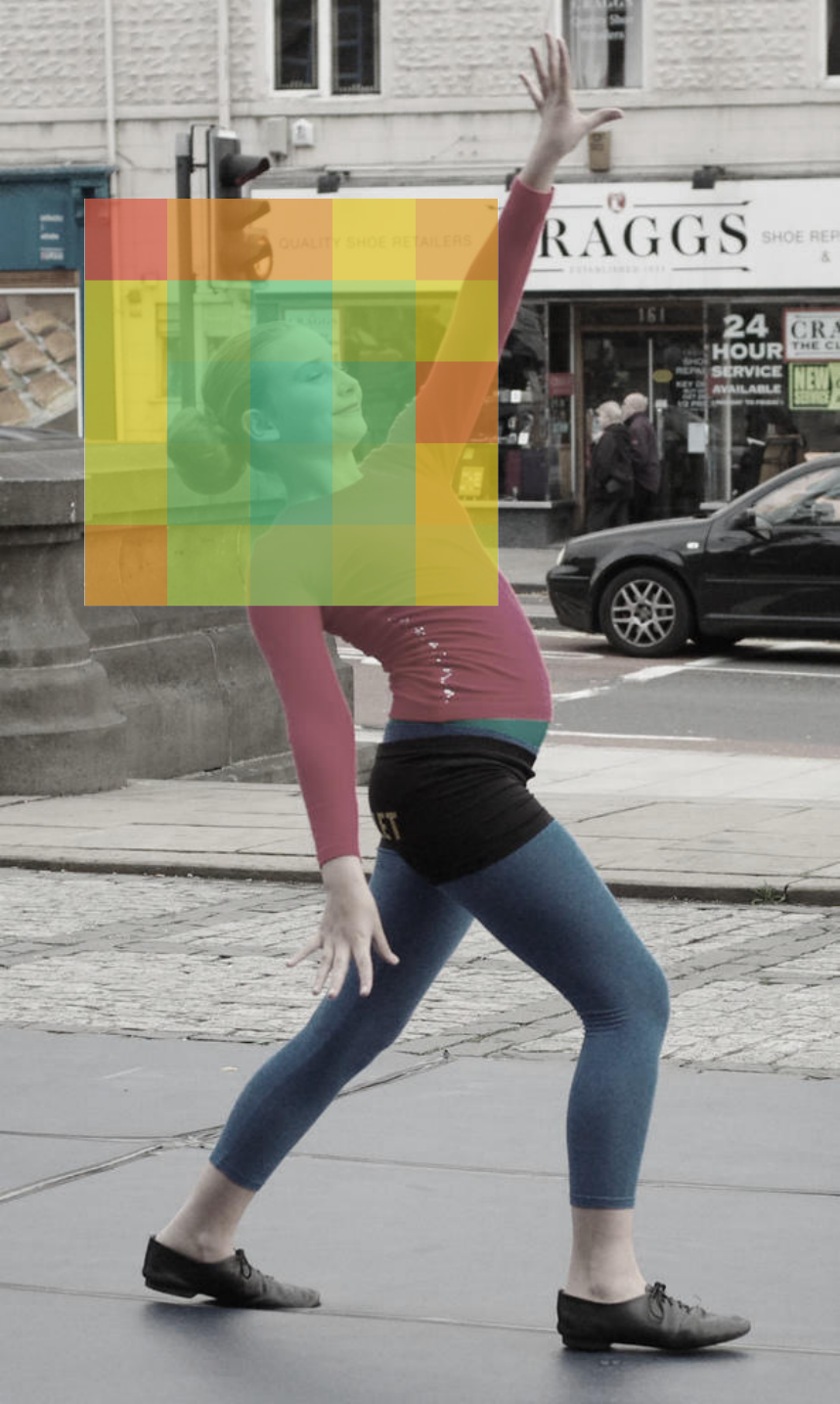} &\includegraphics[width=0.23\columnwidth]{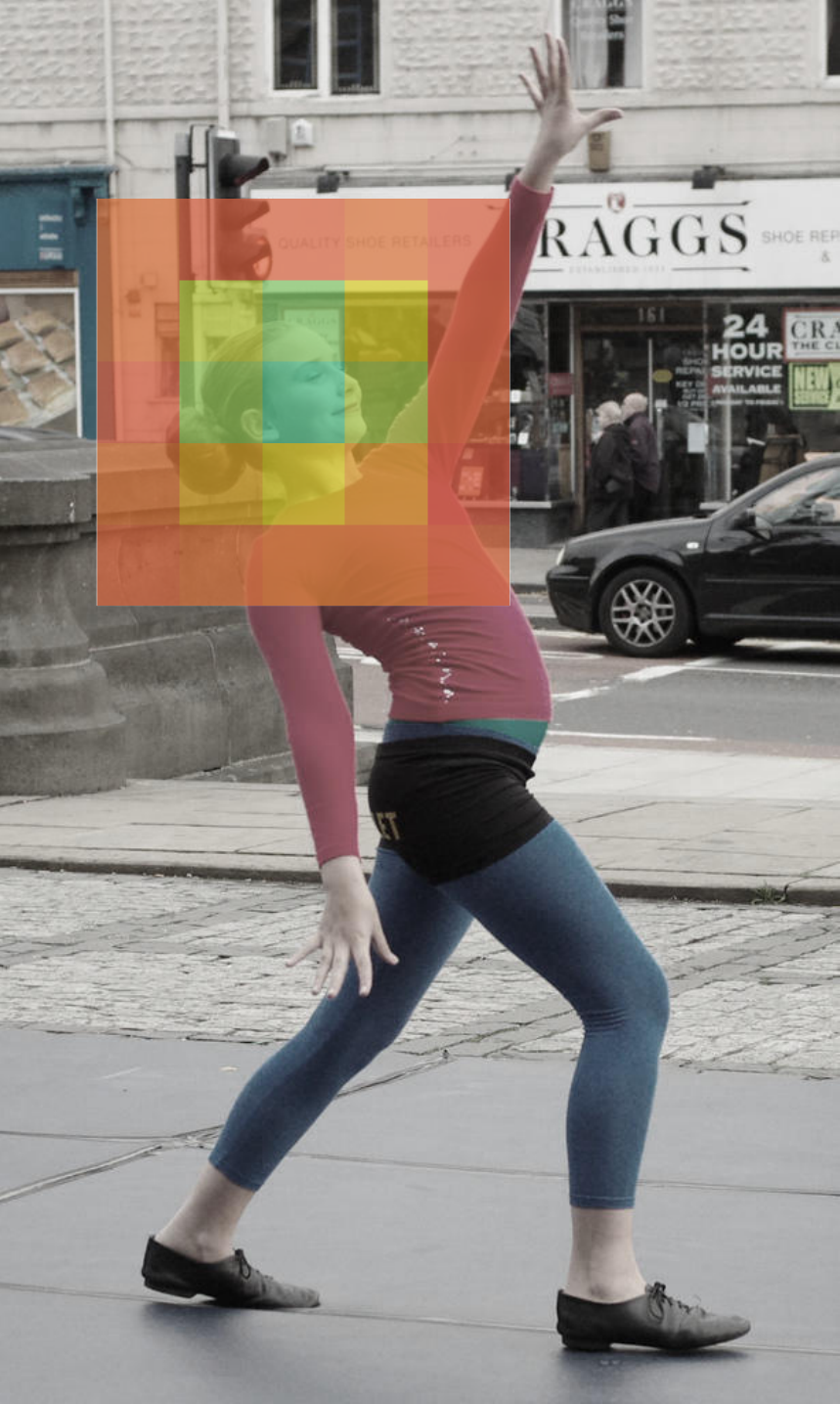} \\
&&&\includegraphics[width=0.2\columnwidth]{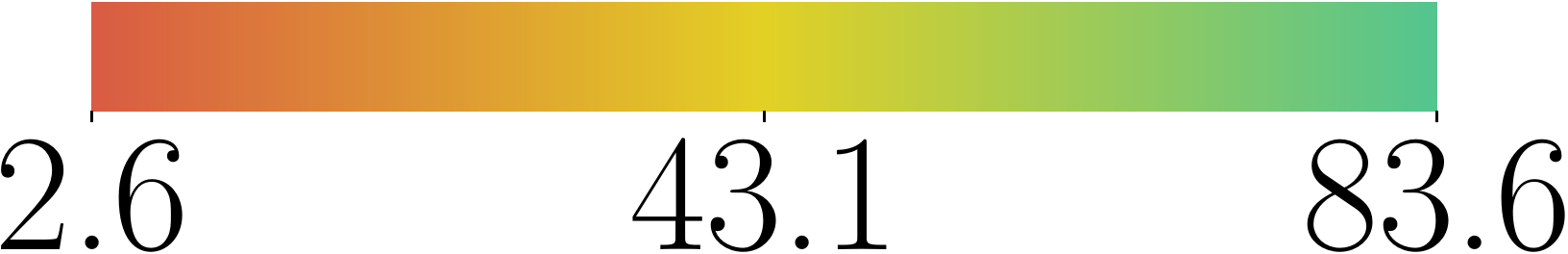} 
&\includegraphics[width=0.2\columnwidth]{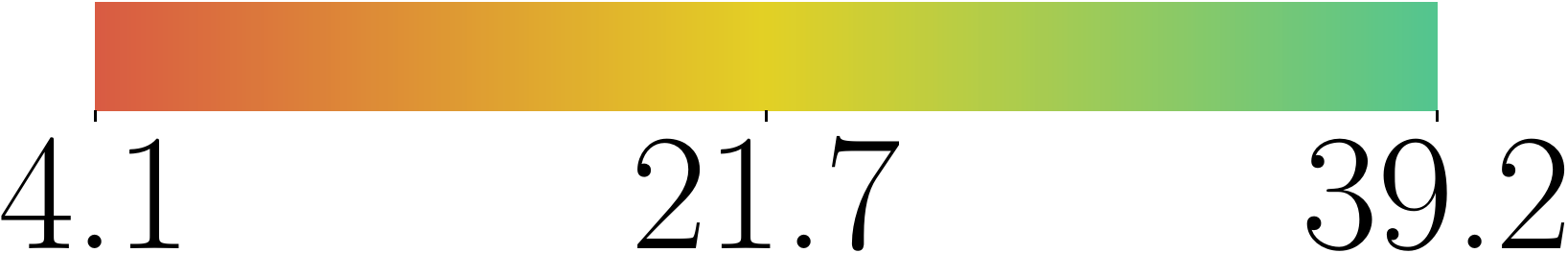} \\
{\rotatebox{90}{\hspace{1.5em}Upper body ($\texttt{u}$) size \hspace{0.0em}}}
&\includegraphics[width=0.23\columnwidth]{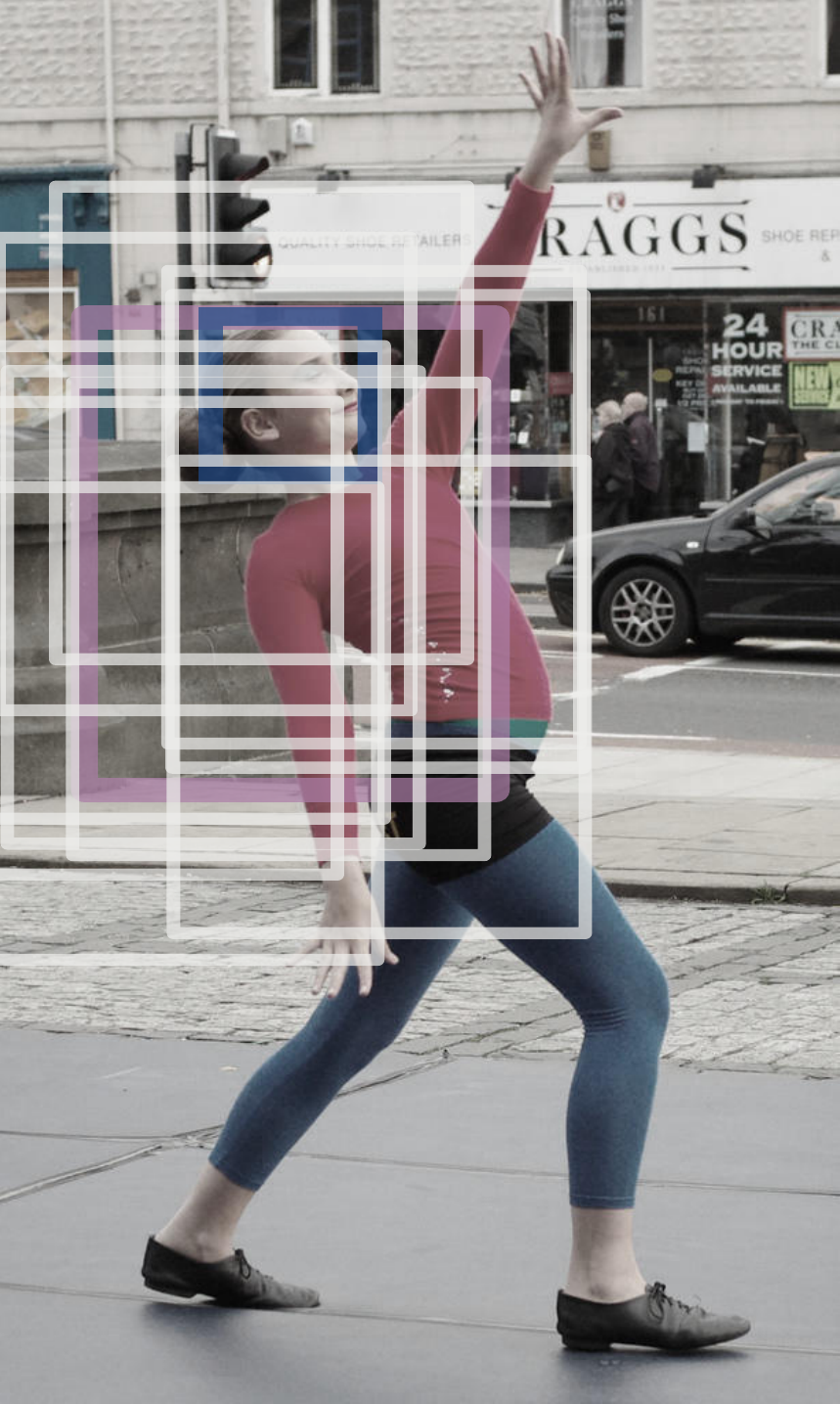} &\includegraphics[width=0.23\columnwidth]{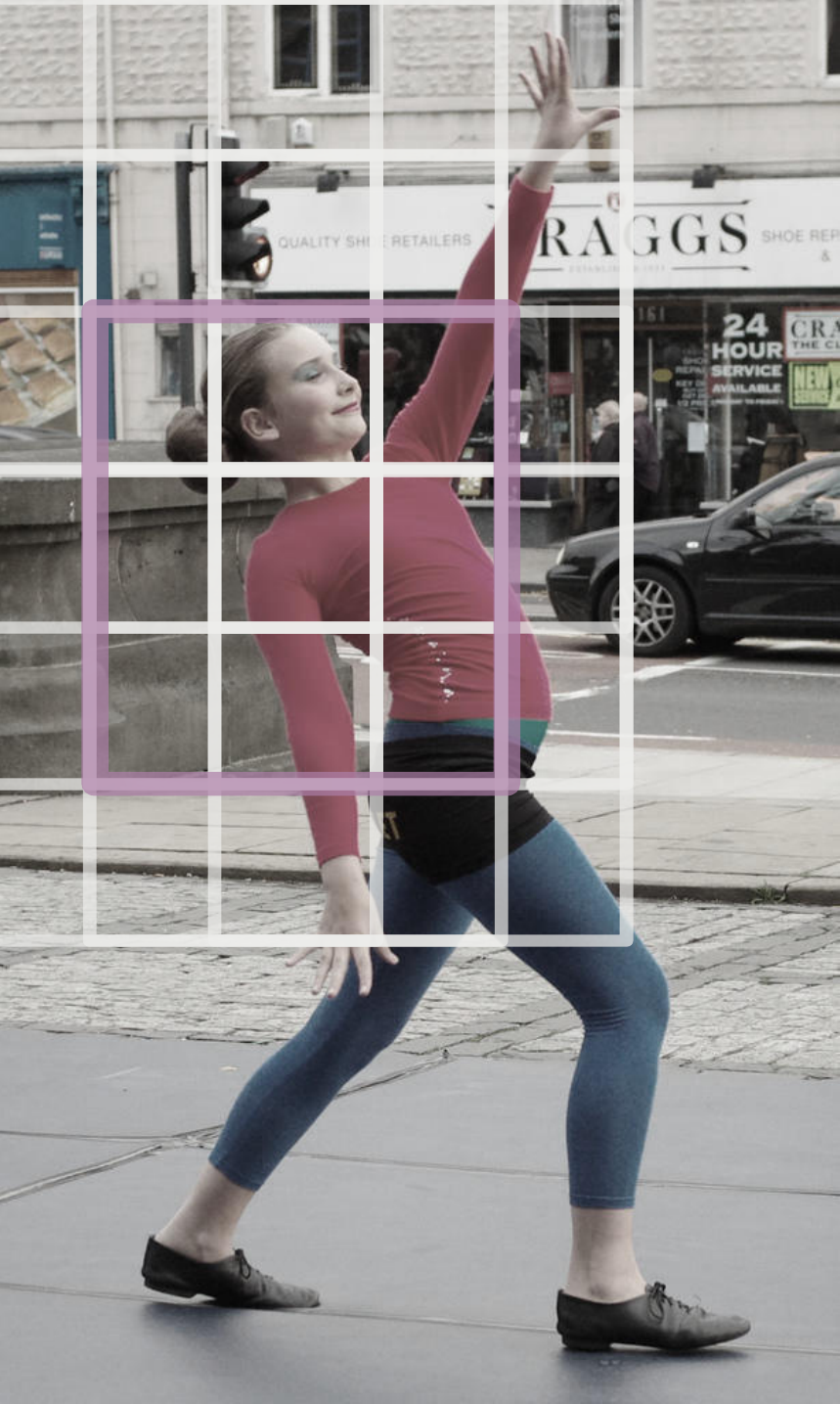} &\includegraphics[width=0.23\columnwidth]{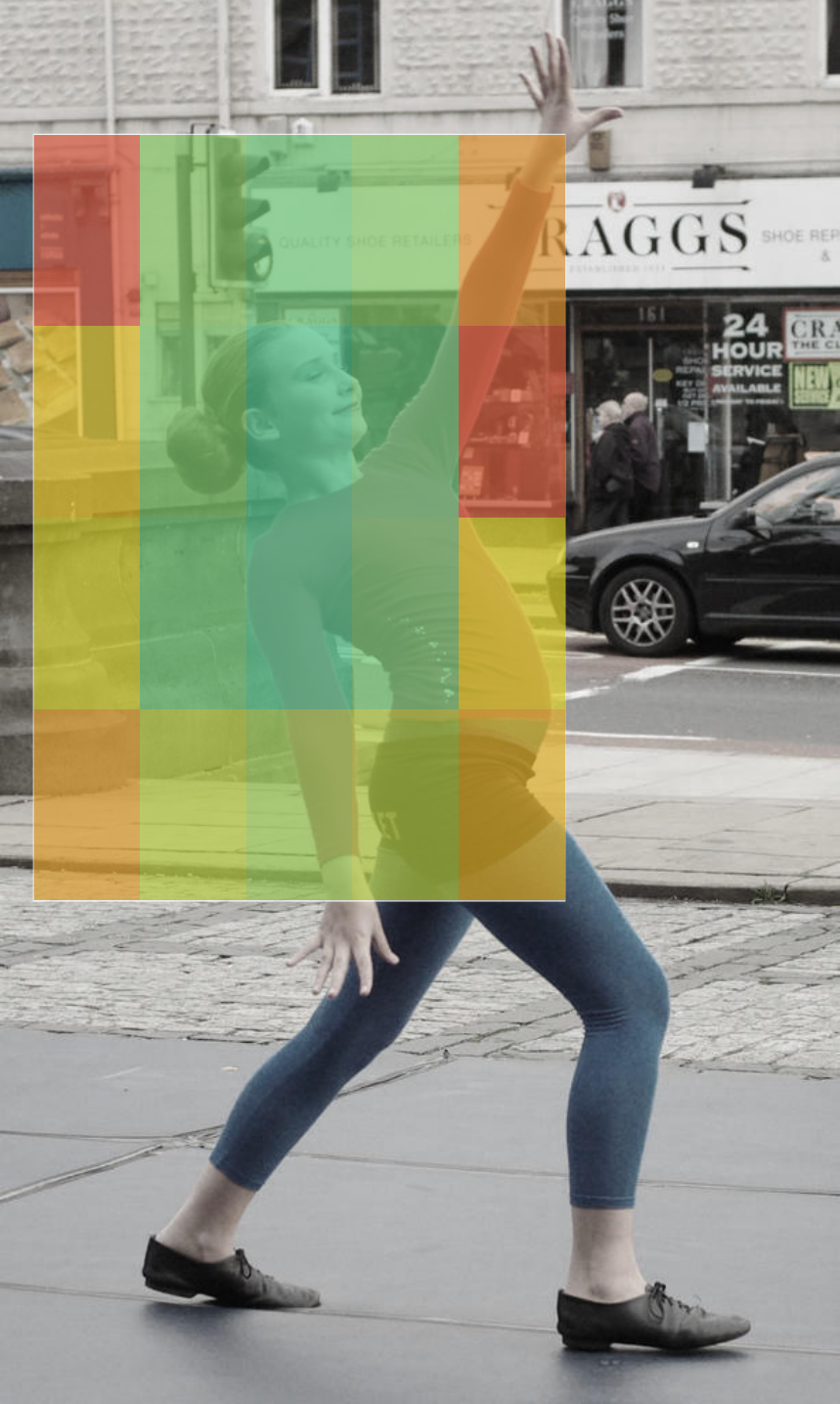} &\includegraphics[width=0.23\columnwidth]{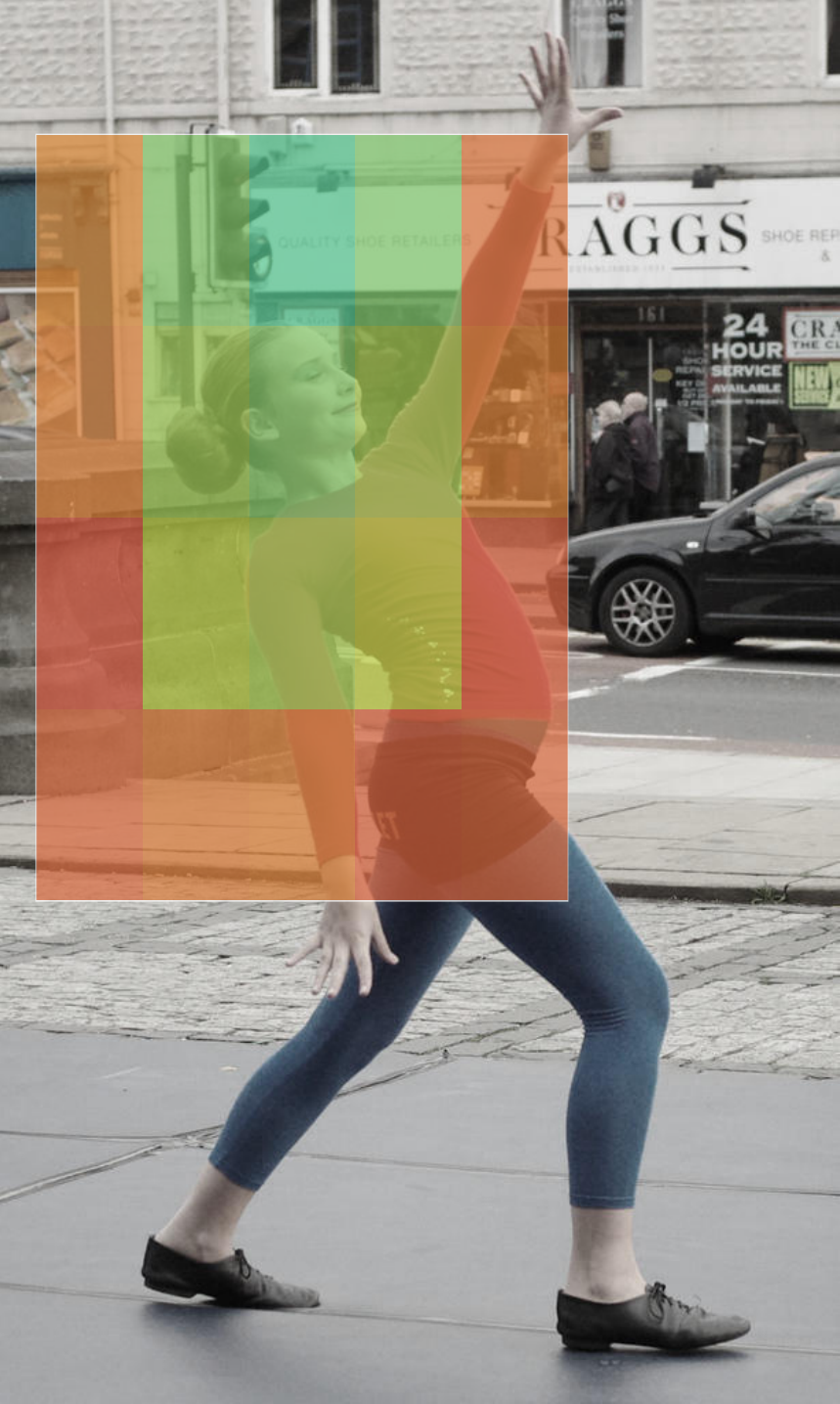} \\
&&&\includegraphics[width=0.2\columnwidth]{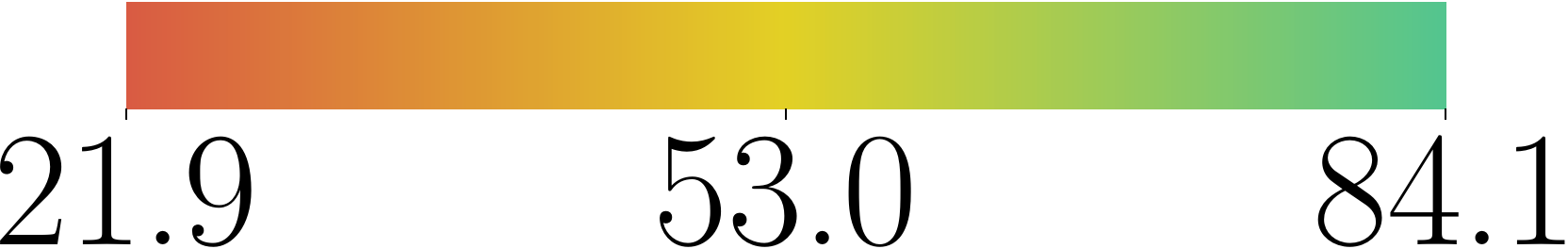} &\includegraphics[width=0.2\columnwidth]{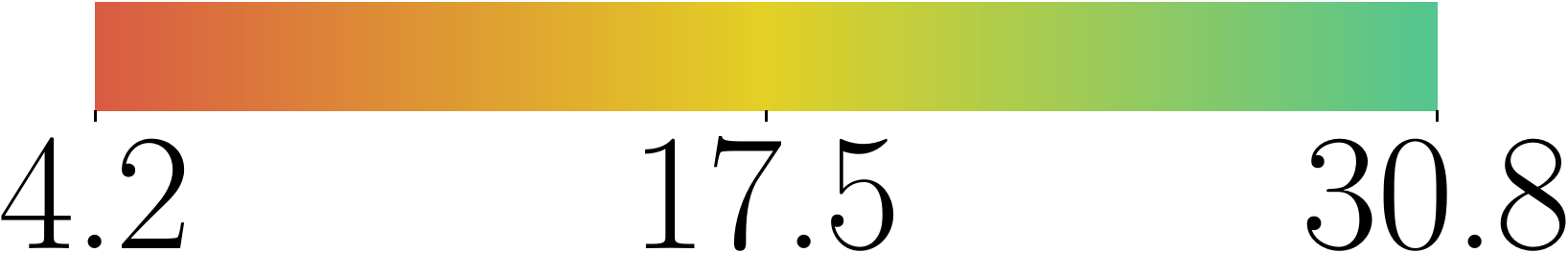} \\
& r-patch & sw & Original split & Day split \\ 
&&&\multicolumn{2}{c}{sw results}

\end{tabular}
\vspace{0em}
\caption{\label{fig:patch-analysis-visualisation}Regions considered for analysis. We consider cues from head ($\texttt{h}$) as well as upper body ($\texttt{u}$) sized patches that are either chosen randomly (r-patch, column 1) or in sliding window manner (sw, column 2). Recognition results for sw are visualised in Original (column 3) and Day (column 4) splits. }
\end{figure}

In order to further justify the choice of the five image regions ($\texttt{f}\texttt{h}\texttt{u}\texttt{b}\texttt{s}$), we compare their informativeness against three  baseline types of cues: (1) \emph{random patch} (r-patch), (2) \emph{sliding window} (sw), and (3) \emph{random initialisation} (r-init). For each type, we consider head sized ($\texttt{h}$, ground truth head size) and upper body sized ($\texttt{u}$, $3\times3$ of head size) regions. See figure \ref{fig:patch-analysis-visualisation} columns 1 and 2 for r-patch and sw.

Specifically, (1) for head sized r-patch we sample regions from within the original upper body region ($\texttt{u}$); for upper body sized r-patch, we sample from within $\pm 1$ head away from the original upper body region. (2) For head sized sw, we set the stride as half of $\texttt{h}$ width/height, while for upper body sized ones, we set the stride as $\texttt{h}$ width/height themselves. The r-patch and sw are fixed across person instances with respect to the respective head locations. (3) The r-init are always based on the original head and upper body regions, but the features are trained with different random initialisations. 

The results for the sliding window regions are shown in figure \ref{fig:patch-analysis-visualisation} columns 3 and 4, under Original and Day splits, respectively. In all sizes and domain gaps, the original head region is the most informative one. The informativeness of head region is amplified under the Day split (larger domain gap), with larger performance gap between head and context regions -- clothing and event changes in the context regions hamper identification. \S\ref{sec:challenges-analysis} contains more in-depth analysis regarding this point.

\begin{figure*}
\centering
\hfill
\subfloat[Original]{
\includegraphics[width=0.8\columnwidth]{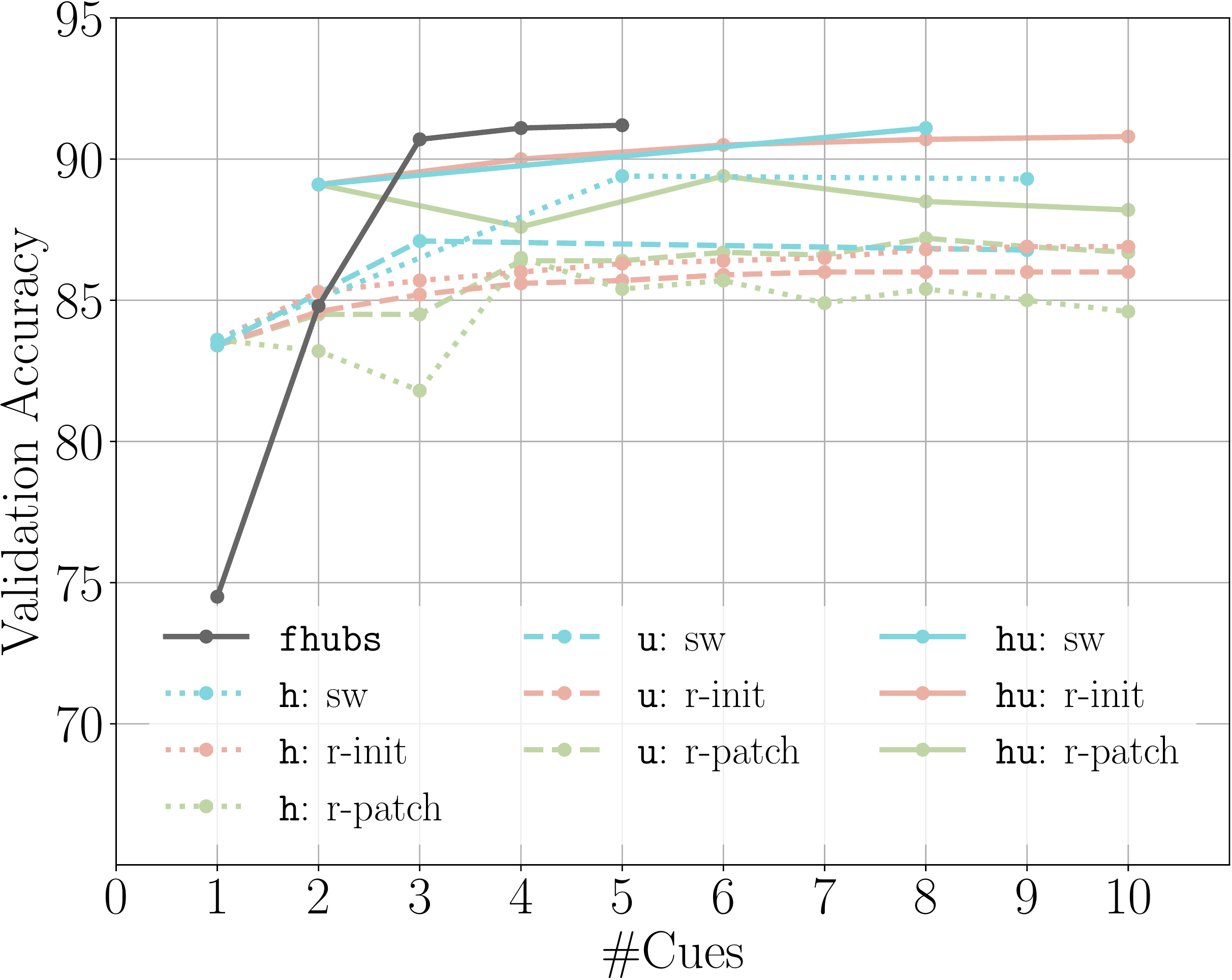} 
}\hfill
\subfloat[Day]{
\includegraphics[width=0.8\columnwidth]{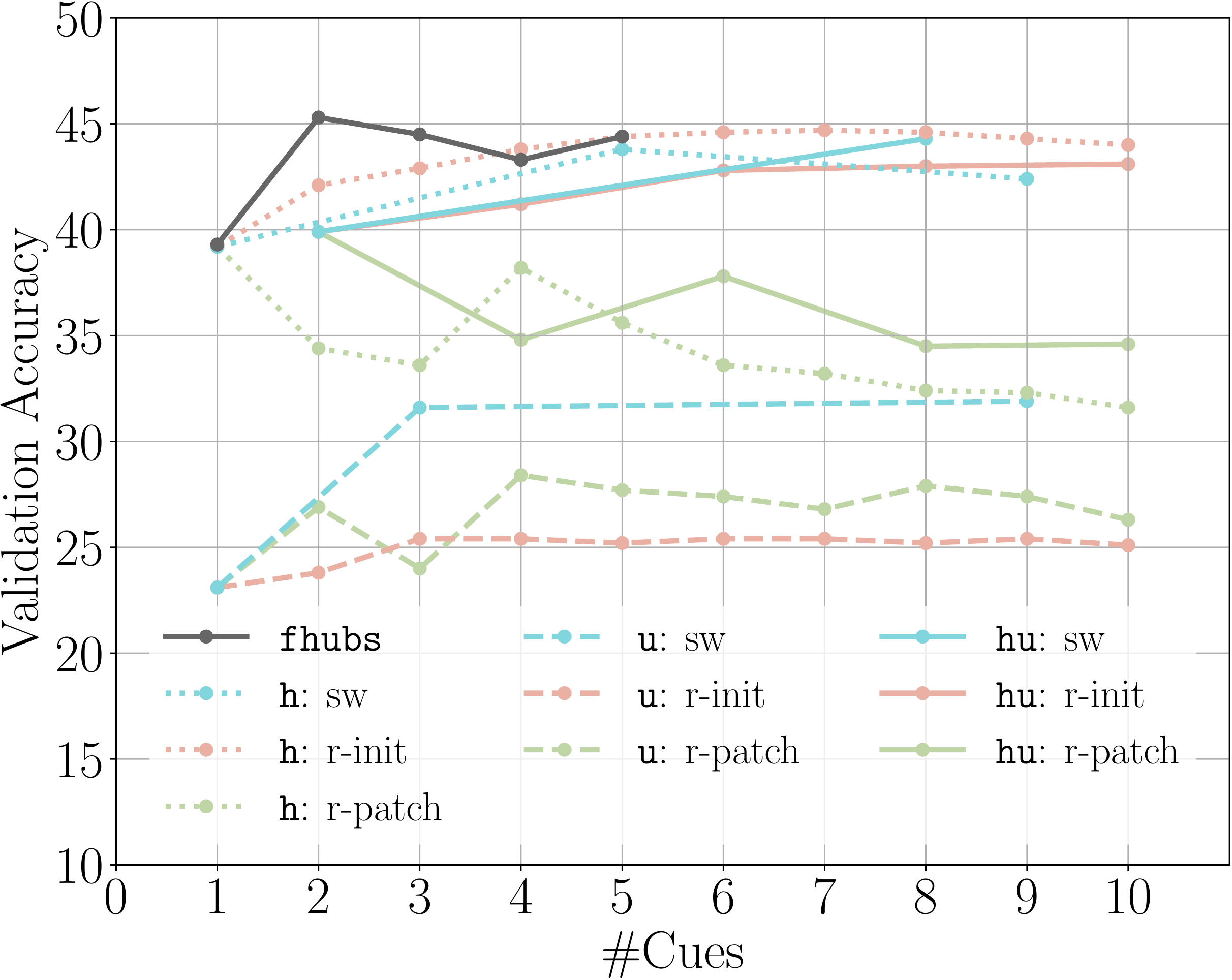} 
}\hfill
\vspace{0em}
\caption{\label{fig:plot-region-choice}Comparison of our region choice $\texttt{fhubs}$ against three types of baseline region types, random patch (r-patch), sliding window (sw), and random initialisation (r-init), based on head ($\texttt{h}$) and upper body ($\texttt{u}$) sized regions. We report the \emph{val} set accuracy against the number of cues used. The combination orders for r-patch and r-init  are random; for sw, we combine regions close to the head first.}
\end{figure*}

We compare the three types of regions against our choice of regions $\texttt{f}\texttt{h}\texttt{u}\texttt{b}\texttt{s}$ and quantitatively. See figure \ref{fig:plot-region-choice} for the plot showing the trade off between the accuracy and the number of cues used. For $\texttt{f}\texttt{h}\texttt{u}\texttt{b}\texttt{s}$, we progressively combine from $\texttt{f}$ to $\texttt{s}$ (5 cues maximum). For random patch (r-patch) and random initialisation (r-init), the combination order is randomised. For sliding window (sw), we combine the regions nearest to head first, and expand the combination diameter. Note that for every baseline region types (r-patch/init and sw), we consider combining head and upper body size boxes together ($\texttt{hu}$).

From figure \ref{fig:plot-region-choice} we observe, most importantly, that our choice of regions $\texttt{fhubs}$ gives the best performance-complexity (measured in number of cues) trade off among the regions and combinations considered under both small and large domain gaps (Original and Day splits). Amongst the baseline types, sw and r-init beat the r-patch in general; it is important to focus on the head region, the most identity-relevant part. In the Original split, it helps to combine $\texttt{h}$ and $\texttt{u}$ for all baseline types; it is important to combine multi-scale cues. However, the best trade off is given by $\texttt{fhubs}$, one that samples the from diverse regions and scales.

\subsubsection*{Conclusion}

Our choice of the regions $\texttt{fhubs}$ efficiently captures diverse identity information at diverse scales (only five cues). $\texttt{fhubs}$ beats the baseline region selection methods including random patches, sliding windows, and ensemble of randomly initialised cues in terms of the performance-complexity (number of cues) trade off.

\subsection{\label{sec:Scene}Scene ($\textnormal{\texttt{s}}$)}

\begin{table}
\begin{centering}
\begin{tabular}{lllcc}
 && Method && Accuracy\tabularnewline
\cline{1-1} \cline{3-3} \cline{5-5}
& \vspace{-0.9em}\tabularnewline
\cline{1-1} \cline{3-3} \cline{5-5}
Gist && \texttt{$\texttt{s}_{\texttt{gist}}$} && 21.56\tabularnewline
PlacesNet scores && \texttt{$\texttt{s}_{\texttt{places 205}}$} && 21.44\tabularnewline
raw PlacesNet && \texttt{$\texttt{s}_{0\texttt{ places}}$} && 27.37\tabularnewline
PlacesNet fine-tuned && \texttt{$\texttt{s}_{3\texttt{ places}}$} && 25.62\tabularnewline
raw AlexNet && \texttt{$\texttt{s}_{0}$} && 26.54\tabularnewline
AlexNet fine-tuned && \texttt{$\texttt{s}=\texttt{s}_{3}$} && 27.06\tabularnewline
\cline{1-1} \cline{3-3} \cline{5-5}
\end{tabular}
\par\end{centering}
\vspace{0.8em}

\caption{\label{tab:validation-set-scene}PIPA \emph{val} set accuracy of different scene cues. See descriptions in \S\ref{sec:Scene}.}
\end{table}

Scene region is the whole image containing the person of interest. 
Other than a fine-tuned AlexNet we considered multiple feature types
to encode the scene information. \texttt{$\texttt{s}_{\texttt{gist}}$}:
using the Gist descriptor \cite{Oliva2001IjcvGist} ($512$ dimensions).
\texttt{$\texttt{s}_{0\texttt{ places}}$}: instead of using AlexNet
pre-trained on ImageNet, we consider an AlexNet (PlacesNet) pre-trained
on $205$ scene categories of the ``Places Database'' \cite{Zhou2014NipsPlaces}
($\sim\negmedspace2.5$ million images). \texttt{$\texttt{s}_{\texttt{places 205}}$}:
Instead of the $4\,096$ dimensions PlacesNet feature vector, we also
consider using the score vector for each scene category ($205$ dimensions).
$\texttt{s}_{0}$,$\texttt{s}_{3}$: finally we consider using AlexNet
in the same way as for body or head (with zero or $300\mbox{k}$ iterations
of fine-tuning on the PIPA person recognition training set). \texttt{$\texttt{s}_{3\texttt{ places}}$: $\texttt{s}_{0\texttt{ places}}$
}fine-tuned for person recognition.

\subsubsection*{Results}

Table \ref{tab:validation-set-scene} compares the different alternatives
on the \emph{val} set. The Gist descriptor \texttt{$\texttt{s}_{\texttt{gist}}$}
performs only slightly below the convnet options (we also tried the
$4\,608$ dimensional version of Gist, obtaining worse results).
Using the raw (and longer) feature vector of \texttt{$\texttt{s}_{0\texttt{ places}}$}
is better than the class scores of \texttt{$\texttt{s}_{\texttt{places 205}}$}.
Interestingly, in this context pre-training for places classification
is better than pre-training for objects classification (\texttt{$\texttt{s}_{0\texttt{ places}}$}
versus \texttt{$\texttt{s}_{0}$}). After fine-tuning $\texttt{s}_{3}$
reaches a similar performance as \texttt{$\texttt{s}_{0\texttt{ places}}$}.\\
Experiments trying different combinations indicate that there is little
complementarity between these features. Since there is not a large
difference between \texttt{$\texttt{s}_{0\texttt{ places}}$} and
\texttt{$\texttt{s}_{3}$}, for the sake of simplicity we use \texttt{$\texttt{s}_{3}$}
as our scene cue \texttt{$\texttt{s}$} in all other experiments.

\subsubsection*{Conclusion}

Scene $\ensuremath{\texttt{s}}$ by itself, albeit weak, can obtain
results far above the chance level. After fine-tuning, scene recognition
as pre-training surrogate task \cite{Zhou2014NipsPlaces} does not
provide a clear gain over (ImageNet) object recognition.

\subsection{\label{sec:Head-or-face}Head ($\textnormal{\texttt{h}}$) or face
($\textnormal{\ensuremath{\texttt{f}}}$)?}

A large portion of work on face recognition focuses on the face region
specifically. In the context of photo albums, we aim to quantify how
much information is available in the head versus the face region. As discussed in \S\ref{subsec:face-detection}, we obtain the face regions $\ensuremath{\texttt{f}}$ from the DPM face detector \cite{Mathias2014Eccv}.

\subsubsection*{Results}

There is a large gap of $\sim\negmedspace10$ percent points performance
between $\texttt{f}$ and $\texttt{h}$ in table \ref{tab:validation-set-regions-accuracy}
highlighting the importance of including the hair and background around
the face.

%When evaluating only on the frontal faces $\texttt{FR}$ of the validation set $\texttt{f}$ reaches $81\%$ accuracy and $70\%$ for non-fronal and non-detections $\texttt{NFR}\cup\texttt{NDET}$. The performance drop between frontal versus handling profile and back views is less dramatic than one could have suspected. See \S \ref{subsec:Head orientation analysis} for more head orientation analysis.

\subsubsection*{Conclusion}

Using $\texttt{h}$ is more effective than $\texttt{f}$, but $\texttt{f}$ result still shows a fair performance. As with other body cues, there is a complementarity between $\texttt{h}$ and $\texttt{f}$; we suggest to use them together.

\begin{table}
\begin{centering}
\begin{tabular}{lllcc}
 && Method && Accuracy\tabularnewline
\cline{1-1} \cline{3-3} \cline{5-5} 
& \vspace{-0.9em} \tabularnewline
\cline{1-1} \cline{3-3} \cline{5-5} 
More data (\S\ref{sec:Additional-training-data}) && $\texttt{h}$ && 83.88\tabularnewline
 && $\texttt{h}+\mbox{\ensuremath{\texttt{h}}}_{\texttt{cacd}}$ && 84.88\tabularnewline
 && $\texttt{h}+\mbox{\ensuremath{\texttt{h}}}_{\texttt{casia}}$ && 86.08\tabularnewline
 && $\texttt{h}+\mbox{\ensuremath{\texttt{h}}}_{\texttt{casia}}+\mbox{\ensuremath{\texttt{h}}}_{\texttt{cacd}}$\hspace*{-1.5em} && 86.26\tabularnewline
\cline{1-1} \cline{3-3} \cline{5-5} 
Attributes (\S\ref{sec:Attributes})  & &$\mbox{\ensuremath{\texttt{h}}}_{\texttt{pipa11m}}$  && 74.63\tabularnewline
 && $\mbox{\ensuremath{\texttt{h}}}_{\texttt{pipa11}}$ & & 81.74\tabularnewline
 && $\texttt{h}+\mbox{\ensuremath{\texttt{h}}}_{\texttt{pipa11}}$ && 85.00\tabularnewline
\arrayrulecolor{gray}
 \cline{3-3} \cline{5-5} 
\arrayrulecolor{black} && $\mbox{\ensuremath{\texttt{u}}}_{\texttt{peta5}}$ && 77.50\tabularnewline
 && $\texttt{u}+\mbox{\ensuremath{\texttt{u}}}_{\texttt{peta5}}$ && 85.18\tabularnewline
\arrayrulecolor{gray}
\cline{3-3} \cline{5-5} 
\arrayrulecolor{black} && $\mbox{\ensuremath{\texttt{A}}}=\mbox{\ensuremath{\texttt{h}}}_{\texttt{pipa11}}+\mbox{\ensuremath{\texttt{u}}}_{\texttt{peta5}}$\hspace*{-1.5em} && 86.17\tabularnewline
 && $\texttt{h}+\texttt{u}$ && 85.77\tabularnewline
 && $\texttt{h}+\texttt{u}+\ensuremath{\texttt{A}}$ && 90.12\tabularnewline
\cline{1-1} \cline{3-3} \cline{5-5} 
\texttt{naeil} (\S\ref{sec:conf-naeil}) && \texttt{naeil}\cite{oh2015person} && 91.70\tabularnewline
\cline{1-1} \cline{3-3} \cline{5-5} 
\end{tabular}
\par\end{centering}
\vspace{0.8em}

\caption{\label{tab:validation-set-extended-data-accuracy}PIPA \emph{val} set accuracy of different cues based on extended data. See \S\ref{sec:Additional-training-data}, \S\ref{sec:Attributes}, and \S\ref{sec:conf-naeil} for details.}
\end{table}

\subsection{\label{sec:Additional-training-data}Additional training data ($\textnormal{\ensuremath{\mbox{\ensuremath{\texttt{h}}}_{\texttt{cacd}}},\,\ensuremath{\mbox{\ensuremath{\texttt{h}}}_{\texttt{casia}}}}$)}

It is well known that deep learning architectures benefit from additional
data. DeepFace \cite{Taigman2014CvprDeepFace} used by \texttt{PIPER} \cite{Zhang2015CvprPiper} is trained over $4.4\cdot10^{6}$
faces of $4\cdot10^{3}$ persons (the private SFC dataset \cite{Taigman2014CvprDeepFace}).
In comparison our cues are trained over ImageNet and PIPA's $29\cdot10^{3}$
faces over $1.4\cdot10^{3}$ persons. To measure the effect of training
on larger data we consider fine-tuning using two open source face recognition
datasets: CASIA-WebFace (CASIA) \cite{Yi2014ArxivLearningFace} and
the ``Cross-Age Reference Coding Dataset'' (CACD) \cite{Chen2014Eccv}.

CASIA contains $0.5\cdot10^{6}$ images of $10.5\cdot10^{3}$ persons
(mainly actors and public figures). When fine-tuning AlexNet
over these identities (using the head area $\mbox{\ensuremath{\texttt{h}}}$),
we obtain the $\mbox{\ensuremath{\texttt{h}}}_{\texttt{casia}}$ cue. 

CACD contains $160\cdot10^{3}$ faces of $2\cdot10^{3}$ persons 
with varying ages. Although smaller in total number of images than CASIA, CACD features greater number of samples per identity ($\sim\negmedspace2\times)$.
The $\mbox{\ensuremath{\texttt{h}}}_{\texttt{cacd}}$ cue is built
via the same procedure as $\mbox{\ensuremath{\texttt{h}}}_{\texttt{casia}}$.

\subsubsection*{Results}

See the top part of table \ref{tab:validation-set-extended-data-accuracy} for the results. $\texttt{h}+\mbox{\ensuremath{\texttt{h}}}_{\texttt{cacd}}$
and $\texttt{h}+\mbox{\ensuremath{\texttt{h}}}_{\texttt{casia}}$ improve
over $\texttt{h}$ (1.0 and 2.2 pp, respectively). Extra convnet training data seems to help. However, due to the mismatch in data distribution, 
$\mbox{\ensuremath{\texttt{h}}}_{\texttt{cacd}}$ and $\mbox{\ensuremath{\texttt{h}}}_{\texttt{casia}}$
on their own are about $\sim\negmedspace5\ \mbox{pp}$ worse than
$\texttt{h}$.

\subsubsection*{Conclusion}

Extra convnet training data helps, even if they are from different type of photos. 

\subsection{\label{sec:Attributes}Attributes ($\textnormal{\texttt{\ensuremath{\mbox{\ensuremath{\texttt{h}}}_{\texttt{pipa11}}},\,\ensuremath{\mbox{\ensuremath{\texttt{u}}}_{\texttt{peta5}}}}}$)}

Albeit overall appearance might change day to day, one could expect
that stable, long term attributes provide means for recognition. We build attribute cues by fine-tuning AlexNet features not for the person recognition task (like for all other cues), but rather for the attribute prediction surrogate task. We consider two sets attributes, one on the head region and the other on the upper body region.

We have annotated identities in the PIPA \emph{train} and \emph{val} sets ($1409+366$ in total) with five long term attributes: age, gender, glasses, hair colour, and hair length (see table \ref{tab:attributes-details} for details). We build $\mbox{\ensuremath{\texttt{h}}}_{\texttt{pipa11}}$ by fine-tuning AlexNet features for the task of head attribute prediction.

For fine-tuning the attribute cue $\mbox{\ensuremath{\texttt{h}}}_{\texttt{pipa11}}$, we consider two approaches: training a single
network for all attributes as a multi-label classification problem
with the sigmoid cross entropy loss, or tuning one network per attribute
separately and concatenating the feature
vectors. The results on the \emph{val} set indicate the latter
($\mbox{\ensuremath{\texttt{h}}}_{\texttt{pipa11}}$) performs better
than the former ($\mbox{\ensuremath{\texttt{h}}}_{\texttt{pipa11m}}$).

For the upper body attribute features, we use the ``PETA pedestrian attribute dataset''
\cite{Deng2014AcmPeta}. The dataset originally has $105$ attributes annotations for $19\cdot10^{3}$ full-body pedestrian images. We chose the five long-term attributes for our study: gender, age (young adult, adult), black hair, and short hair
(details in table \ref{tab:attributes-details}). We choose to use the upper-body $\texttt{u}$ rather than the full body $\texttt{b}$ for attribute prediction -- the crops are much less noisy. We train the AlexNet feature on upper body of PETA images with the attribute prediction task to obtain the cue $\mbox{\ensuremath{\texttt{u}}}_{\texttt{peta5}}$.

\subsubsection*{Results}

See results in table \ref{tab:validation-set-extended-data-accuracy}. Both PIPA ($\mbox{\ensuremath{\texttt{h}}}_{\texttt{pipa11}}$)
and PETA ($\mbox{\ensuremath{\texttt{u}}}_{\texttt{peta5}}$) annotations
behave similarly ($\sim\negmedspace1\ \mbox{pp}$ gain over $\texttt{h}$
and $\texttt{u}$), and show complementary
($\sim\negmedspace5\ \mbox{pp}$ gain over $\texttt{h}\negmedspace+\negmedspace\texttt{u}$).
Amongst the attributes considered, gender contributes the most to
improve recognition accuracy (for both attributes datasets). 

\subsubsection*{Conclusion}

Adding attribute information improves the performance. 

\begin{table}
\begin{centering}
\begin{tabular}{lllll}
Attribute && Classes& & Criteria\tabularnewline
\cline{1-1} \cline{3-3} \cline{5-5}
& \vspace{-0.9em}\tabularnewline
\cline{1-1} \cline{3-3} \cline{5-5}
Age && Infant && {\footnotesize{}Not walking (due to young age)}\tabularnewline
 && Child & &{\footnotesize{}Not fully grown body size}\tabularnewline
 && Young Adult && {\footnotesize{}Fully grown \& Age $<45$}\tabularnewline
 && Middle Age && {\footnotesize{}$45\leq\mbox{Age}\leq60$}\tabularnewline
 && Senior && {\footnotesize{}Age$\geq60$}\tabularnewline
\cline{1-1} \cline{3-3} \cline{5-5} 
Gender && Female && {\footnotesize{}Female looking}\tabularnewline
 && Male && {\footnotesize{}Male looking}\tabularnewline
\cline{1-1} \cline{3-3} \cline{5-5}
Glasses && None && {\footnotesize{}No eyewear}\tabularnewline
 && Glasses && {\footnotesize{}Transparant glasses}\tabularnewline
 && Sunglasses && {\footnotesize{}Glasses with eye occlusion}\tabularnewline
\cline{1-1} \cline{3-3} \cline{5-5}
Haircolour && Black && {\footnotesize{}Black}\tabularnewline
 && White && {\footnotesize{}Any hint of whiteness}\tabularnewline
 && Others && {\footnotesize{}Neither of the above}\tabularnewline
\cline{1-1} \cline{3-3} \cline{5-5}
Hairlength && No hair && {\footnotesize{}Absolutely no hair on the scalp}\tabularnewline
 && Less hair && {\footnotesize{}Hairless for $>\frac{1}{2}$ upper scalp}\tabularnewline
 && Short hair && {\footnotesize{}When straightened,$<10$ cm}\tabularnewline
 && Med hair && {\footnotesize{}When straightened, $<$chin level}\tabularnewline
 && Long hair && {\footnotesize{}When straightened, $>$chin level}\tabularnewline
 \cline{1-1} \cline{3-3} \cline{5-5}
\end{tabular}
\par\end{centering}
\begin{centering}
\par\end{centering}
\vspace{0.8em}

\caption{\label{tab:attributes-details}PIPA attributes details.}
\end{table}

\subsection{\label{sec:conf-naeil}Conference version final model ($\textnormal{\ensuremath{\mbox{\ensuremath{\texttt{naeil}}}}}$) \cite{oh2015person}}

% $\texttt{P}=\texttt{f}\negthinspace+\negthinspace\texttt{h}\negthinspace+\negthinspace\texttt{u}\negthinspace+\negthinspace\texttt{b}$

The final model in the conference version of this paper combines five vanilla regional cues ($\texttt{\ensuremath{\mbox{\ensuremath{\texttt{P}}}_{s}}}=\texttt{P}\negthinspace+\negthinspace\texttt{s}$), two head cues trained with extra data ($\textnormal{\ensuremath{\mbox{\ensuremath{\texttt{h}}}_{\texttt{cacd}}}$, $\ensuremath{\mbox{\ensuremath{\texttt{h}}}_{\texttt{casia}}}}$), and ten attribute cues ($\mbox{\ensuremath{\texttt{h}}}_{\texttt{pipa11}}$, $\mbox{\ensuremath{\texttt{u}}}_{\texttt{peta5}}$), resulting in 17 cues in total. We name this method $\texttt{naeil}$ \cite{oh2015person}\footnote{``naeil'', \includegraphics[height=0.8em]{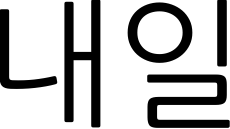}, means ``tomorrow'' and sounds like ``nail''.}. 

\subsubsection*{Results}

See table \ref{tab:validation-set-extended-data-accuracy} for the results. \texttt{naeil}, by combining all the cues considered naively, achieves the best result 91.70\% on the \emph{val} set.

\subsubsection*{Conclusion}

Cues considered thus far are complementary, and the combined model \texttt{naeil} is effective.

\subsection{\label{sec:deepid}DeepID2+ face recognition module ($\textnormal{\ensuremath{\mbox{\ensuremath{\texttt{h}}}_{\texttt{deepid}}}}$)  \cite{Sun2014ArxivDeepId2plus}}

Face recognition performance have improved significantly in recent years with better architectures and larger open source datasets \cite{Huang2007Lfw,Zhu2013Iccv,Taigman2014CvprDeepFace,Sun2014ArxivDeepId2plus,Ding2015Arxiv,Schroff2015ArxivFaceNet,Parkhi15,chen2016unconstrained,wen2016discriminative}. In this section, we study how much face recognition helps in person recognition. While DeepFace \cite{Taigman2014CvprDeepFace} used by the \texttt{PIPER} \cite{Zhang2015CvprPiper} would have enabled more direct comparison against the \texttt{PIPER}, it is not publicly available. We thus choose the DeepID2+ face recogniser \cite{Sun2014ArxivDeepId2plus}. Face recognition technology is still improving quickly, and larger and larger face datasets are being released -- the analysis in this section would be an underestimate of current and future face recognisers.

The DeepID2+ network is a siamese neural network that takes 25 different crops of head as input, with the joint verification-identification loss. The training is based on large databases consisting of CelebFaces+\cite{sun2014deep}, WDRef\cite{Chen2012}, and LFW\cite{Huang2007Lfw} -- totalling $2.9\cdot10^{5}$ faces of $1.2\cdot10^{4}$ persons. At test time, it ensembles the predictions from the 25 crop regions obtained by facial landmark detections. The resulting output is $1\,024$ dimensional head feature that we denote as $\textnormal{\ensuremath{\mbox{\ensuremath{\texttt{h}}}_{\texttt{deepid}}}}$.

Since the DeepID2+ pipeline begins with the facial landmark detection, the DeepID2+ features are not available for instances with e.g. occluded or backward orientation heads. As a result, only $52\,709$ out of $63\,188$ instances ($83.42\%$) have the DeepID2+ features available, and we use vectors of zeros as features for the rest. % ($10\,479$ heads). 

\subsubsection*{Results - Original split}

See table \ref{tab:deepid-val} for the \emph{val} set results for $\textnormal{\ensuremath{\mbox{\ensuremath{\texttt{h}}}_{\texttt{deepid}}}}$ and related combinations. $\textnormal{\ensuremath{\mbox{\ensuremath{\texttt{h}}}_{\texttt{deepid}}}}$ in itself is weak ($68.46\%$) compared to the vanilla head feature $\texttt{h}$, due to the missing features for the back-views. However, when combined with $\texttt{h}$, the performance reaches $85.86\%$ by exploiting information from strong DeepID2+ face features and the viewpoint robust $\texttt{h}$ features.

Since the feature dimensions are not homogeneous ($4\,096$ versus $1\,024$), we try $L_{2}$ normalisation of $\texttt{h}$ and $\textnormal{\ensuremath{\mbox{\ensuremath{\texttt{h}}}_{\texttt{deepid}}}}$ before concatenation ($\texttt{h}\oplus\mbox{\ensuremath{\texttt{h}}}_{\texttt{deepid}}$). This gives a further $3\%$ boost ($88.74\%$) -- better than $\texttt{h}+\mbox{\ensuremath{\texttt{h}}}_{\texttt{cacd}}+\mbox{\ensuremath{\texttt{h}}}_{\texttt{casia}}$, previous best model on the head region ($86.26\%$).

\subsubsection*{Results - Album, Time and Day splits}

Table \ref{tab:deepid-val} also shows results for the Album, Time, and Day splits on the \emph{val} set. While the general head cue \texttt{h} degrades significantly on the Day split, $\texttt{h}_{\texttt{deepid}}$ is a reliable cue with roughly the same level of recognition in all four splits ($60\negmedspace\sim\negmedspace70\%$). This is not surprising, since face is largely invariant over time, compared to hair, clothing, and event.

On the other splits as well, the complementarity of $\texttt{h}$ and $\textnormal{\ensuremath{\mbox{\ensuremath{\texttt{h}}}_{\texttt{deepid}}}}$ is guaranteed only when they are $L_2$ normalised before concatenation. The $L_2$ normalised concatenation $\texttt{h}\oplus\mbox{\ensuremath{\texttt{h}}}_{\texttt{deepid}}$ envelops the performance of individual cues on all splits.

\subsubsection*{Conclusion}

DeepID2+, with face-specific architecture/loss and massive amount of training data, contributes highly useful information for the person recognition task. However, being only able to recognise face-visible instances, it needs to be combined with orientation-robust $\texttt{h}$ to ensure the best performance. Unsurprisingly, having a specialised face recogniser helps more in the setup with larger appearance gap between training and testing samples (Album, Time, and Day splits). Better face recognisers will further improve the results in the future.

\begin{table}
\begin{centering}
\begin{tabular}{lccccc}
 Method && Original & Album & Time & Day\tabularnewline
\cline{1-1} \cline{3-6}
 &\vspace{-1em}\tabularnewline
\cline{1-1} \cline{3-6}
$\texttt{h}$ && 83.88 & 77.90 & 70.38 & 40.71\tabularnewline
$\mbox{\ensuremath{\texttt{h}}}_{\texttt{deepid}}$ && 68.46 & 66.91 & 64.16 & 60.46\tabularnewline
$\texttt{h}+\mbox{\ensuremath{\texttt{h}}}_{\texttt{deepid}}$ && 85.86 & 80.54 & 73.31 & 47.86\tabularnewline
$\texttt{h}\oplus\mbox{\ensuremath{\texttt{h}}}_{\texttt{deepid}}$ && 88.74 & 85.72 & 80.88 & 66.91\tabularnewline
\cline{1-1} \cline{3-6}
$\texttt{naeil}$\cite{oh2015person} && 91.70 & 86.37 & 80.66 & 49.21\tabularnewline
 $\texttt{naeil}+\mbox{\ensuremath{\texttt{h}}}_{\texttt{deepid}}$ && 92.11 & 86.77 & 81.08 & 51.02\tabularnewline
$\texttt{naeil2}$ && {93.42} & {89.95} & {85.87} & {70.58}\tabularnewline
\cline{1-1} \cline{3-6}
\end{tabular}
\par\end{centering}
\vspace{0.8em}

\caption{\label{tab:deepid-val}PIPA \emph{val} set accuracy of methods involving $\mbox{\ensuremath{\texttt{h}}}_{\texttt{deepid}}$. The optimal combination weights are $\lambda^{\star}=[0.60\;1.05\;1.00\;1.50]$ for Original, Album, Time, and Day splits, respectively.\protect \\
$\oplus$ means $L_{2}$ normalisation before concatenation.}

\end{table}

\subsection{\label{sec:naeil+deepid}Combining $\texttt{naeil}$ with $\texttt{h}_\texttt{deepid}$ ($\textnormal{\ensuremath{\mbox{\ensuremath{\texttt{naeil2}}}}}$)}

We build the final model of the journal version, namely the $\texttt{naeil2}$ by combining $\texttt{naeil}$ and $\texttt{h}_\texttt{deepid}$. As seen in \S\ref{sec:deepid}, naive concatenation is likely to fail due to even larger difference in dimensionality ($4\,096\times 17=69\,632$ versus $1\,024$). We consider $L_{2}$ normalisation of $\texttt{naeil}$ and $\texttt{h}_\texttt{deepid}$, and then performing a weighted concatenation.

\begin{equation}
\label{eq:weighted-sum}
\texttt{naeil}\oplus_\lambda\mbox{\ensuremath{\texttt{h}}}_{\texttt{deepid}} = \frac{\texttt{naeil}}{||\texttt{naeil}||_2}+\lambda\cdot\frac{\mbox{\ensuremath{\texttt{h}}}_{\texttt{deepid}}}{||\mbox{\ensuremath{\texttt{h}}}_{\texttt{deepid}}||_2},
\end{equation}
where, $\lambda>0$ is a parameter and $+$ denotes a concatenation.

\subsubsection*{Optimisation of $\lambda$ on \emph{val} set}

$\lambda$ determines how much relative weight to be given to $\texttt{h}_\texttt{deepid}$. As we have seen in \S\ref{sec:deepid}, the amount of additional contribution from $\texttt{h}_\texttt{deepid}$ is different for each split. In this section, we find $\lambda^\star$, the optimal values for $\lambda$, for each split over the \emph{val} set. The resulting combination of $\texttt{naeil}$ and $\texttt{h}_\texttt{deepid}$ is our final method, $\texttt{naeil2}$. $\lambda^\star$ is searched on the equi-distanced points $\{0,0.05,0.1,\cdots,3\}$.

\begin{figure}
\begin{centering}
\hspace*{\fill}\includegraphics[width=0.7\columnwidth]{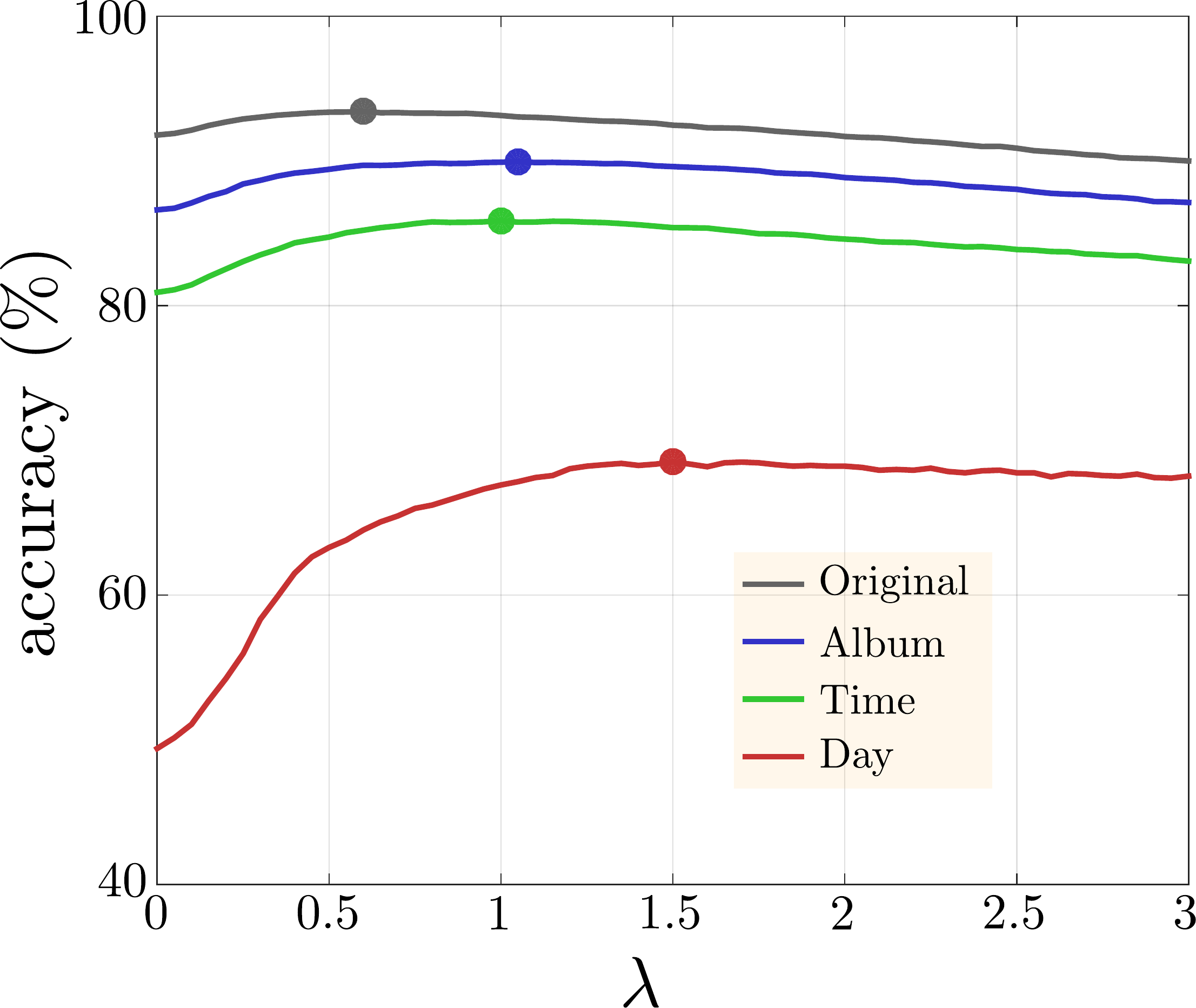}\hspace*{\fill}\vspace{0.5em}
\par\end{centering}
\caption{\label{fig:lambda-deepid+naeil}PIPA \emph{val} set accuracy of $\texttt{naeil}\oplus_\lambda\mbox{\ensuremath{\texttt{h}}}_{\texttt{deepid}}$
for varying values of $\lambda$. Round dots denote the maximal \emph{val} accuracy.}
\end{figure}

See figure \ref{fig:lambda-deepid+naeil} for the \emph{val} set performance of $\texttt{naeil}\oplus_\lambda\mbox{\ensuremath{\texttt{h}}}_{\texttt{deepid}}$ with varying values of $\lambda$. The optimal weights are found at  $\lambda^{\star}=[0.60\;1.05\;1.00\;1.50]$ for Original, Album, Time, and Day splits, respectively. The relative importance of $\mbox{\ensuremath{\texttt{h}}}_{\texttt{deepid}}$ is greater on splits with larger appearance changes. For each split, we denote $\texttt{naeil2}$ as the combination $\texttt{naeil}$ and $\texttt{h}_\texttt{deepid}$ based on the optimal weights.

Note that the performance curve is rather stable for $\lambda\geq1.5$ in all splits. In practice, when the expected amount of appearance changes of subjects are unknown, our advice would be to choose $\lambda\approx 1.5$. Finally, we remark that the weighted sum can also be done for the $17$ cues in $\texttt{naeil}$; finding the optimal cue weights is left as a future work.

\subsubsection*{Results}

See table \ref{tab:deepid-val} for the results of combining $\texttt{naeil}$ and $\texttt{h}_\texttt{deepid}$. Naively concatenated, $\texttt{naeil}+\texttt{h}_\texttt{deepid}$ performs worse than $\texttt{h}_\texttt{deepid}$ on the Day split ($51.02\%$ vs $60.46\%$). However, the weighted combination \texttt{naeil2} achieves the best performance on all four splits.

\subsubsection*{Conclusion}

When combining $\texttt{naeil}$ and $\texttt{h}_\texttt{deepid}$, a weighted combination is desirable, and the resulting final model \texttt{naeil2} beats all the previously considered models on all four splits.

\section{\label{sec:Test-set-results}PIPA test set results and comparison}

\begin{table*}
\begin{centering}
\begin{tabular}{cclcllcllccccc}
&&        && \multicolumn{2}{c}{Special modules} && \multicolumn{2}{c}{General features} &&  \tabularnewline
\cline{5-6} \cline{8-9}
&& Method && Face rec. & Pose est. && Data & Arch. && Original & Album & Time & Day\tabularnewline
\cline{3-3} \cline{5-6} \cline{8-9} \cline{11-14}
&\vspace{-1em}\tabularnewline
\cline{3-3} \cline{5-6} \cline{8-9} \cline{11-14}
&& Chance level && \ding{55} & \ding{55} && $-$ & $-$ && \hspace{0.5em}0.78 & \hspace{0.5em}0.89 & \hspace{0.5em}0.78 & \hspace{0.5em}1.97\tabularnewline
\cline{1-1} \cline{3-3} \cline{5-6} \cline{8-9} \cline{11-14}
\multirow{6}{*}{\rotatebox{90}{Head\hspace{0.0em}}}&& $\mbox{\ensuremath{\texttt{h}}}_{\texttt{rgb}}$ && \ding{55} & \ding{55} && $-$ & $-$ && 33.77 & 27.19 & 16.91 & \hspace{0.5em}6.78\tabularnewline
&& $\mbox{\ensuremath{\texttt{h}}}$ && \ding{55} & \ding{55} && I+P & Alex && 76.42 & 67.48 & 57.05 & 36.48\tabularnewline
&& $\texttt{h}\negthinspace+\negthinspace\mbox{\ensuremath{\texttt{h}}}_{\texttt{casia}}\negthinspace+\negthinspace\mbox{\ensuremath{\texttt{h}}}_{\texttt{cacd}}$ && \ding{55} & \ding{55} && I+P+CC & Alex && 80.32 & 72.82 & 63.18 & 45.45\tabularnewline
&& \texttt{\small{}$\mbox{\ensuremath{\texttt{h}_{\texttt{deepid}}}}$} && DeepID2+\cite{Sun2014ArxivDeepId2plus} & \ding{55} && $-$ & $-$ && 68.06 & 65.49 & 60.69 & 61.49\tabularnewline
&& $\texttt{h}\oplus\mbox{\ensuremath{\texttt{h}}}_{\texttt{deepid}}$ && DeepID2+\cite{Sun2014ArxivDeepId2plus} & \ding{55} && I+P & Alex && 85.94 & 81.95 & 75.85 & 66.00\tabularnewline
 && \texttt{DeepFace}\cite{Zhang2015CvprPiper} && DeepFace\cite{Taigman2014CvprDeepFace} & \ding{55} && $-$ & $-$ && 46.66 & $-$ & $-$ & $-$\tabularnewline
\cline{1-1} \cline{3-3} \cline{5-6} \cline{8-9} \cline{11-14}
\multirow{7}{*}{\rotatebox{90}{Body\hspace{0.0em}}}&& $\mbox{\ensuremath{\texttt{b}}}$ && \ding{55} & \ding{55} &&  I+P & Alex && 69.63 & 59.29 & 44.92 & 20.38\tabularnewline
&& $\texttt{h}\negthinspace+\negthinspace\texttt{b}$ && \ding{55} & \ding{55} && I+P & Alex && 83.36 & 73.97 & 63.03 & 38.15\tabularnewline
&& $\texttt{P}=\texttt{f}\negthinspace+\negthinspace\texttt{h}\negthinspace+\negthinspace\texttt{u}\negthinspace+\negthinspace\texttt{b}$ && \ding{55} & \ding{55} && I+P & Alex && 85.33 & 76.49 & 66.55 & 42.17\tabularnewline
 && \texttt{GlobalModel}\cite{Zhang2015CvprPiper} && \ding{55} & \ding{55} && I+P & Alex && 67.60 & $-$ & $-$ & $-$\tabularnewline
&& \texttt{PIPER}\cite{Zhang2015CvprPiper} && DeepFace\cite{Taigman2014CvprDeepFace} & Poselets\cite{Bourdev2009IccvPoselets} && I+P & Alex && 83.05 & $-$ & $-$ & $-$\tabularnewline
&& \texttt{Pose}\cite{kumar2017pose} && \ding{55} & Pose group && I+P+V & Alex && 89.05 & 82.37 & 74.84 & 56.73\tabularnewline
&& \texttt{COCO}\cite{liu_2017_coco} && \ding{55} & Part det.\cite{ren15fasterrcnn} && I+P & Goog,Res && \textbf{92.78} & 83.53 & 77.68 & 61.73\tabularnewline
\cline{1-1} \cline{3-3} \cline{5-6} \cline{8-9} \cline{11-14}
\multirow{4}{*}{\rotatebox{90}{Image\hspace{0.0em}}} && $\texttt{\ensuremath{\mbox{\ensuremath{\texttt{P}}}_{s}}}=\texttt{P}\negthinspace+\negthinspace\texttt{s}$ && \ding{55} & \ding{55} && I+P & Alex && 85.71 & 76.68 & 66.55 & 42.31\tabularnewline
&& $\texttt{naeil}=\texttt{\ensuremath{\mbox{\ensuremath{\texttt{P}}}_{s}}}\negthinspace+\negthinspace\texttt{E}$\cite{oh2015person} && \ding{55} & \ding{55} && I+P+E & Alex && 86.78 & 78.72 & 69.29 & 46.54\tabularnewline
&& \texttt{Contextual}\cite{Li_2016_CVPR} && DeepID\cite{sun2014deep} & \ding{55} && I+P & Alex && 88.75 & 83.33 & 77.00 & 59.35\tabularnewline
%\texttt{Contextual+album \cite{Li_2016_CVPR}} & 93.91 & 83.44 & 80.23 & 61.62\tabularnewline
\cline{3-3} \cline{5-6} \cline{8-9} \cline{11-14}
&& $\texttt{naeil2}$ (this paper) && DeepID2+\cite{Sun2014ArxivDeepId2plus} & \ding{55} && I+P+E & Alex && {90.42} & \textbf{86.30} & \textbf{80.74} & \textbf{70.58}\tabularnewline
\cline{1-1} \cline{3-3} \cline{5-6} \cline{8-9} \cline{11-14}
\end{tabular}
\par\end{centering}
\vspace{0.8em}

\caption{\label{tab:test-set-accuracy-four-splits}PIPA \emph{test} set accuracy (\%) of the proposed method and prior arts on the four splits. For each method, we indicate any face recognition or pose estimation module included, and the data and convnet architecture for other features. \protect \\
Cues on extended data $\texttt{E}=\mbox{\ensuremath{\texttt{h}}}_{\texttt{casia}}\negthinspace+\negthinspace\mbox{\ensuremath{\texttt{h}}}_{\texttt{cacd}}\negthinspace+\negthinspace\texttt{\ensuremath{\mbox{\ensuremath{\texttt{h}}}_{\texttt{pipa11}}}+\ensuremath{\mbox{\ensuremath{\texttt{u}}}_{\texttt{peta5}}}}$.\protect \\
$\oplus$ means concatenation after $L_{2}$ normalisation.\protect \\
In the data column, I indicates ImageNet\cite{Deng2009CvprImageNet} and P indicates PIPA \emph{train} set. CC means CACD\cite{Chen2014Eccv}$+$CASIA\cite{Yi2014ArxivLearningFace} and E means CC$+$PETA\cite{Deng2014AcmPeta}. V indicates the VGGFace dataset \cite{Parkhi15}.\protect \\
In the architecture column, (Alex,Goog,Res) refers to (AlexNet\cite{Krizhevsky2012Nips},GoogleNetv3\cite{szegedy2016rethinking},ResNet50\cite{He_2016_CVPR}).
}
\end{table*}

%All experiments in this section are limited to a person recognition scenario where head boxes are provided by human annotations, and all test faces belong to a known finite set, following the protocol in \cite{Zhang2015CvprPiper}. (\S\ref{sec:Towards-open-world} is concerned with forgoing this assumption.) 

In this section, we measure the performance of our final model and key intermediate results on the PIPA \emph{test} set, and compare against the prior arts. See table \ref{tab:test-set-accuracy-four-splits} for a summary.

\subsection{\label{subsec:face-rgb-baseline}Baselines}

We consider two baselines for measuring the inherent difficulty of the task. First baseline is the ``chance level'' classifier, which does not see the image content and simply picks the most commonly occurring class. It provides the lower bound for any recognition method, and gives a sense of how large the gallery set is.

Our second baseline is the raw RGB nearest neighbour classifier $\mbox{\ensuremath{\texttt{h}}}_{\texttt{rgb}}$. It uses the raw downsized ($40\negmedspace\times\negmedspace40\ \mbox{pixels}$) and blurred RGB head crop as the feature. The identity of the Euclidean distance nearest neighbour training image is predicted at test time. By design, $\mbox{\ensuremath{\texttt{h}}}_{\texttt{rgb}}$ is only able to recognize
near identical head crops across the $\mbox{\emph{test}}_{\nicefrac{0}{1}}$ splits.

\subsubsection*{Results}

See results for ``chance level'' and $\mbox{\ensuremath{\texttt{h}}}_{\texttt{rgb}}$ in table \ref{tab:test-set-accuracy-four-splits}. While the ``chance level'' performance is low ($\leq2\%$ in all splits), we observe that $\mbox{\ensuremath{\texttt{h}}}_{\texttt{rgb}}$ performs unreasonably well on the Original split (33.77\%). This shows that the Original splits share many nearly identical person instances across the split, and the task is very easy. On the harder splits, we see that the $\mbox{\ensuremath{\texttt{h}}}_{\texttt{rgb}}$ performance diminishes, reaching only 6.78\% on the Day split. Recognition on the Day split is thus far less trivial -- simply taking advantage of pixel value similarity would not work.

\subsubsection*{Conclusion}

Although the gallery set is large enough, the task can be made arbitrarily easy by sharing many similar instances across the splits (Original split). We have remedied the issue by introducing three more challenging splits (Album, Time, and Day) on which the naive RGB baseline ($\mbox{\ensuremath{\texttt{h}}}_{\texttt{rgb}}$) no longer works (\S\ref{subsec:PIPA-splits}).

\subsection{Methods based on head}

We consider our four intermediate models ($\textnormal{\texttt{h}}$, $\texttt{h}\negthinspace+\negthinspace\mbox{\ensuremath{\texttt{h}}}_{\texttt{casia}}\negthinspace+\negthinspace\mbox{\ensuremath{\texttt{h}}}_{\texttt{cacd}}$, $\mbox{\ensuremath{\texttt{h}_{\texttt{deepid}}}}$, $\texttt{h}\oplus\mbox{\ensuremath{\texttt{h}}}_{\texttt{deepid}}$) and a prior work \texttt{DeepFace} \cite{Zhang2015CvprPiper,Taigman2014CvprDeepFace}.

We observe the same trend as described in the previous sections on the \emph{val} set (\S\ref{sec:Head-or-face}, \ref{sec:deepid}). Here, we focus on the comparison against \texttt{DeepFace} \cite{Taigman2014CvprDeepFace}. Even without a specialised face module, $\textnormal{\texttt{h}}$ already performs better than \texttt{DeepFace} (76.42\% versus 46.66\%, Original split). We believe this is for two reasons: (1) \texttt{DeepFace} only takes face regions as input, leaving out valuable hair and background information (\S\ref{sec:Head-or-face}), (2) \texttt{DeepFace} only makes predictions on 52\% of the instances where the face can be registered. Note that $\mbox{\ensuremath{\texttt{h}_{\texttt{deepid}}}}$ also do not always make prediction due to the failure to estimate the pose (17\% failure on PIPA), but performs better than \texttt{DeepFace} in the considered scenario (68.06\% versus 46.66\%, Original split).

\subsection{Methods based on body}

We consider three of our intermediate models ($\textnormal{\texttt{b}}$, $\texttt{h}\negthinspace+\negthinspace\texttt{b}$, $\texttt{P}=\texttt{f}\negthinspace+\negthinspace\texttt{h}\negthinspace+\negthinspace\texttt{u}\negthinspace+\negthinspace\texttt{b}$) and four prior arts (\texttt{GlobalModel}\cite{Zhang2015CvprPiper}, \texttt{PIPER}\cite{Zhang2015CvprPiper}, \texttt{Pose}\cite{kumar2017pose}, \texttt{COCO}\cite{liu_2017_coco}). \texttt{Pose} \cite{kumar2017pose} and \texttt{COCO} \cite{liu_2017_coco} methods appeared after the publication of the conference version of this paper \cite{oh2015person}. See table \ref{tab:test-set-accuracy-four-splits} for the results. 

Our body cue $\mbox{\ensuremath{\texttt{b}}}$ and Zhang et al.'s \texttt{GlobalModel} \cite{Zhang2015CvprPiper} are the same methods implemented independently. Unsurprisingly, they perform similarly (69.63\% versus 67.60\%, Original split). 

Our $\texttt{h}\negthinspace+\negthinspace\texttt{b}$ method is the minimal system matching Zhang et al.'s \texttt{PIPER} \cite{Zhang2015CvprPiper} ($83.36\%$ versus 83.05\%, Original split). The feature vector of $\texttt{h}\negthinspace+\negthinspace\texttt{b}$ is about $50$ times smaller than \texttt{PIPER}, and do not make use of face recogniser or pose estimator.

In fact, \texttt{PIPER} captures the head region via one of its poselets. Thus, $\texttt{h}\negthinspace+\negthinspace\texttt{b}$ extracts cues from a subset of \texttt{PIPER}'s ``\texttt{GlobalModel+Poselets}''
\cite{Zhang2015CvprPiper}, but performs better (83.36\% versus $78.79\%$, Original split).

\subsubsection*{Methods since the conference version\cite{oh2015person}}

\texttt{Pose} by Kumar et al. \cite{kumar2017pose} uses extra keypoint annotations on the PIPA \emph{train} set to generate pose clusters, and train separate models for each pose cluster (PSM, pose-specific models). By performing a form of pose normalisation they have improved the results significantly: 2.27 pp and 10.19 pp over \texttt{naeil} on Original and Day splits, respectively.

\texttt{COCO} by Liu et al. \cite{liu_2017_coco} proposes a novel metric learning loss for the person recognition task. Metric learning gives an edge over classifier-based methods by enabling recognition of unseen identities without re-training. They further use Faster-RCNN detectors \cite{ren15fasterrcnn} to localise face and body more accurately. The final performance is arguably good in all four splits, compared to \texttt{Pose} \cite{kumar2017pose} or \texttt{naeil} \cite{oh2015person}. However, one should note that the face, body, upper body, and full body features in \texttt{COCO} are based on GoogleNetv3 \cite{szegedy2016rethinking} and ResNet50 \cite{He_2016_CVPR} -- the numbers are not fully comparable to all the other methods that are largely based on AlexNet.

\subsection{Methods based on full image}

We consider our two intermediate models ($\texttt{\ensuremath{\mbox{\ensuremath{\texttt{P}}}_{s}}}=\texttt{P}\negthinspace+\negthinspace\texttt{s}$, $\texttt{naeil}=\texttt{\ensuremath{\mbox{\ensuremath{\texttt{P}}}_{s}}}\negthinspace+\negthinspace\texttt{E}$) and \texttt{Contextual} \cite{Li_2016_CVPR}, a method which appeared after the conference version of this paper \cite{oh2015person}.

Our \texttt{naeil} performs better than \texttt{PIPER} \cite{Zhang2015CvprPiper} (86.78\% versus 83.05\%, Original split), while having a 6 times smaller feature vector and not relying on face recogniser or pose estimator.

\subsubsection*{Methods since the conference version\cite{oh2015person}}

\texttt{Contextual} by Li et al. \cite{Li_2016_CVPR} makes use of person co-occurrence statistics to improve the results. It performs 1.97 pp and 12.81 pp better than \texttt{naeil} on Original and Day splits, respectively. However, one should note that \texttt{Contextual} employs a face recogniser DeepID \cite{sun2014deep}. We have found that a specialised face recogniser improves the recognition quality greatly on the Day split (\S\ref{sec:deepid}).

\subsection{\label{subsec:naeil2}Our final model \texttt{naeil2}}

\texttt{naeil2} is a weighted combination of \texttt{naeil} and $\texttt{h}_{\texttt{deepid}}$ (see \S\ref{sec:naeil+deepid} for details). Observe that by attaching a face recogniser module on \texttt{naeil}, we achieve the best performance on Album, Time, and Day splits. In particular, on the Day split, \texttt{naeil2} makes a 8.85 pp boost over the second best method \texttt{COCO} \cite{liu_2017_coco} (table \ref{tab:test-set-accuracy-four-splits}). On the Original split, \texttt{COCO} performs better (2.36 pp gap), but note that \texttt{COCO} uses more advanced feature representations (GoogleNet and ResNet).

Since \texttt{naeil2} and \texttt{COCO} focus on orthogonal techniques, they can be combined to yield even better performances.

\subsection{Computational cost}

We report computational times for some pipelines in our method. The feature training takes 2-3 days on a single GPU machine. The SVM training takes 42 seconds for $\texttt{h}$ ($4\,096$ dim) and $1\,118$ seconds for \texttt{naeil} on the Original split (581 classes, $6\,443$ samples). Note that this corresponds to a realistic user scenario in a photo sharing service where $\sim\negmedspace500$ identities are known to the user and the average number of photos per identity is $\sim\negmedspace10$.

%Compared to PIPER \cite{Zhang2015CvprPiper}, our framework is computationally efficient in two aspects. First, our system does not need to learn to assign weights for different cues. Second, the PIPER feature has roughly $4096\times108$ dimensions, requiring far more memory and training time than our final system ($4096\times17+1024$ dim).

\section{\label{sec:challenges-analysis}Analysis}

In this section, we provide a deeper analysis of individual cues towards the final performance. In particular, we measure how contributions from individual cues (e.g. face and scene) change when either the system has to generalise across time or head viewpoint. We study the performance as a function of the number of training samples per identity, and examine the distribution of identities according to their recognisability.

\begin{figure}
\centering{}\hspace*{\fill}%
\begin{center}
\includegraphics[width=0.8\columnwidth]{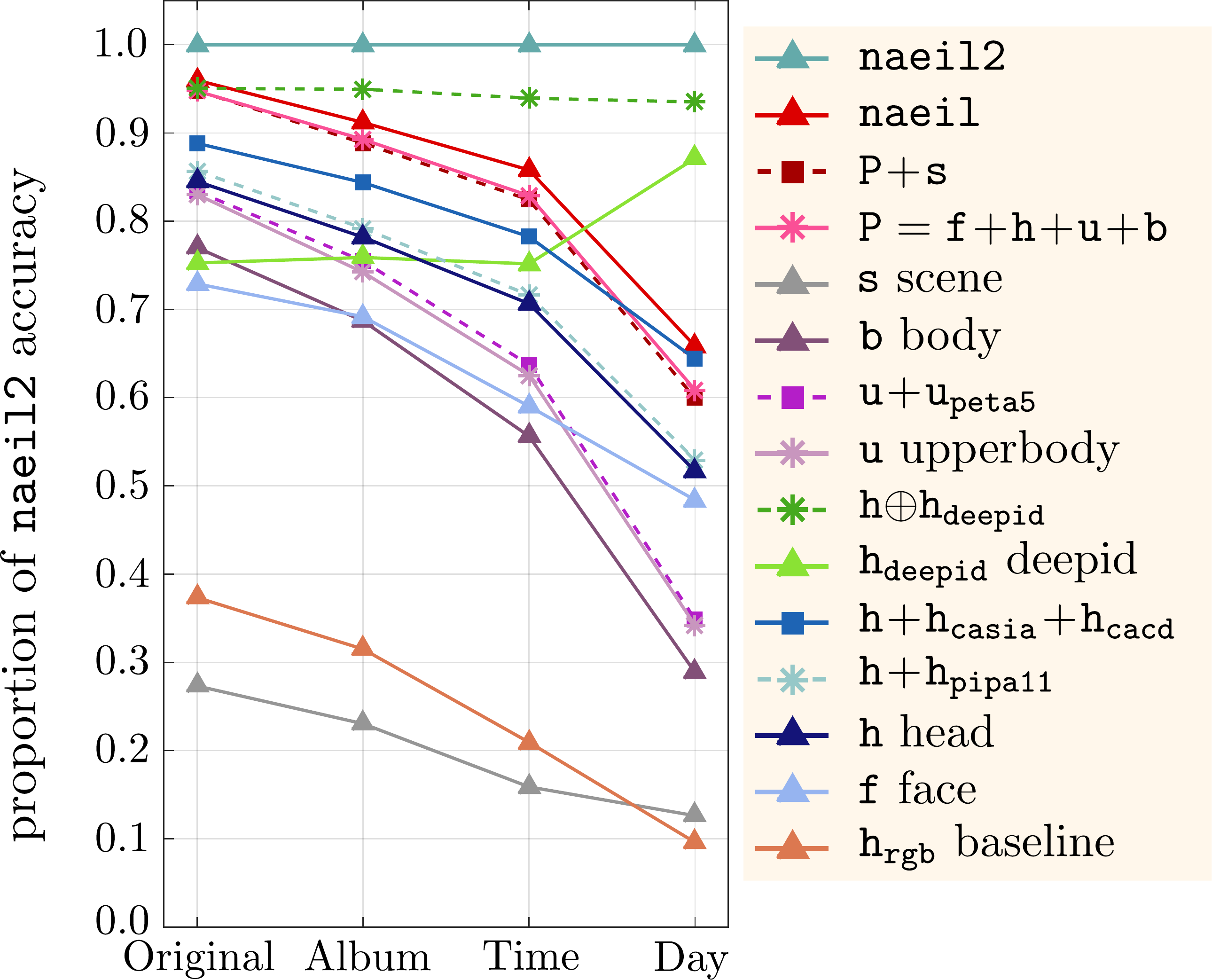}
\vspace{0em}
\caption{\label{fig:splits-accuracy-relative}PIPA \emph{test} set relative accuracy of various methods in the four splits, against the final system \texttt{naeil2}.}
\par\end{center}%
\end{figure}

\subsection{Contribution of individual cues \label{subsec:Importance-of-features}}

We measure the contribution of individual cues towards the final system \texttt{naeil2} (\S\ref{sec:naeil+deepid}) by dividing the accuracy for each intermediate method by the performance of \texttt{naeil2}. We report results in the four splits in order to determine which cues contribute more when there are larger time gap between training and testing samples and vice versa.

\subsubsection*{Results}

See figure \ref{fig:splits-accuracy-relative} for the relative performances in four splits. The cues based more on context (e.g. $\texttt{b}$ and $\texttt{s}$) see greater drop from the Original to Day split, whereas cues focused on face $\texttt{f}$ and head $\texttt{h}$ regions tend to drop less. Intuitively, this is due to the greater changes in clothing and events in the Day split.

On the other hand, $\mbox{\ensuremath{\texttt{h}}}_{\texttt{deepid}}$ increases in its relative contribution from Original to Day split, nearly explaining 90\% of \texttt{naeil2} in the Day split. $\mbox{\ensuremath{\texttt{h}}}_{\texttt{deepid}}$ provides valuable invariant face feature especially when the time gap is great. However, on the Original split $\mbox{\ensuremath{\texttt{h}}}_{\texttt{deepid}}$ only reaches about 75\% of \texttt{naeil2}. Head orientation robust $\texttt{naeil}$ should be added to attain the best performance. 

\subsubsection*{Conclusion}

Cues involving context are stronger in the Original split; cues around face, especially the $\mbox{\ensuremath{\texttt{h}}}_{\texttt{deepid}}$, are robust in the Day split. Combining both types of cues yields the best performance over all considered time/appearance changes.

\subsection{\label{subsec:Head-orientation-analysis}Performance by viewpoint}

We study the impact of test instance viewpoint on the proposed systems. Cues relying on face are less likely to be robust to occluded faces, while body or context cues will be robust against viewpoint changes. We measure the performance of models on the head orientation partitions defined by a DPM head detector (see \S\ref{subsec:face-detection}): frontal $\texttt{FR}$, non-frontal $\texttt{NFR}$, and
no face detected $\texttt{NFD}$. $\texttt{NFD}$ subset is a proxy for back-view and occluded-face instances. 

\subsubsection*{Results}

\begin{figure}
\begin{centering}
\begin{center}
\includegraphics[width=0.8\columnwidth]{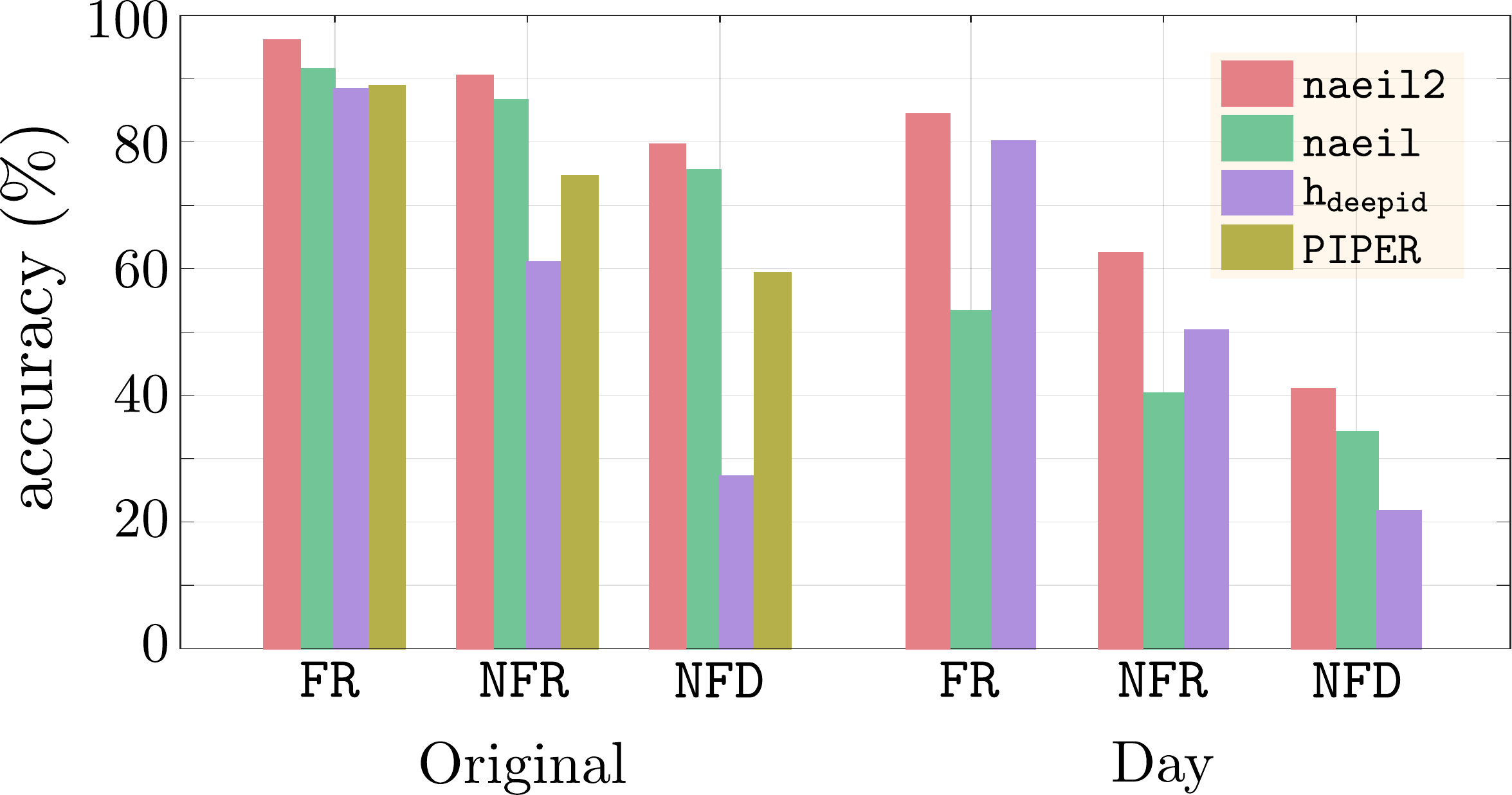}
\caption{\label{fig:threeway-bar}PIPA \emph{test} set accuracy of methods on the frontal ($\texttt{FR}$), non-frontal ($\texttt{NFR}$), and no face detected ($\texttt{NFD}$) subsets. Left: Original split, right: Day split.}
\par\end{center}
\end{centering}
\end{figure}

Figure \ref{fig:threeway-bar} shows the accuracy of methods on the three head orientation subsets for the Original and Day splits. All the considered methods show worse performance from frontal $\texttt{FR}$ to non-frontal $\texttt{NFR}$ and no face detected $\texttt{NFD}$ subsets. However, in the Original split, \texttt{naeil2} still robustly predicts the identities even for the \texttt{NFD} subset ($\sim\negmedspace 80\%$ accuracy). On the Day split, \texttt{naeil2} also do struggle on the \texttt{NFD} subset ($\sim\negmedspace 20\%$ accuracy). Recognition of  \texttt{NFD} instances under the Day split constitutes the main remaining challenge of person recognition.

\begin{figure}
\begin{centering}
\begin{center}
\includegraphics[width=0.8\columnwidth]{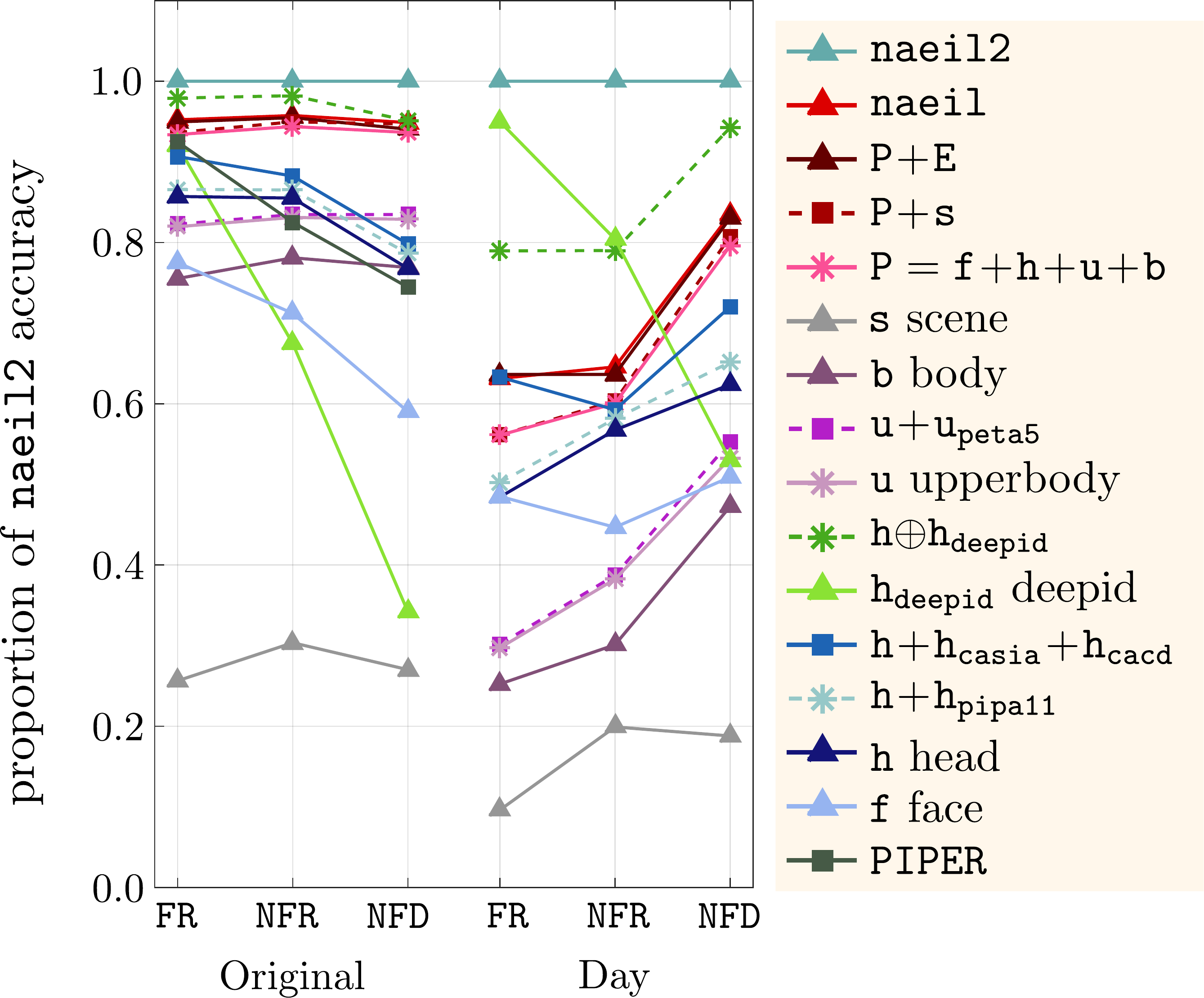}
\caption{\label{fig:threeway-relative}PIPA \emph{test} set relative accuracy of frontal ($\texttt{FR}$), non-frontal ($\texttt{NFR}$), and non-detection ($\texttt{NDET}$) head orientations, relative to the final model \texttt{naeil2}. Left: Original split, right: Day split. }
\par\end{center}
\end{centering}
\end{figure}

In order to measure contributions from individual cues in different head orientation subsets, we report the relative performance against the final model \texttt{naeil2} in figure \ref{fig:threeway-relative}. The results are reported on the Original and Day splits. Generally, cues based on more context (e.g. $\texttt{b}$ and $\texttt{s}$) are more robust when face is not visible than the face specific cues (e.g. $\texttt{f}$ and $\texttt{h}$). Note that $\mbox{\ensuremath{\texttt{h}}}_{\texttt{deepid}}$ performance drops significantly in $\texttt{NDET}$, while $\texttt{naeil}$ generally improves its relative performance in harder viewpoints. \texttt{naeil2} envelops the performance of the individual cues in all orientation subsets.

\subsubsection*{Conclusion}

$\texttt{naeil}$ is more viewpoint robust than $\mbox{\ensuremath{\texttt{h}}}_{\texttt{deepid}}$, making a contrast to the time-robustness analysis (\S\ref{subsec:Importance-of-features}). The combined model \texttt{naeil2} takes the best of both worlds. The remaining challenge for person recognition lies on the no face detected \texttt{NFD} instances under the Day split. Perhaps image or social media metadata could be utilised (e.g. camera statistics, time and GPS location, social media friendship graph). 

\begin{figure*}
\begin{centering}
\begin{center}
\hspace*{\fill}\includegraphics[width=0.8\columnwidth]{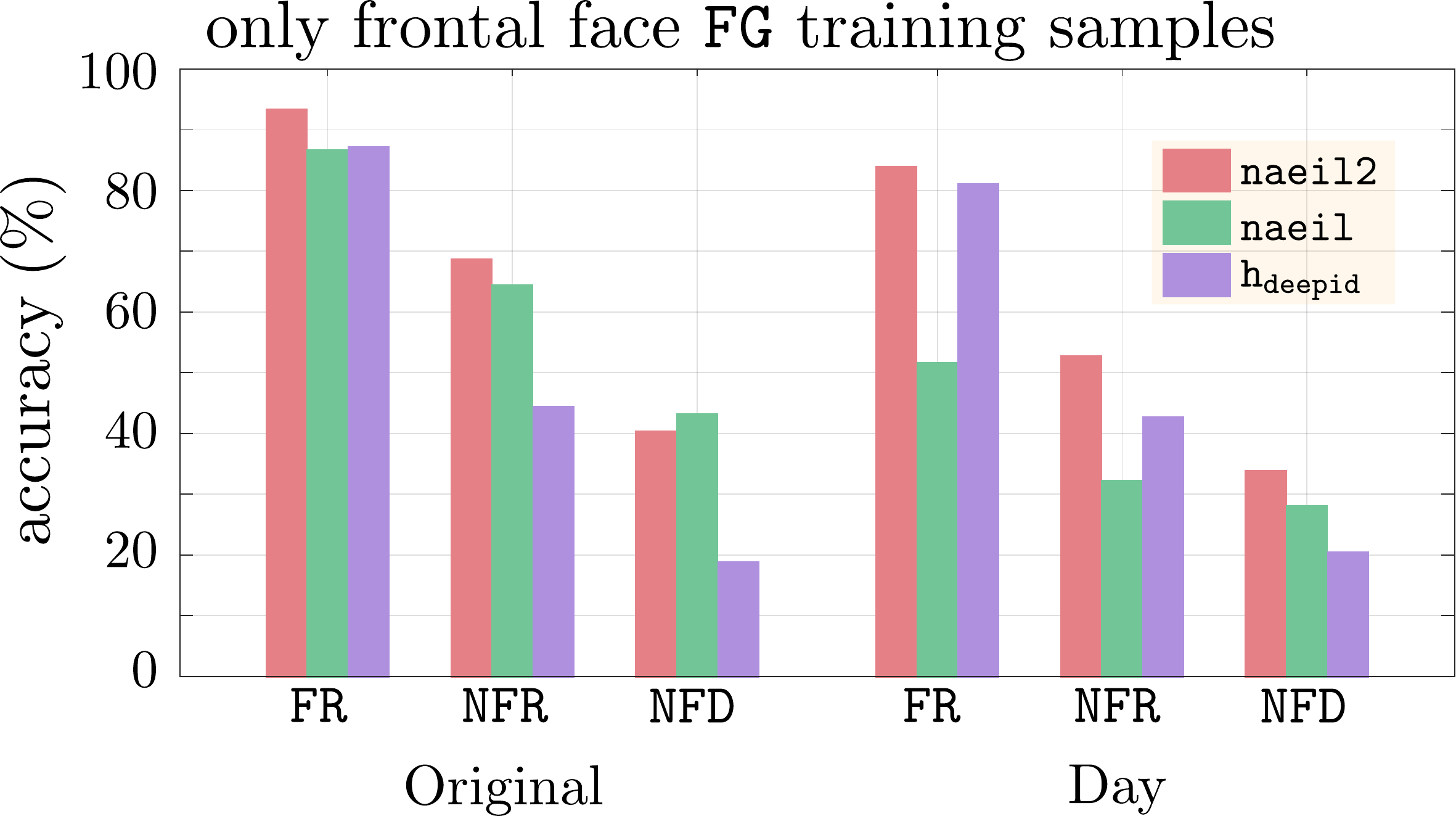}\hspace*{\fill}\includegraphics[width=0.8\columnwidth]{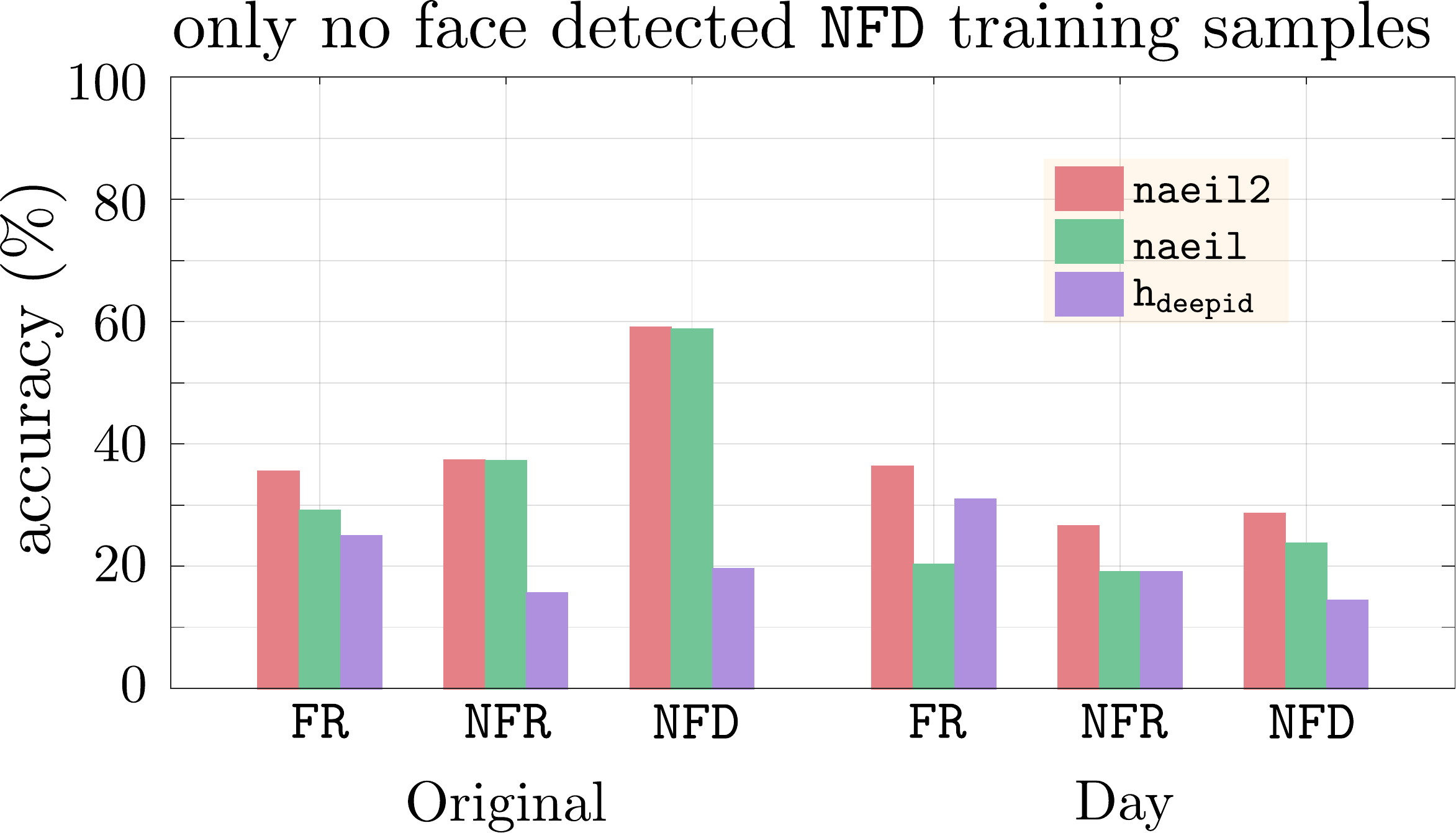}\hspace*{\fill}
\caption{\label{fig:threeway-svm}PIPA \emph{test} set performance when the identity classifier (SVM) is only trained on either frontal ($\texttt{FR}$, left) or no face detected ($\texttt{NFD}$, right) subset. Related scenario: a robot has only seen frontal views of people; who is this person shown from the back view? }
\par\end{center}
\end{centering}
\end{figure*}

\subsection{Generalisation across viewpoints\label{subsec:SVM-orientation}}

Here, we investigate the viewpoint generalisability of our models. For example, we challenge the system to identify a person from the back, having only shown frontal face samples.

\subsubsection*{Results}

Figure \ref{fig:threeway-svm} shows the accuracies of the methods, when they are trained either only on the frontal subset $\texttt{FR}$ (left plot) or only on the no face detected subset $\texttt{NFD}$ (right plot). When trained on $\texttt{FR}$, \texttt{naeil2} has difficulties generalising to the $\texttt{NFD}$ subset ($\texttt{FR}$ versus $\texttt{NFD}$ performance is $\sim\negmedspace95\%$ to $\sim\negmedspace40\%$ in Original; $\sim\negmedspace85\%$ to $\sim\negmedspace35\%$ in Day). However, the absolute performance is still far above the random chance (see \S\ref{subsec:face-rgb-baseline}), indicating that the learned identity representations are to a certain degree generalisable. The \texttt{naeil} features are more robust in this case than $\mbox{\ensuremath{\texttt{h}}}_{\texttt{deepid}}$, with less dramatic drop from \texttt{FR} to \texttt{NFD}.
 
When no face is given during training (training on \texttt{NFD} subset), identities are much harder to learn in general. The recognition performance is low even for no-generalisation case: $\sim\negmedspace60\%$ and $\sim\negmedspace30\%$ for Original and Day, respectively, when trained and tested on \texttt{NFD}.

\subsubsection*{Conclusion}

$\texttt{naeil2}$ does generalise marginally across viewpoints, largely attributing to the \texttt{naeil} features. It seems quite hard to learn identity specific features (either generalisable or not) from back-views or occluded faces (\texttt{NFD}).

\subsection{Viewpoint distribution does not matter for feature training\label{subsec:Feature-learning-orientation}}

We examine the effect of the ratio of head orientations in the feature training set on the quality of the head feature $\ensuremath{\texttt{h}}$. We fix the number of training examples that consists only of frontal \texttt{FR} and non-frontal faces \texttt{NFR}, while varying their ratio.

One would hypothesize that the maximal viewpoint robustness of the feature is achieved at a balanced mixture of \texttt{FR} and \texttt{NFR} for each person; also that $\ensuremath{\texttt{h}}$ trained with $\texttt{FR}$ ($\texttt{NFR}$) subset is relatively strong at predicting $\texttt{FR}$ ($\texttt{NFR}$) subset (respectively). 

\subsubsection*{Results}

\begin{figure}
\begin{centering}
\hspace*{\fill}\includegraphics[width=0.5\columnwidth]{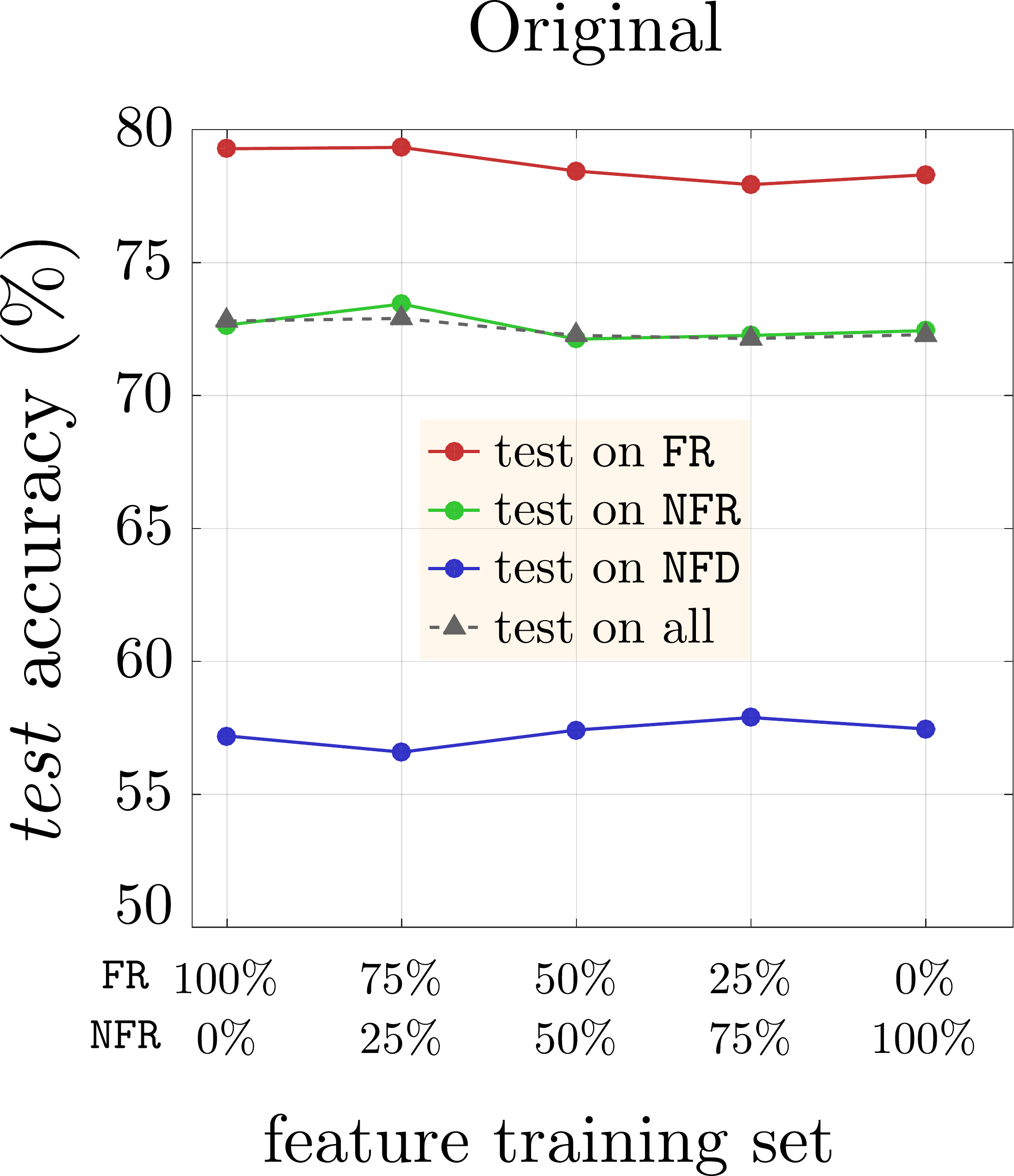}\hspace*{\fill}\includegraphics[width=0.5\columnwidth]{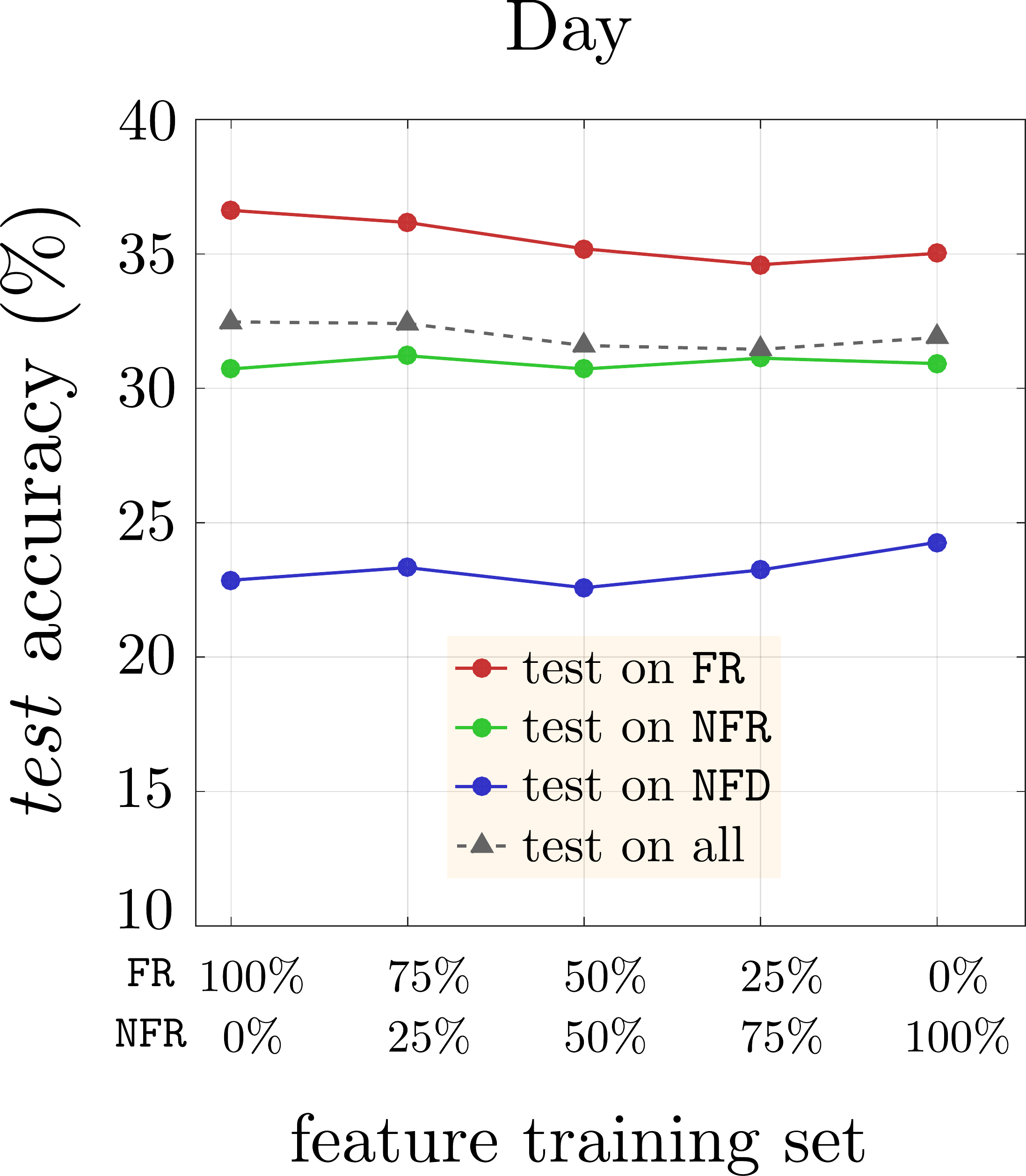}\hspace*{\fill}\vspace{0.5em}
\par\end{centering}
\caption{\label{fig:threeway-feat} Train the feature $\ensuremath{\texttt{h}}$ with different mixtures of frontal $\texttt{FR}$ and non-frontal $\texttt{NFR}$ heads. The viewpoint wise performance is shown for the Original (left) and Day (right) splits.}
\end{figure}

Figure \ref{fig:threeway-feat} shows the performance of $\ensuremath{\texttt{h}}$ trained with various $\texttt{FR}$ to $\texttt{NFR}$ ratios on \texttt{FR}, \texttt{NFR}, and \texttt{NFD} subsets. Contrary to the hypothesis, changing the distribution of head orientations in the feature training has $<3\%$ effect on their performances across all viewpoint subsets in both Original and Day splits. 

\subsubsection*{Conclusion}

No extra care is needed to control the distribution of head orientations in the feature training set to improve the head feature $\texttt{h}$. Features on larger image regions (e.g. \texttt{u} and \texttt{b}) are expected to be even less affected by the viewpoint distribution.

\subsection{Input resolution \label{subsec:Analysis-of-remaining-factors}}

This section provides analysis on the impact of input resolution. We aim to identify methods that are robust in different range of resolutions.

\subsubsection*{Results}

\begin{figure}
\centering{}\hspace*{\fill}%
\begin{center}
\includegraphics[width=0.8\columnwidth]{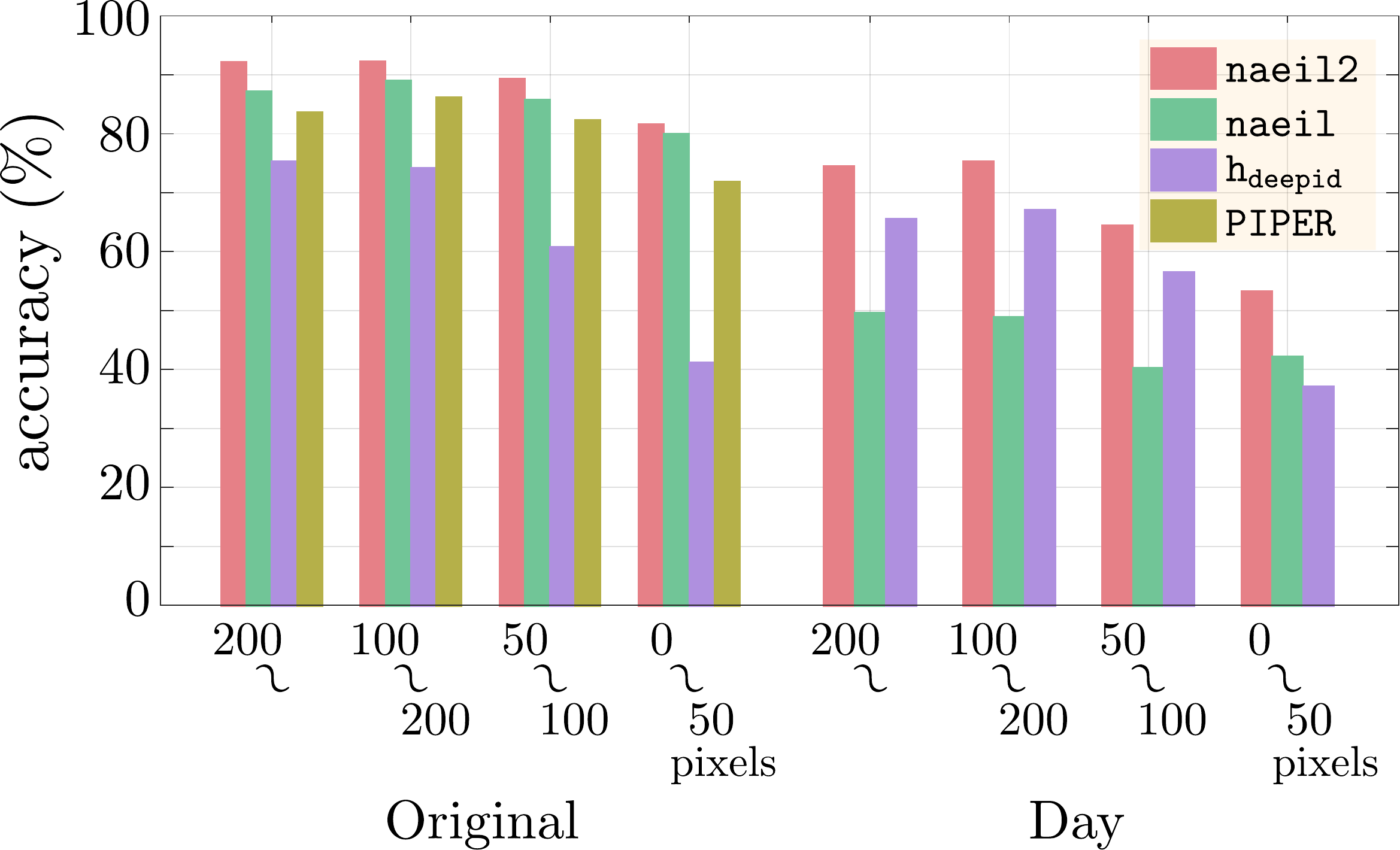}
\vspace{0em}
\caption{\label{fig:remaining_factors}PIPA \emph{test} set accuracy of systems at different levels of input resolution. Resolution is measured in terms of the head height (pixels).}
\par\end{center}%
\end{figure}

Figure \ref{fig:remaining_factors} shows the performance with respect to the input resolution (head height in pixels). The final model \texttt{naeil2} is robust against low input resolutions, reaching $\sim\negmedspace80\%$ even for instances with $<50$ pixel heads on Original split. On the day split, \texttt{naeil2} is less robust on low resolution examples ($\sim\negmedspace55\%$).

Component-wise, note that $\texttt{naeil}$ performance is nearly invariant to the resolution level. $\texttt{naeil}$ tends to be more robust for low resolution input than the $\texttt{h}_{\texttt{deepid}}$ as it is based on body and context features and do not need high resolution faces. 

\subsubsection*{Conclusion}

For low resolution input $\texttt{naeil}$ should be exploited, while for high resolution input $\texttt{h}_{\texttt{deepid}}$ should be exploited. If unsure, \texttt{naeil2} is a good choice -- it envelops the performance of both in all resolution levels.

\subsection{Number of training samples\label{subsec:Importance-of-training}}

We are interested in two questions: (1) if we had more samples per identity, would person recognition be solved with the current method? (2) how many examples per identity are enough to gather substantial amount of information about a person? To investigate the questions, we measure the performance of methods at different number of training samples per identity. We perform 10 independent runs per data point with  fixed number of training examples per identity (subset is uniformly sampled at each run).

\subsubsection*{Results}

\begin{figure}
\begin{centering}
\hspace*{\fill}\includegraphics[width=0.8\columnwidth]{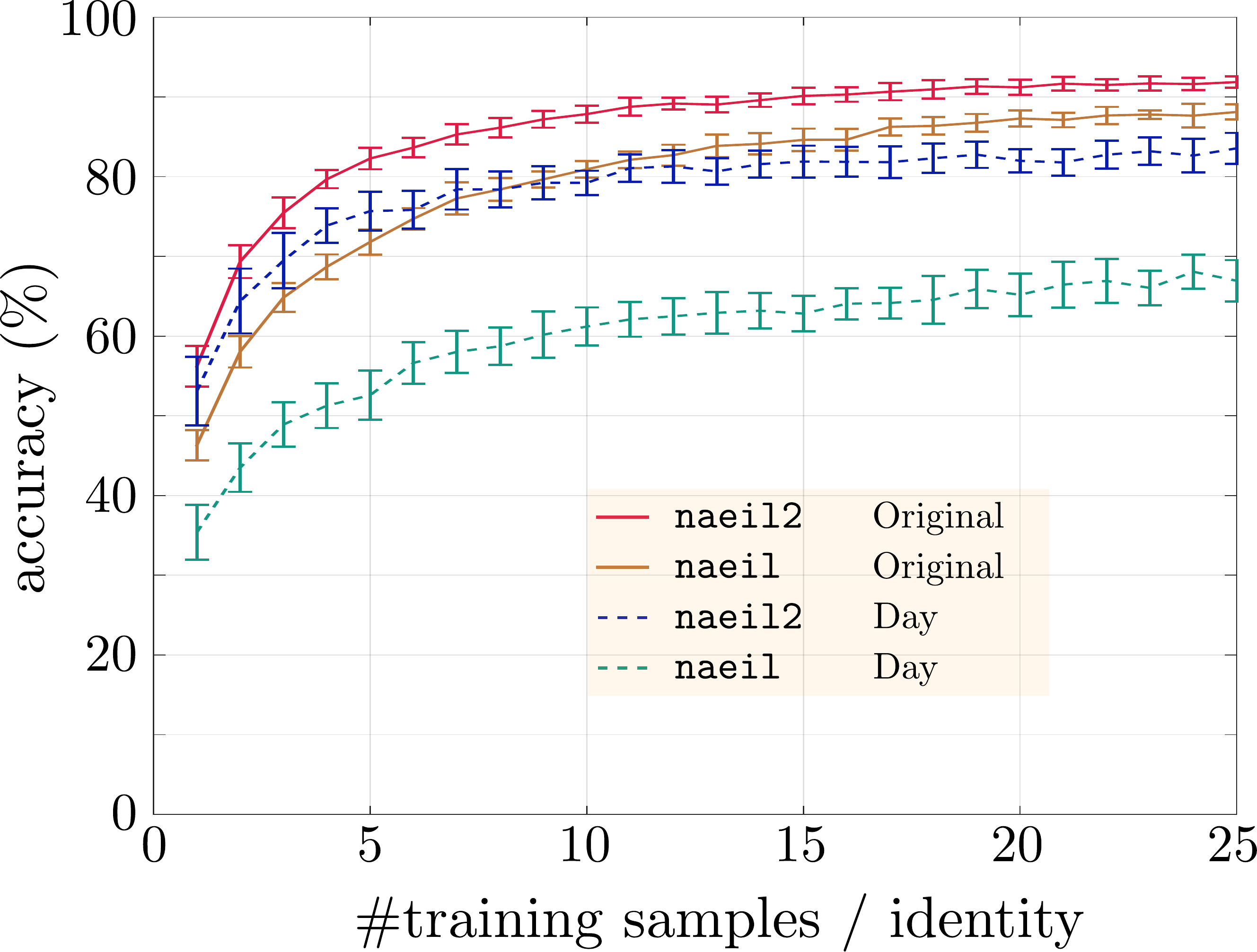}\hspace*{\fill}
\par\end{centering}
\begin{centering}
\vspace{0em}
\par\end{centering}
\caption{\label{fig:numtrain-accuracy}Recognition accuracy at different number of training samples per identity. Error bars indicate $\pm 1$ standard deviation from the mean.}
\end{figure}

Figure \ref{fig:numtrain-accuracy} shows the trend of recognition performances of methods with respect to different levels of training sample size. \texttt{naeil2} saturates after $10\sim15$ training examples per person in Original and Day splits, reaching $\sim\negmedspace92\%$ and $\sim\negmedspace83\%$, respectively, at $25$ examples per identity. At the lower end, we observe that 1 example per identity is already enough to recognise a person far above the chance level ($\sim\negmedspace67\%$ and $\sim\negmedspace35\%$ on Original and Day, respectively).

\subsubsection*{Conclusion}

Adding a few times more examples per person will not push the performance to 100\%. Methodological advances are required to fully solve the problem. On the other hand, the methods already collect substantial amount of identity information only from single sample per person (far above chance level). 

\subsection{Distribution of per-identity accuracy \label{subsec:per-id-accuracy}}

Finally, we study how much proportion of the identities are easy to recognise and how many are hopeless. We study this by computing the distribution of identities according to their per-identity recognition accuracies.

\subsubsection*{Results}

\begin{figure}
\begin{centering}
\hspace*{\fill}\includegraphics[width=0.5\columnwidth]{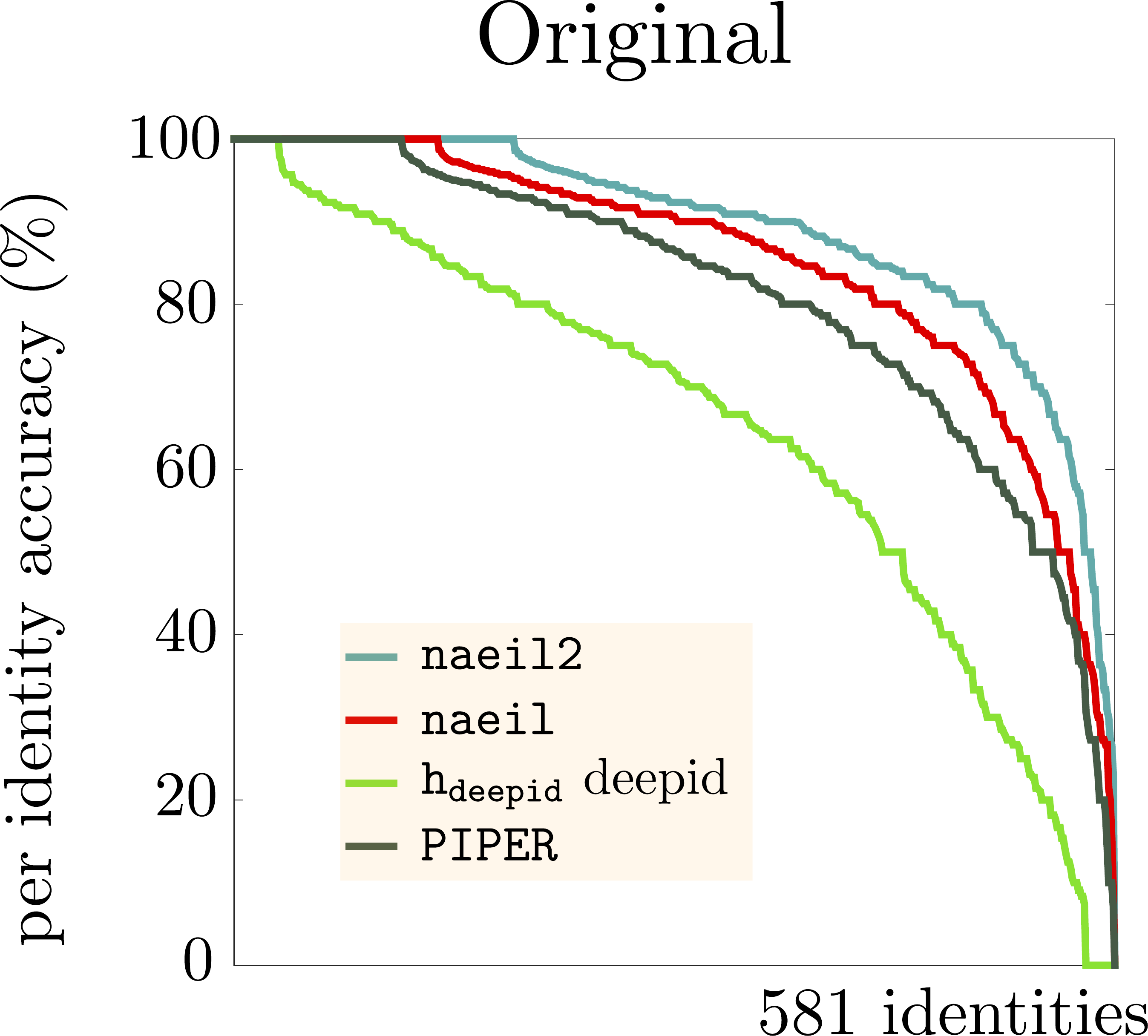}\hspace*{\fill}\includegraphics[width=0.5\columnwidth]{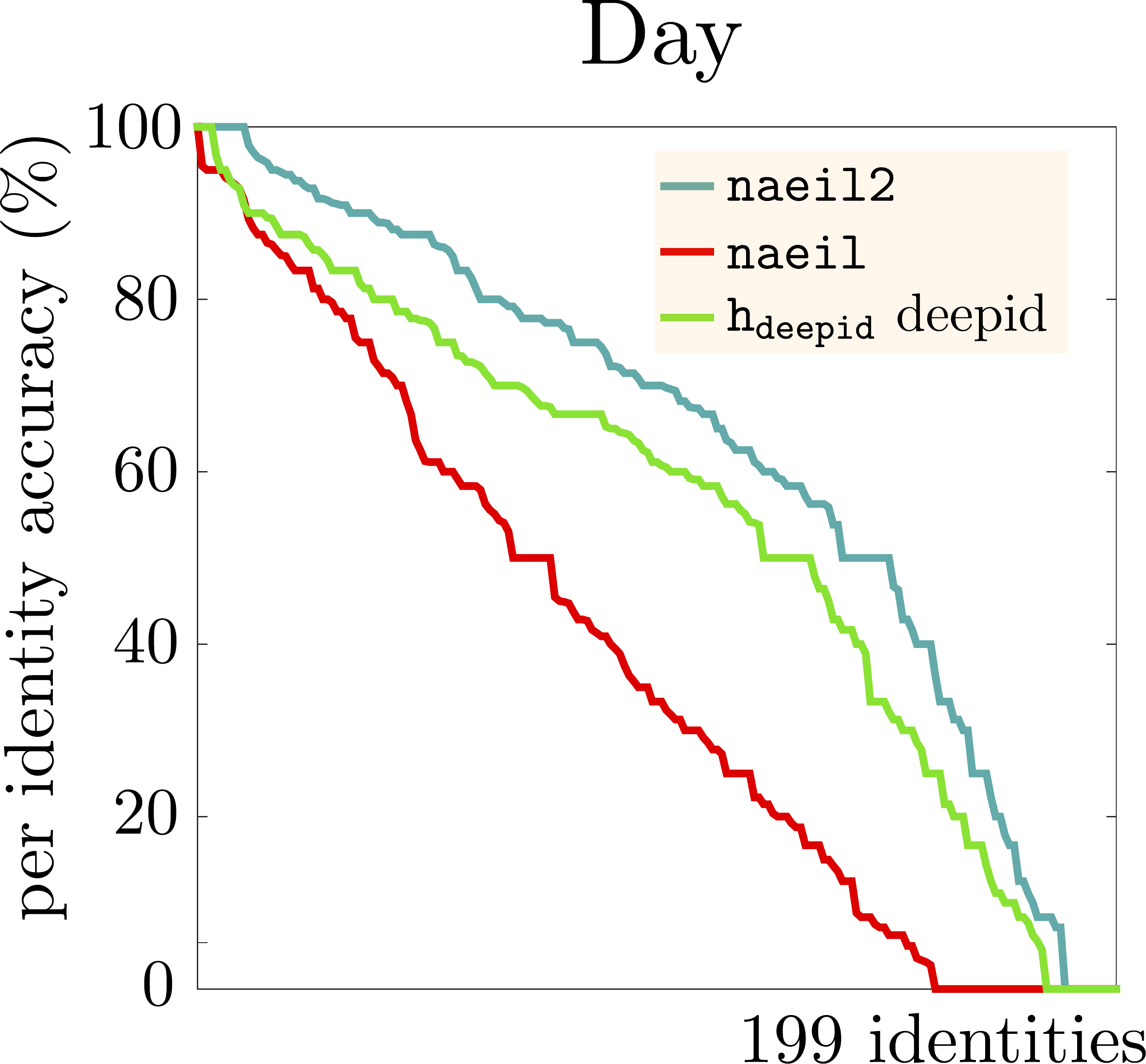}\hspace*{\fill}
\par\end{centering}
\begin{centering}
\vspace{0em}
\par\end{centering}
\caption{\label{fig:per-id-accuracy}Per identity accuracy of on the Original and Day splits. The identities are sorted according to the per identity accuracy for each method separately.}
\end{figure}

Figure \ref{fig:per-id-accuracy} shows the per identity accuracy for each identity in a descending order for each considered method. On the Original split, \texttt{naeil2} gives $100\%$ accuracy for $185$ out of the $581$ test identities, whereas there was only one identity where the method totally fails. On the other hand, on the Day split there are $11$ out of the $199$ test identities for whom \texttt{naeil2} achieves $100\%$ accuracy and $12$ identities with zero accuracy. In particular, \texttt{naeil2} greatly improves the per-identity accuracy distribution over $\texttt{naeil}$, which gives zero prediction for $40$ identities.

\subsubsection*{Conclusion}

In the Original split, \texttt{naeil2} is doing well on many of the identities already. In the Day split, the $\texttt{h}_{\texttt{deepid}}$ feature has greatly improved the per-identity performances, but \texttt{naeil2} still misses some identities. It is left as future work to focus on the hard identities.

\section{\label{sec:open-world}Open-world recognition}

So far, we have focused on the scenario where the test instances are always from a closed world of gallery identities. However, for example when person detectors are used to localise instances, as opposed to head box annotations, the detected person may not be one of the gallery set. One may wonder how our person recognisers would perform when the test instance could be an unseen identity.

In this section, we study the task of ``open-world person recognition''. The test identity may be either from a gallery set (training identities) or from a background set (unseen identities). We consider the scenario where test instances are given by a face detector \cite{Mathias2014Eccv} while the training instance locations have been annotated by humans.

Key challenge for our recognition system is to tell apart gallery identities from background faces, while simultaneously classifying the gallery identities. Obtained from a detector, the background faces may contain any person in the crowd or even non-faces. We will introduce a simple modification of our recognition systems' test time algorithm to let them further make the gallery versus background prediction. We will then discuss the relevant metrics for our systems' open-world performances.

\subsection{Method}

At test time, body part crops are inferred from the detected face region ($\texttt{f}$). First, $\texttt{h}$ is regressed from $\texttt{f}$, using the PIPA \emph{train} set statistics on the scaling and displacement transformation from \texttt{f} to \texttt{h}. All the other regions ($\texttt{u}$, $\texttt{b}$, $\texttt{s}$) are computed based on $\texttt{h}$ in the same way as in \S\textcolor{red}{3.2} of main paper.

To measure if the inferred head region \texttt{h} is sound and compatible with the models trained on \texttt{h} (as well as $\texttt{u}$ and $\texttt{b}$), we train the head model \texttt{h} on head annotations and test on the heads inferred from face detections. The recognition performance is $87.74\%$, while when trained and tested on the head annotations, the performance is $89.85\%$. We see a small drop, but not significant -- the inferred regions to be largely compatible. %($\texttt{FR}\cup\texttt{NFR}$) subset

The gallery-background identity detection is done by thresholding the final SVM score output. Given a recognition system and test instance $x$, let $\mathcal{S}_{k}\left(x\right)$ be the SVM score for identity $k$. Then, we apply a thresholding parameter $\tau>0$ to predict background if $\underset{k}{\max}\,\,\mathcal{S}_{k}\left(x\right)<\tau$, and predict the argmax gallery identity otherwise.

\subsection{Evaluation metric}

\begin{figure}
\begin{centering}
\hspace*{\fill}\includegraphics[width=0.8\columnwidth]{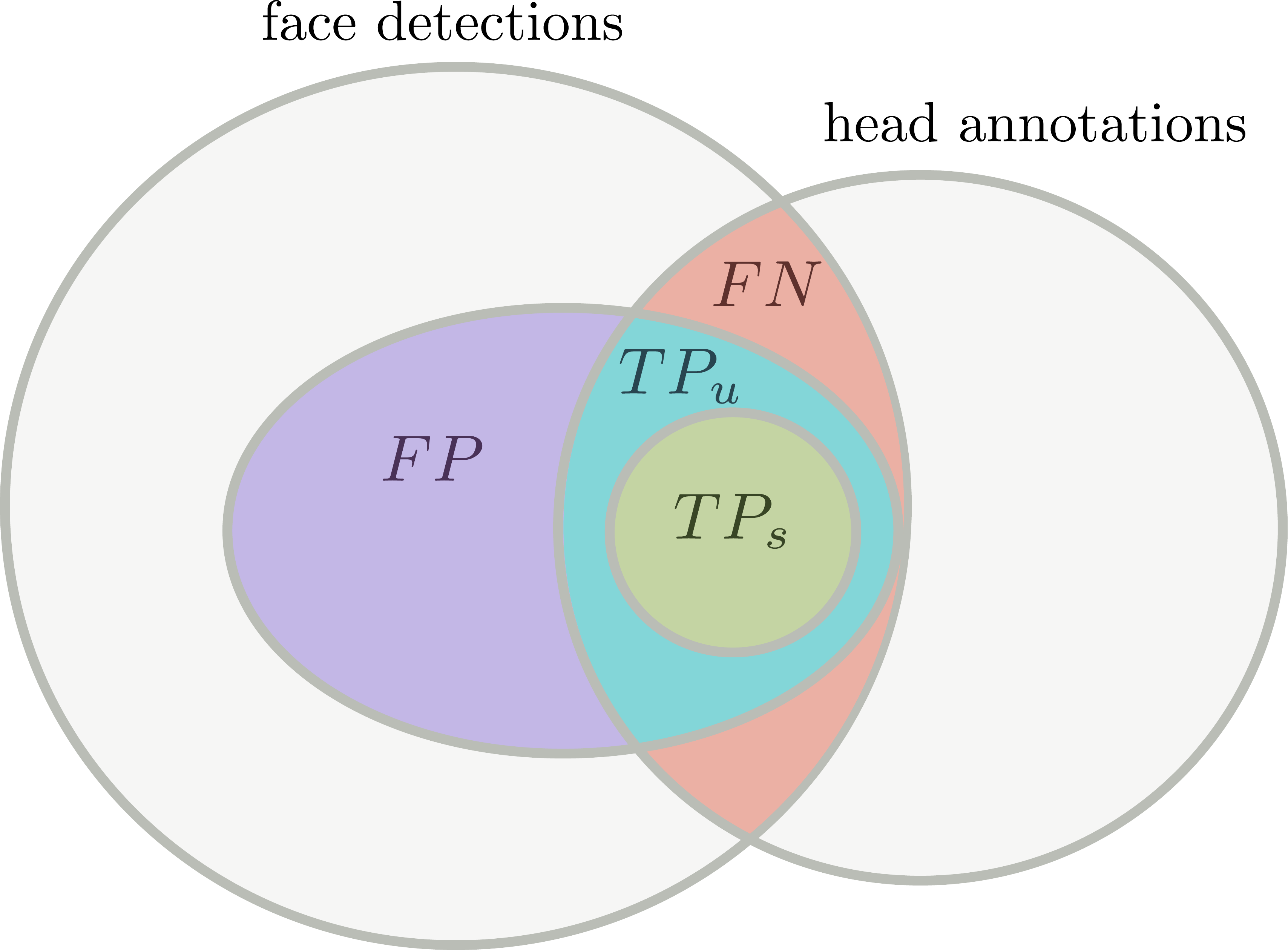}\hspace*{\fill}
\par\end{centering}
\caption{\label{fig:open-metric}Diagram of various subsets generated by a person recognition system in an open world setting (cf. Figure \textcolor{red}{2} of main paper). $TP_s$: sound true positive, $TP_u$: unsound true positive, $FP$: false positive, $FN$: false negative. See text for the definitions.}
\end{figure}

The evaluation metric should measure two aspects simultaneously: (1) ability to tell apart background identities, (2) ability to classify gallery identities. We first introduce a few terms to help defining the metrics. Refer to figure \ref{fig:open-metric} for a visualisation. We say a detected test instance $x$ is a ``foreground prediction'' if $\underset{k}{\max}\,\,\mathcal{S}_{k}\left(x\right)\ge\tau$. A foreground prediction is either a true positive ($TP$) or a false positive ($FP$), depending on whether $x$ is a gallery identity or not. If $x$ is a $TP$, it is either a sound true positive $TP_s$ or an unsound true positive $TP_u$, depending on the classification result $\underset{k}{\arg\max}\,\,\mathcal{S}_{k}\left(x\right)$. A false negative ($FN$) is incurred if a gallery identity is predicted
as background. 

We first measure the system's ability to screen background identities while at the same time classifying the gallery identities. The \textbf{recognition recall (RR)} at threshold $\tau$ is defined as follows
\begin{equation}
\label{eq:RR}
\mathrm{RR}(\tau)=\frac{\left|TP_s\right|}{\left|\mbox{face det.}\cap\mbox{head anno.}\right|}=\frac{\left|TP_s\right|}{\left|TP\cup FN\right|}.
\end{equation}
To factor out the performance of face detection, we constrain our evaluation to the intersection between face detections and head annotations (the denominator $TP\cup FN$). Note that the metric is a decreasing function of $\tau$, and when $\tau\rightarrow-\infty$ the corresponding system is operating under the closed world assumption.

The system enjoys high RR when $\tau$ is decreased, but the system then predicts many background cases as foreground ($FP$). To quantify the trade-off we introduce a second metric: 
\textbf{false positive per image (FPPI)}. Given a threshold $\tau>0$, FPPI is defined as
\begin{equation}
\label{eq:FPPI}
\mathrm{FPPI}(\tau)=\frac{\left|FP\right|}{\left|\mbox{images}\right|},
\end{equation}
measuring how many wrong foreground predictions the system makes per image. It is also a decreasing function of $\tau$. When $\tau\rightarrow\infty$, the FPPI attains zero.

\subsection{Results}

\begin{figure*}
\begin{centering}
\hspace*{\fill}\includegraphics[width=0.8\columnwidth]{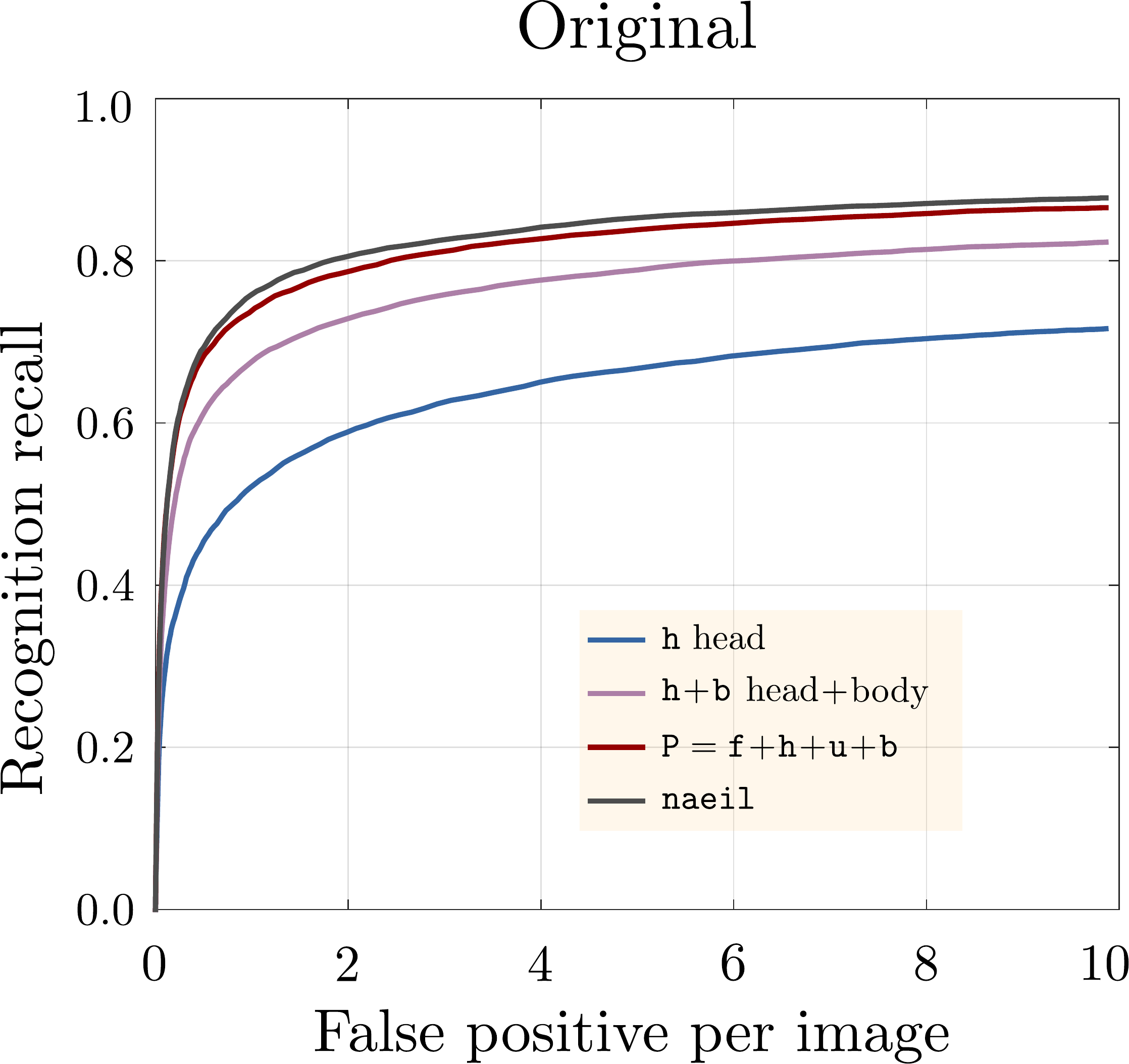}\hspace*{\fill}\includegraphics[width=0.8\columnwidth]{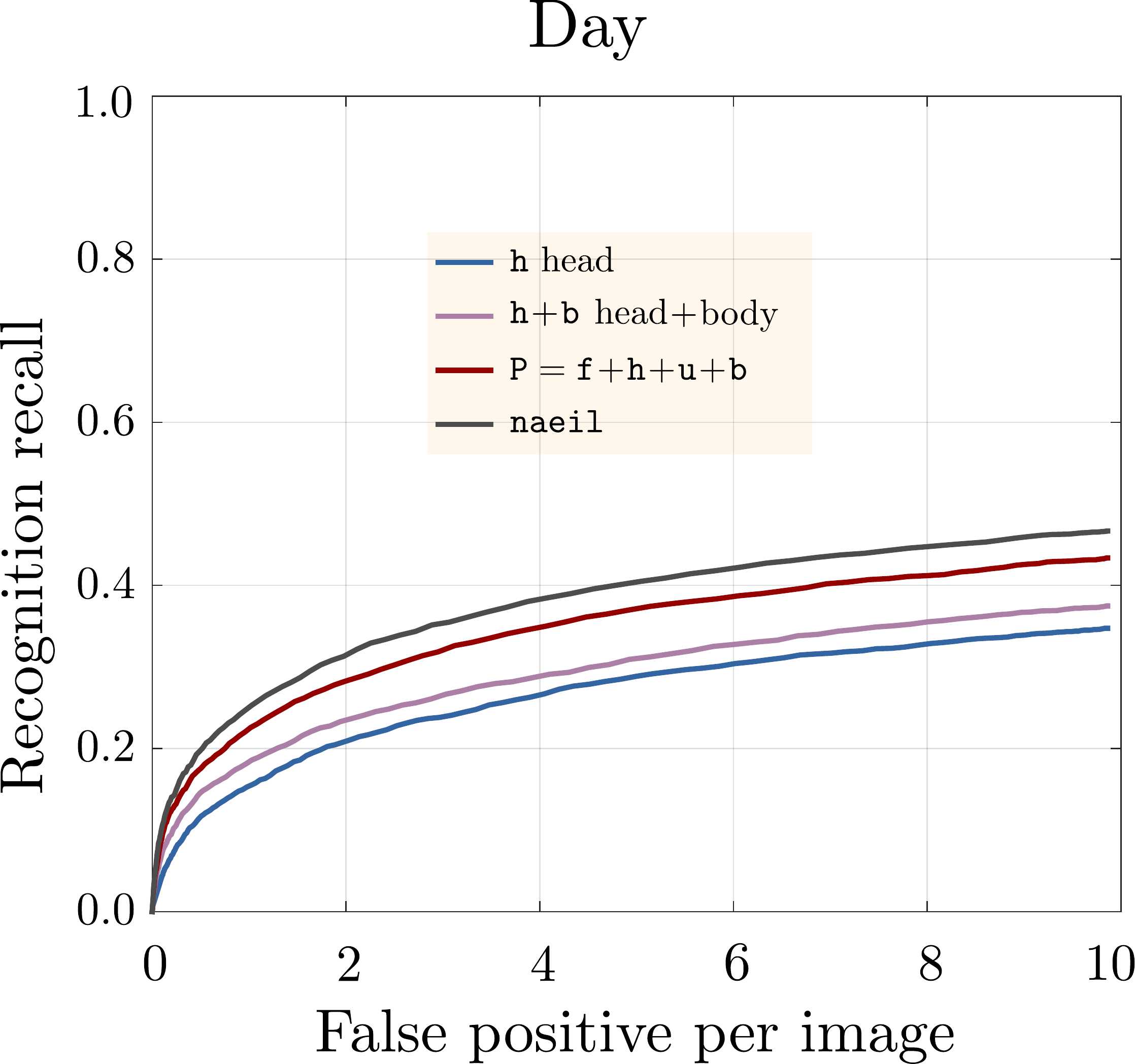}\hspace*{\fill}
\par\end{centering}
\caption{\label{fig:open-world-results}Recognition recall (RR) versus false positive per image (FPPI) of our recognition systems in the open world setting. Curves are parametrised by $\tau$ -- see text for details.}
\end{figure*}

Figure \ref{fig:open-world-results} shows the recognition rate (RR) versus false positive per image (FPPI) curves parametrised by $\tau$. As $\tau\to\infty$, $RR(\tau)$ approaches the close world performance on the face detected subset ($\texttt{FR}\cup\texttt{NFR}$): $87.74\%$ (Original) and $46.67\%$ (Day) for $\texttt{naeil}$. In the open-world case, for example when the system makes one FPPI, the recognition recall for $\texttt{naeil}$ is $76.25\%$ (Original) and $25.29\%$ (Day). Transitioning from the open world to close world, we see quite some drop, but one should note that the set of background face detections is more than $7\times$ greater than the foreground faces.

Note that the DeepID2+ \cite{Sun2014ArxivDeepId2plus} is not a public method, and so we cannot compute $\mbox{\ensuremath{\texttt{h}}}_{\texttt{deepid}}$ features ourselves; we have not included the $\mbox{\ensuremath{\texttt{h}}}_{\texttt{deepid}}$ or \texttt{naeil2} results in this section.

\subsection{Conclusion}

Although performance is not ideal, a simple SVM score thresholding scheme can make our systems work in the open world recognition scenario.

\section{\label{sec:Conclusion}Conclusion}

We have analysed the problem of person recognition in social media photos where people may appear with occluded faces, in diverse poses, and in various social events. We have investigated efficacy of various cues, including the face recogniser DeepID2+\cite{Sun2014ArxivDeepId2plus}, and their time and head viewpoint generalisability. For better analysis, we have contributed additional splits on PIPA \cite{Zhang2015CvprPiper} that simulate different amount of time gap between training and testing samples.

We have made four major conclusions. (1) Cues based on face and head are robust across time (\S\ref{subsec:Importance-of-features}). (2) Cues based on context are robust across head viewpoints (\S\ref{subsec:Head-orientation-analysis}). (3) The final model \texttt{naeil2}, a combination of face and context cues, is robust across both time and viewpoint and achieves a $\sim\negmedspace9$ pp improvement over a recent state of the art on the challenging Day split (\S\ref{subsec:naeil2}). (4) Better convnet architectures and face recognisers will improve the performance of the \texttt{naeil} and \texttt{naeil2} frameworks in the future \S\ref{subsec:naeil2}).

The remaining challenges are mainly the large time gap and occluded face scenarios (\S\ref{subsec:Head-orientation-analysis}). One possible direction is to exploit non-visual cues like GPS and time metadata, camera parameters, or social media album/friendship graphs. Code and data are publicly available at \url{https://goo.gl/DKuhlY}.

%\bibliographystyle{ieee}
%\bibliography{refs.bib}

% or


%\section{}

\ifCLASSOPTIONcompsoc
  \section*{Acknowledgments}
\else
  \section*{Acknowledgment}
\fi

This research was supported by the German Research Foundation (DFG CRC 1223).

\ifCLASSOPTIONcaptionsoff
  \newpage
\fi



\bibliographystyle{IEEEtran}
\bibliography{refs.bib}
%\begin{thebibliography}{1}


%\end{thebibliography}


\begin{IEEEbiography}[{\includegraphics[width=1in,height=1.25in,clip,keepaspectratio]{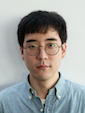}}]{Seong Joon Oh}
received the master's degree in mathematics from the University of Cambridge in 2014 and his PhD in computer vision from the Max Planck Institute for Informatics in 2018. He currently works as a research scientist at LINE Plus Corporation, South Korea. His research interests are computer vision, machine learning, security, and privacy.

\end{IEEEbiography}

\begin{IEEEbiography}[{\includegraphics[width=1in,height=1.25in,clip,keepaspectratio]{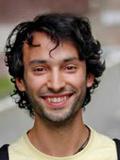}}]{Rodrigo Benenson}
received his electronics engineering diploma from UTFSM, Valparaiso in 2004 and his PhD in robotics from INRIA/Mines
Paristech in 2008. He was a postdoctoral associate with the KULeuven VISICS team from 2008 to 2012 and with the Max Planck Institute for
Informatics, Saarbr\"{u}cken until 2017. He currently works as a research scientist at Google, Z\"{u}rich.
His main interests in computer vision are weakly supervised learning and mobile robotics.

\end{IEEEbiography}

\begin{IEEEbiography}[{\includegraphics[width=1in,height=1.25in,clip,keepaspectratio,trim={50 0 50 0}]{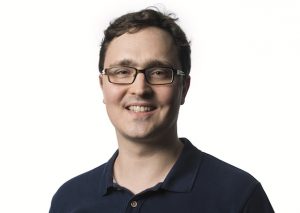}}]{Mario Fritz}
is faculty member at the CISPA Helmholtz Center i.G. He is working at the intersection of machine learning and computer vision with privacy and security. He did his postdoc at the International Computer Science Institute and UC Berkeley on a Feodor Lynen Research Fellowship of the Alexander von Humboldt Foundation. He has received funding from Intel, Google and a collaborative research center on ``Methods and Tools for Understanding and Controlling Privacy''.  He has served as an area chair for major vision conferences (ICCV, ECCV, ACCV, BMVC) and is associate editor for TPAMI. 

\end{IEEEbiography}

\begin{IEEEbiography}[{\includegraphics[width=1in,height=1.25in,clip,keepaspectratio]{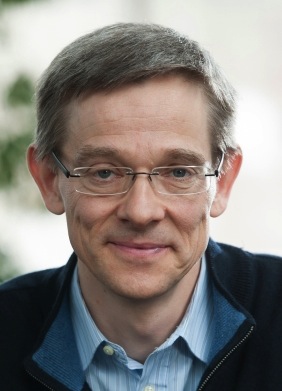}}]{Bernt Schiele}
received the masters degree in computer science from the University of Karlsruhe and INP Grenoble in 1994 and the PhD degree from INP Grenoble in computer vision in 1997. He was a postdoctoral associate and visiting assistant professor with MIT between 1997 and 2000. From 1999 until 2004, he was an assistant professor with ETH Zurich and, from 2004 to 2010, he was a full professor of computer science with TU Darmstadt. In 2010, he was appointed a scientific member of the Max Planck Society and the director at the Max Planck Institute for Informatics.
Since 2010, he has also been a professor at Saarland University. His main interests are computer vision, perceptual computing, statistical learning methods, wearable computers, and integration of multimodal sensor data. He is particularly interested in developing methods which work under realworld conditions.

\end{IEEEbiography}


%\newpage



%\vfill


\end{document}